\documentclass[final,3p,times]{elsarticle}

\usepackage{amssymb}
\usepackage{amsthm}

\usepackage{xcolor}
\usepackage{bbm}
\usepackage{subcaption}
\usepackage{array,multirow,graphicx}
\usepackage{float}
\usepackage{amsmath}
\usepackage{amsfonts}
\usepackage{amsmath}
\usepackage{multirow} 
\usepackage{graphicx}
\usepackage{url}
\usepackage{tabularx}
\usepackage{ltablex}

\journal{Neurocomputing}

\begin{document}

\begin{frontmatter}



\title{Beyond the Visible: A Survey on Cross-spectral Face Recognition}


\tnotetext[cor1]{Corresponding author.}


\author{David Anghelone\fnref{label1}\corref{cor1}}\ead{david.anghelone@inria.fr}
\author{Cunjian Chen\fnref{label2,label3}\corref{cor1}}\ead{cunjian.chen@monash.edu}
\author{Arun Ross\fnref{label2}}
\author{Antitza Dantcheva\fnref{label1}}


\affiliation[label1]{organization={Centre Inria d'Université Côte d'Azur},
             country={France}
            }

\affiliation[label2]{organization={Michigan State University},
            country={USA}
            }

\affiliation[label3]{organization={Monash University},
            country={Australia}
            }            
            
\begin{abstract}
Cross-spectral face recognition (CFR) refers to recognizing individuals using face images stemming from different spectral bands, such as infrared versus visible.
While CFR is inherently more challenging than classical face recognition due to significant variation in facial appearance caused by the \textit{modality gap}, it is useful in many scenarios including night-vision biometrics and detecting presentation attacks. Recent advances in deep neural networks (DNNs) have resulted in significant improvement in the performance of CFR systems. Given these developments, the contributions of this survey are three-fold. First, we provide an overview of CFR, by formalizing the CFR problem and presenting related applications. Secondly, we discuss the appropriate spectral bands for face recognition and discuss recent CFR methods, placing emphasis on \textit{deep neural networks}. In particular we describe techniques that have been proposed to extract and compare heterogeneous features emerging from different spectral bands. We also discuss the datasets that have been used for evaluating CFR methods. Finally, we discuss the challenges and future lines of research on this topic.
\end{abstract}





\end{frontmatter}

\section{Introduction}\label{sec:introduction}

\begin{figure}[t!]
    \centering
    \includegraphics[width=1\textwidth]{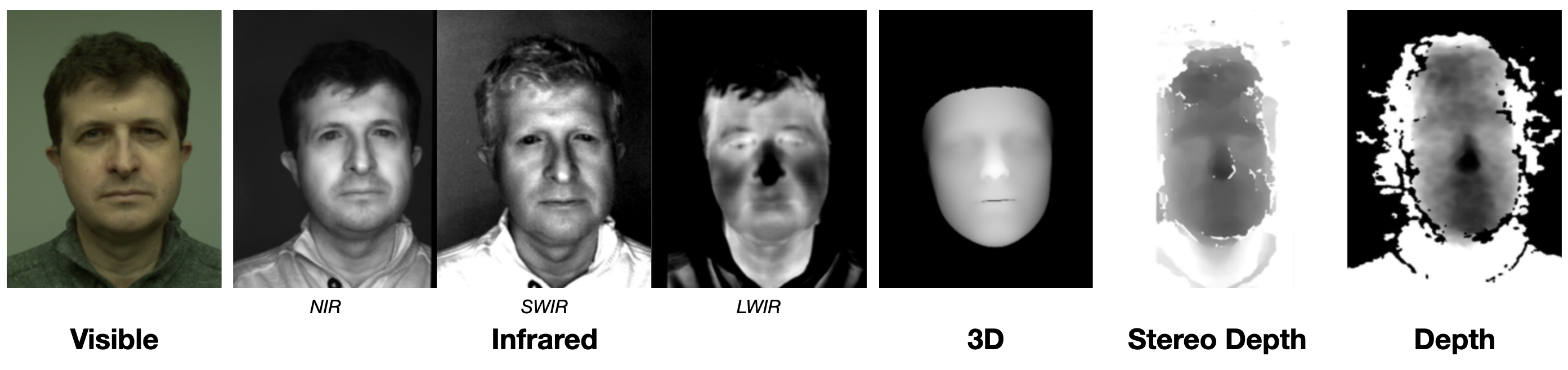}
    \caption{Examples of \textbf{heterogeneous face images} of the same subject captured in different modalities, i.e., domains. Images are from the MCXFace dataset~\cite{George_IEEETIFS_2022}.}
    \label{fig:modality}
\end{figure}

Face Recognition (FR) has been a very active research field for the last several decades \cite{Zhao2003,Abate07,Li2011,mei2018deep}. Advances in deep convolutional neural networks (CNNs) and the associated seminal work on DeepFace \cite{taigman2014deepface} have brought significant progress in this area, tackling a number of challenges including variations in pose, illumination and expression (PIE), as well as resolution and unconstrained settings.
While such work have predominantly focused on the visible (VIS) spectrum, considering additional spectra allows for increased robustness \cite{hu2017heterogeneous,BOURLAI201614}, in particular in the presence of different \textit{poses}, \textit{illumination variations}, \textit{noise}, as well as \textit{occlusions}. Further benefits include incorporating the \textit{absolute size of objects}, as well as \textit{robustness to presentation attacks} such as the use of makeup and masks to circumvent a system \cite{spinoulas2020multispectral}. 
Therefore, comparing RGB face images against face images acquired beyond the visible spectrum, often referred to as Cross-spectral Face Recognition (CFR), 
is of particular significance in designing FR systems for \textit{defense, surveillance, and public safety}~\cite{hu2017heterogeneous} and falls under the broader category of Heterogeneous Face Recognition (HFR) that addresses the broader concept of recognizing faces across diverse and non-uniform conditions.
Specifically, \textit{heterogeneous} refers to face images being represented in different modalities, e.g., visible, infrared, 2D-3D, resolutions, hand-drawn sketches, and depth (see Figure \ref{fig:modality}). 

The focus of this survey is on CFR, and specifically on comparing visible face images with images acquired beyond the visible spectrum, namely the infrared spectrum.
We note that CFR is more challenging than traditional FR for both,\textit{ human examiners}, as well as \textit{computer vision algorithms} due to the following three factors. Firstly, there is large \textit{intra-spectral variation}, where within the same modality, face samples of the same subject can exhibit larger appearance variations than face samples of different subjects.
Secondly, the \textit{modality gap} is of concern, where appearance variation between two face samples of the same subject can be larger than that of two samples belonging to two different subjects (see Figure~\ref{fig:modality}). This can result in degraded face recognition performance. 
Finally, the \textit{limited} availability of \textit{training samples} of cross-modality face image pairs can make it difficult to design effective CFR methods using deep neural networks.


\begin{figure}[htb!]
\centering
    \includegraphics[scale = 0.25]{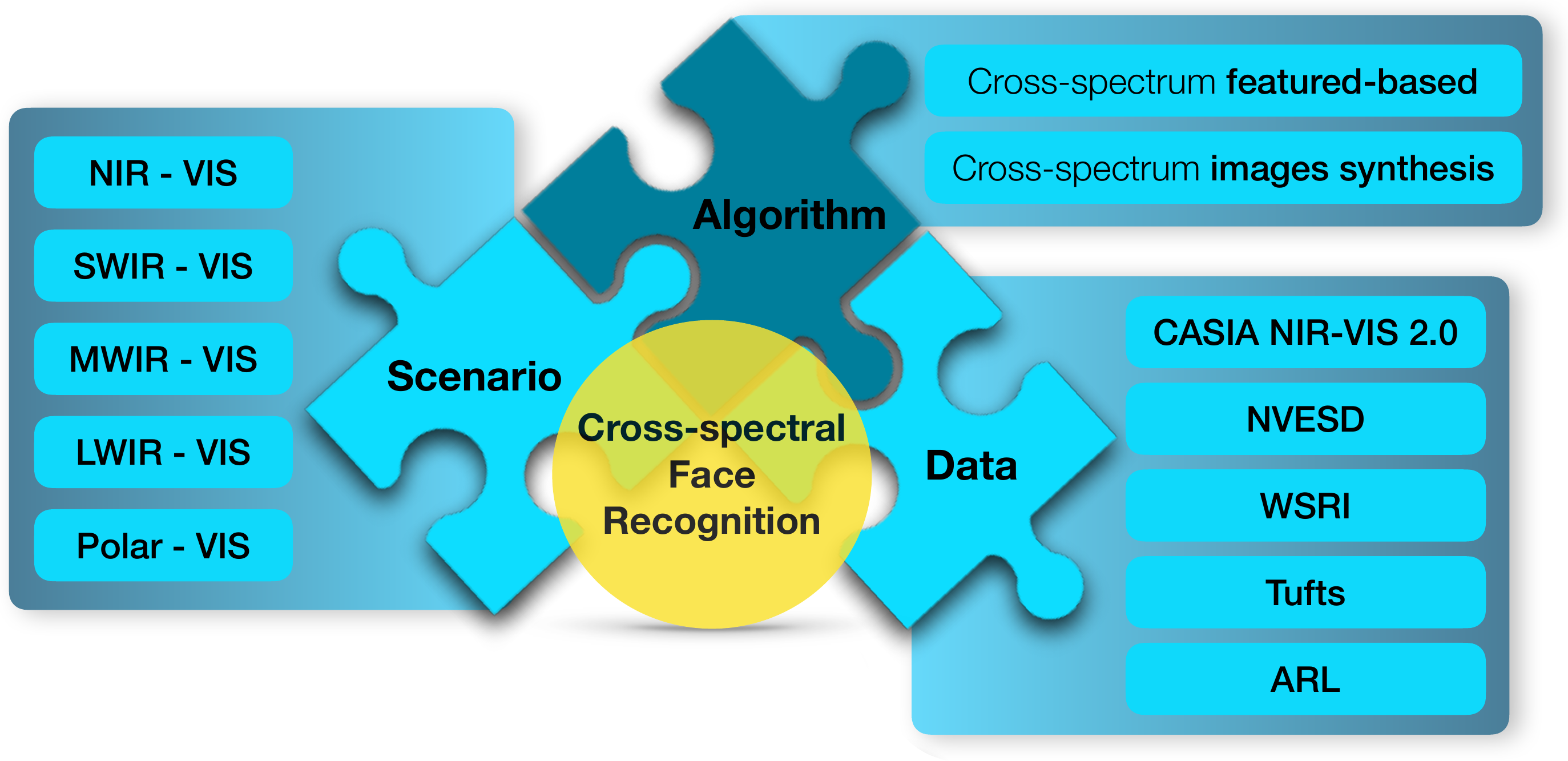}
    \caption{An overview of cross-spectral face recognition}
    \label{fig:overview}
\end{figure}

Recent advances in CNNs and generative adversarial networks (GANs) have allowed for remarkable improvements in CFR~\cite{ChenR19, ZhangRHSP19,He2019CrossspectralFC,DiAttribute2020}. The goal of this survey article is to provide an overview of these advances.
Specifically, we describe the proposed methods, used datasets, as well as the results and associated insights. In this context, we consider the electromagnetic spectrum ranging from near infrared (NIR) to more challenging bands such as midwave infrared (MWIR) and longwave infrared (LWIR), as illustrated in Figure~\ref{fig:overview}. 
Unlike other introductory overview articles including \cite{SHOJAGHIASS20142807,hu2017heterogeneous, munir2019extensive}, we focus on highlighting recent advances based on \textit{deep learning} in addressing CFR, and techniques that have been developed to bridge the modality gap.  Table \ref{tab:our_survey_contribution} revisits existing survey papers and highlights novel contributions we present in this survey.

In particular, we showcase the two main approaches in CFR: (i) cross-spectrum feature-based algorithms, which extract and compare features, and (ii) cross-spectrum image synthesis algorithms, which generate images from one spectrum to another for matching and recognition. 
The former involves comparing infrared probes against visible face image galleries within a common feature subspace. However, following the success of GAN-based image synthesis, the latter attempts to synthesize a ``pseudo-visible'' images from infrared images. These synthesized face images allow both, academic and commercial off-the-shelf (COTS) FR systems trained on the visible spectrum to be used for comparison.

\begin{table}[]
    \centering
    \caption{Summary of other related overview articles and the novelty of our work.}
    \begin{tabularx}{20cm}{|p{4cm}|p{5.5cm}|p{6cm}|}
    \hline
        \textbf{Survey reference} &  \textbf{Description} & \textbf{Novelty proposed by our survey}  \\ \hline \hline
        
        \textbf{R.S. Ghiass \textit{et al.} \cite{SHOJAGHIASS20142807} (2014)} Infrared Face Recognition: A Comprehensive Review of Methodologies and Databases
         & Reviews methodologies and databases for infrared face recognition, discussing the robustness of thermal imaging to facial expressions, pose variations, and the use of blood vessel patterns.
         & \textbf{Incorporation of Recent Deep Learning Advances:} We introduce the latest deep learning techniques, specifically tailored for CFR, surpassing traditional methods reviewed \cite{SHOJAGHIASS20142807}. \textbf{Focus on Multimodal Integration:} We explore the integration of multiple spectral bands, aimed at enhancing recognition accuracy and robustness, a more comprehensive approach than prior reviews.
         
         \\ \hline
            
        \textbf{S. Ouyang \textit{et al.} \cite{OuyangHSL16} (2016)} A Survey on Heterogeneous Face Recognition: Sketch, Infra-red, 3D and Low-resolution
         & Provides a detailed description of datasets and benchmarks relevant to evaluation and offers an extensive review of established techniques.
         & \textbf{Focus on CFR:} Our survey is focused specifically on CFR including NIR, SWIR, MWIR, LWIR and Polar-LWIR, compared to NIR \cite{OuyangHSL16}. \textbf{Deep Learning-based Approaches:} We revisit recently proposed deep learning methods for CFR, as opposed to traditional pattern recognition methods \cite{OuyangHSL16}.
         
         \\ \hline
         
        \textbf{S. Hu \textit{et al.} \cite{hu2017heterogeneous} (2017)} Heterogeneous Face Recognition: Recent Advances in Infrared-to-Visible Matching
        & Focuses on methods for matching infrared images to visible spectrum images, highlighting recent advances and challenges in heterogeneous face recognition.
        & \textbf{Bridging the Modality Gap:} We provide an in-depth analysis of state-of-the-art techniques for bridging the modality gap using deep learning. \textbf{Comprehensive Dataset Review:} We extensively revisit newly available datasets, enhancing the evaluation framework for CFR methods.
         
         \\ \hline 
         
        \textbf{R. Munir and R.A. Khan \cite{munir2019extensive} (2019)}
        An Extensive Review on Spectral Imaging in Biometric Systems: Challenges and Advancements
        & Discusses challenges and advancements in spectral imaging for biometric systems, including various spectral bands and their applications.
        & \textbf{Detailed Exploration of Cross-Spectral Techniques:} We delve into CFR methods with a focus on practical applications. \textbf{Future Directions and Challenges:} We outline future research directions and unresolved challenges, providing a road map for ongoing and future work in the field. \textbf{Comparison of Spectral Bands:} We here compare the effectiveness of different spectral bands, providing insights into optimal spectrum utilization for face recognition.
         \\ \hline 
    
    \end{tabularx}
    
    \label{tab:our_survey_contribution}
    
\end{table}

The remainder of this paper is organized as follows. We formalize CFR in Section \ref{sec: HFR Formalization} and present some applications in Section \ref{sec:application}. We discuss the various spectral bands, i.e., sensing modalities, that are relevant to face recognition, in Section \ref{sec:spectra}. In Section \ref{sec:HFR} we describe the general modules of a CFR system, viz., face image pre-processing and feature extraction.
The various deep learning based CFR methods that have been proposed in the literature are explained in Section \ref{sec:Reflective_algo} and \ref{sec:Emissive_algo}, while the relevant datasets used for research in this field are presented in Section \ref{sec:datasets}.
Finally, in  Section \ref{sec:Future Direction}, we discuss open research problems that are currently being addressed in the field of CFR.

\newpage
\section{Formalization of \textit{CFR}}\label{sec: HFR Formalization}
Inspired by previous work \cite{DSU2019, weiss2016survey}, we firstly formalize the FR-task and then the CFR-task as follows.

Let $\mathcal{M}$ be an electromagnetic spectral space modality associated with a marginal distribution $\mathbb{P}$ over a $d$-dimensional feature space $\mathcal{X} \subset \mathbb{R}^d$ and a label space $\mathcal{Y} \subset \mathbb{N}$.

Given a $n$-face database $X = \{ \textbf{x}_i \}_{i=1}^{n} $, where $\textbf{x}_{i} \in \mathcal{X}$, and their corresponding $n$-identities $Y = \{ y_j \}_{j=1}^{n} $ ($y_j \in \mathcal{Y}$), a FR-model is defined as a parametric function $\mathcal{F}_{FR,\Theta}$ described by the random variables $X$ (feature space), $Y$ (label space) and deep learning model parameters,  $\Theta$, which extract features from a CNN. 

\begin{equation}
\begin{array}{ccccc}
\mathcal{F}_{FR,\Theta} & : & X\times Y & \to & [0,1] \\
 & & (\textbf{x}_i,y_j) & \mapsto & \mathbb{P}(Y = y_j \vert X = \textbf{x}_i , \Theta), \\
\end{array} 
\end{equation}   

where, $i,j \in [1,n]$.

Thus, the FR-model aims to deduce parameters, {\color{blue} $\Theta$}, that increase the probability of correct recognition to 1, assuming that the correct face is in the gallery. More specifically, for all $k$-index identities $\in [1,n]$ 
 
\begin{equation}
    \mathcal{F}_{FR,{\color{blue} \Theta}}(\textbf{x}_k , y_k ) = \mathbb{P}(Y = y_k \vert X = \textbf{x}_k, {\color{blue} \Theta}) = 1.
\end{equation}

However, in the case of CFR, we compare acquired face images in two different modalities (or spectra or domains): the source $\mathcal{M}^s$ and the target $\mathcal{M}^t$. Each of them is associated with a marginal distribution $\mathbb{P}$ over the $d$-dimensional feature space for both source $X^s = \{ \textbf{x}_i^s\}_{i=1}^{n} \subset \mathcal{X}^s$ and target $X^t = \{ \textbf{x}_i^t\}_{i=1}^{n} \subset \mathcal{X}^t$, respectively. Note that they share the same set of labels $Y = \{ y_j \}_{j=1}^{n} \subset \mathcal{Y}$. 

The CFR-function, $\mathcal{F}_{CFR,\Theta}$, is formalized as follows. Towards finding the deep learning model parameters, {\color{red} $\Theta$}, for all $k$-index identities $\in [1,n]$, we have

\begin{equation}\label{eq:HFR_formalization}
\mathbb{P}(Y = y_k \vert X^s = \textbf{x}_k^s, {\color{red} \Theta}) = \mathbb{P}(Y = y_k \vert X^t = \textbf{x}_k^t, {\color{red} \Theta}) = \textit{t},
\end{equation}
where, $\textit{t} \in [0,1]$ is a predefined threshold.

Algorithms reviewed in this paper seek to provide strategies for determining {\color{red} $\Theta$}, formalized in Equation \eqref{eq:HFR_formalization}.

During enrollment, a \textit{gallery face image} is captured and processed and being stored as \textit{reference template}, representing the biometric information of an individual.
At the time of authentication, a \textit{probe face image} is captured and processed in the same way as in the enrollment and compared against a reference template of a claimed identity (verification) or against all stored reference templates (identification) \cite{jain2004introduction}.

Mathematically, both heterogeneous verification and identification are formalized as follows. 
The user is either \textit{genuine} -- the claim is true, or an \textit{impostor} -- the claim is false. They are represented by the set $\Omega = \{ \omega_{true} , \omega_{false} \}$, respectively.

In \textit{CFR}-verification, an input feature vector $\textbf{x}_q^{t} \in \mathcal{X}^{t}$ stemming from the target modality $\mathcal{M}^t$ and a claimed identity $y_{j} \in \mathcal{Y}$ are embedded as the pair $(\textbf{x}_q^{t},y_{j})$. The comparison algorithm (Section \ref{subsubseq:Convolutional neural networks}) analyzes the extracted templates $\textbf{x}_q^{t}$, and a function $\mathcal{S}$ (Section \ref{seq:Face Matching}) computes a \textit{similarity score} \eqref{eq:similarity score} according to the stored source template $\textbf{x}_j^{s} \in X^s$. If the score exceeds a predefined threshold $t$, it is considered a match. Thus,

\begin{equation}
(\textbf{x}_q^{t}, y_{j}) \in 
\begin{cases}
\omega_{true}, &\text{if \ } \mathcal{S}(\textbf{x}_j^{s} , \textbf{x}_q^{t}) \geq t \\
\omega_{false}, & \text{otherwise}.
\end{cases}
\end{equation}

In \textit{CFR}-identification, an input feature vector $\textbf{x}_q^{t} \in \mathcal{X}^{t}$ derived from the target modality $\mathcal{M}^t$ is provided to the biometric system. The system attempts to search in the gallery for the appropriate identity $y_k$ correlated with $\textbf{x}_k^{s} \in X^s$, where $k \in [1,n]$. The associated template is able to provide the highest similarity score \eqref{eq:similarity score}. In addition, the highest score exceeding the predefined threshold $t$ leads to a match.
Otherwise the user is classified as unknown $y_{unknown}$. Thus,

\begin{equation}
\textbf{x}_q^{t} \in 
\begin{cases}
y_k, & \text{if \ } \underset{k \in [1,n]}{max}\{   \mathcal{S}(\textbf{x}_k^{s} , \textbf{x}_q^{t}) \geq t \}   \\
y_{\text{unknown}}, & \text{otherwise}.
\end{cases}
\end{equation}







\section{Applications of \textit{CFR}}\label{sec:application} 

\begin{figure}
    \centering
    \includegraphics[width=1\textwidth]{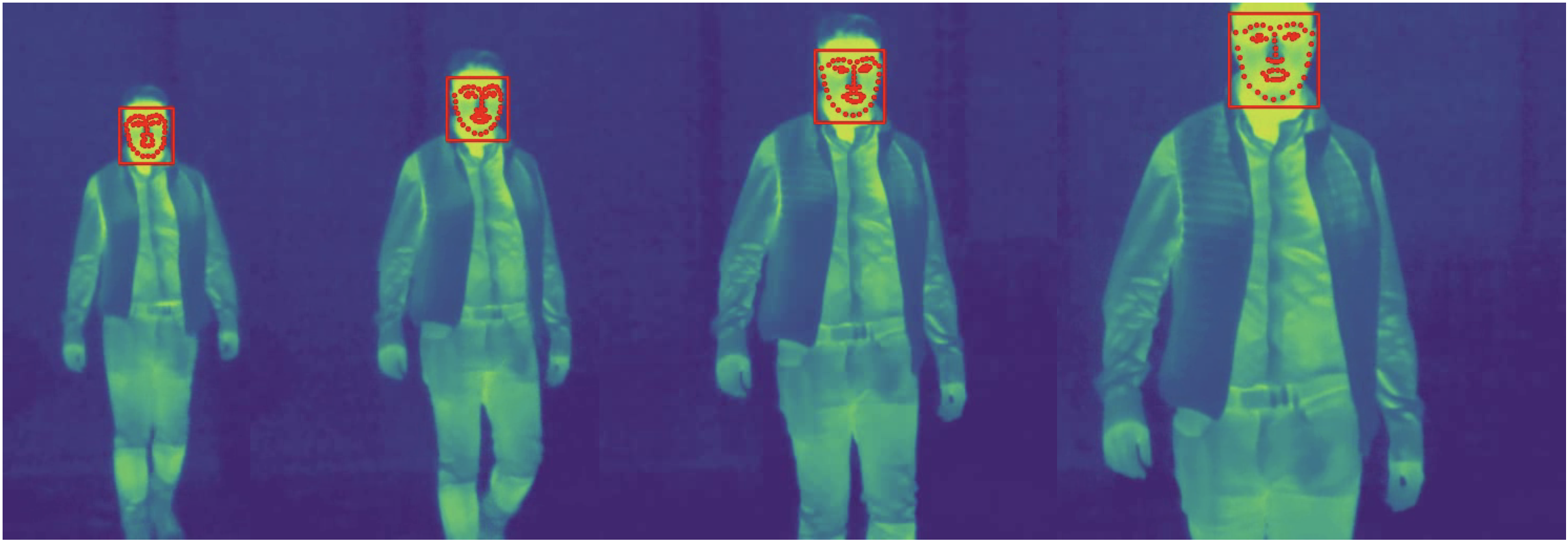}
    \caption{A monitoring system equipped with a thermal sensor applies face and facial landmark detection to a video sequence captured in the wild at night, successfully detecting the person despite challenging conditions, paving the way for subsequent identification steps. Images are from Anghelone \textit{et al.}~\cite{anghelone2022tfld}.}
    \label{fig:TFLD_David_detection}
\end{figure}

\textit{FR}-systems have been widely deployed and operate mainly in the visible spectrum \cite{prati2019sensors}, due to the ubiquity of low-cost cameras.
As motivated in the introduction, \textit{CFR}-systems have been beneficial in additional applications, for example, in \textit{Intelligence Video Surveillance} (IVS). 
Beyond \textit{access control} where acquisition is controlled w.r.t. user collaboration, the application of CFR has strong potential to provide new capabilities in \textit{law enforcement}, \textit{military} and the \textit{AI community}, especially for commercial purposes. 
In this context, IR sensors offer several advantages over traditional visible light imaging. They enable long-range monitoring and capture biometric modalities at variable standoff distances. IR sensors can also capture images through certain types of obscuring materials and perform well in low-light or nighttime scenarios. However, identifying individuals in real-world situations often involves uncooperative behavior. The unconstrained environment necessitates capturing images of subjects at a distance from the camera, accounting for various facial poses, expressions, and even occlusions.

\paragraph{Law Enforcement and Homeland Security}
A pertinent application for CFR has been in law enforcement and homeland security \cite{krivsto2018overview}. In this context, IR-sensors can be used for long-range monitoring and capturing biometric modalities at variable standoff distances, as well as in low-light and night-time scenarios. These settings are depicted in Figure \ref{fig:TFLD_David_detection}. In such cases, the acquired image often has to be compared against a gallery consisting of visible spectrum face images for identification purposes. This capability enhances the effectiveness of monitoring suspects and ensuring security in public spaces. 

\paragraph{Military Applications}
Similar applications are found in military settings involving identifying individuals in challenging environments \cite{BOURLAI201614}. Night-vision devices, which often operate in the infrared spectrum, can capture images in complete darkness. These images can be cross-referenced with visible spectrum databases, facilitating the identification and tracking of persons of interest in combat or surveillance operations.

\paragraph{Commercial Applications}
In the commercial sector, CFR can improve security measures in places such as shopping malls and airports. AI-powered security systems using CFR can manage access control more effectively, recognizing individuals across different lighting conditions and enhancing overall safety.

\paragraph{Presentation Attack Detection}
CFR-systems are also beneficial for deflecting presentation attacks \cite{george2020face, Heusch_TBIOM_2020}, which involve attempts to deceive biometric systems using photographs, masks, or other fraudulent methods. By leveraging images from multiple spectral bands, these systems can better discern between genuine and fake biometric data, enhancing security measures against such attacks.

\paragraph{Sketch-Visual Identification}
On a wider scale, HFR (which is a superset of CFR) offers notable application in sketch-visual identification. Law enforcement agencies can use HFR to match a suspect's sketch with a visible spectrum face image from a mugshot database. This method is particularly useful when photographic evidence is unavailable, and only a hand-drawn sketch of the suspect exists \cite{ouyang2016survey}.






\section{Different Spectra}\label{sec:spectra}

\begin{figure*}
    \centering
    \includegraphics[width=1\textwidth]{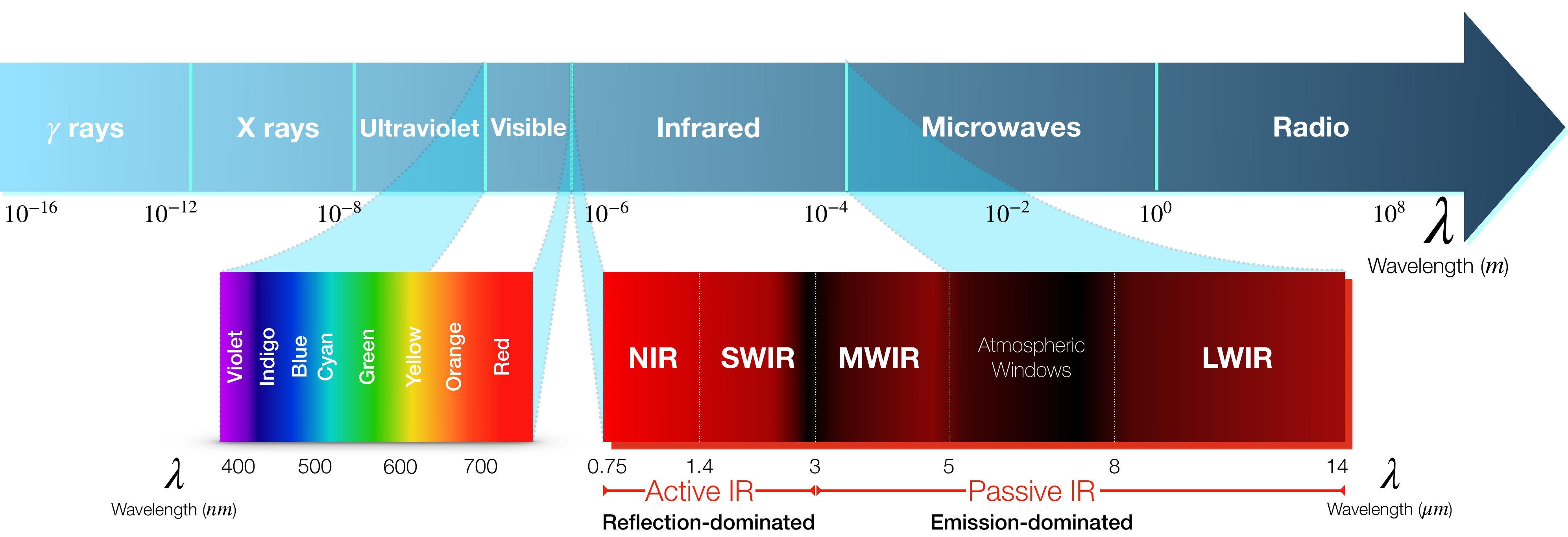}
    \caption{\textbf{Electromagnetic spectrum.} Modalities and their associated wavelengths, highlighting the visible and infrared radiations.}
    \label{fig:electromagnetic spectrum}
    \label{tab:IR_spectrum}
\end{figure*}

We next proceed to introduce the infrared (IR) spectral bands, which we illustrate in Figure \ref{fig:electromagnetic spectrum}, that have been employed in CFR. 

Wolff \textit{et al.} \cite{ wolff2005face}, Buddharaju \textit{et al.} \cite{4107566}, Kong \textit{et al.} \cite{KONG2005103},  Bhowmik \textit{et al.} \cite{bhowmik2011thermal_SPECTRA}, and Bourlai and Hornak \cite{BOURLAI201614} describe  \textit{infrared light} as an invisible, heat-associated energy that can be sensed when radiation or warmth are reflected or emitted from an object. 
Unlike ultraviolet rays \cite{narang2015can} , IR-waves penetrate the skin without damage to health.
We note that, in principle, IR sensors capture either the face-\textit{reflection} of infrared light or the heat face-\textit{emission} stemming from the subcutaneous superficial blood vessels. In particular, face contains blood capillaries and forms a subcutaneous network. More physiology-details including the uniqueness of individual's physiology can be found in the work of Buddharaju \textit{et al.} \cite{4107566}.

IR bands have been defined according to the standard \textit{ISO}-20473:2007 as near-infrared (NIR) between	0.78–3 $\mu$m, mid-infrared (MIR) between 3–50 $\mu$m and far-infrared (FIR) 50–1000 $\mu$m.
Deviating from that, the \textit{CIE}-International Commission on Illumination has divided the IR-spectrum into the bands IR-A (0.7 $\mu$m – 1.4 $\mu$m), IR-B (1.4 $\mu$m – 3 $\mu$m) and IR-C 	
(3 $\mu$m – 1000 $\mu$m). 

In this paper, we use a division of the IR bands based on physical properties, which determine \textbf{reflection-dominated} and \textbf{emission-dominated} regions. These regions, also denoted as \textit{Passive} and \textit{Active} IR, respectively, are further divided into sub-spectra, which we summarize in Figure \ref{tab:IR_spectrum}. 
We note that a similar scheme has been employed by manufacturers of IR sensors, where specific sensors have been developed, 
associated to active or passive IR. 
  




\begin{figure}[b]
    \centering
    \includegraphics[width=1\textwidth]{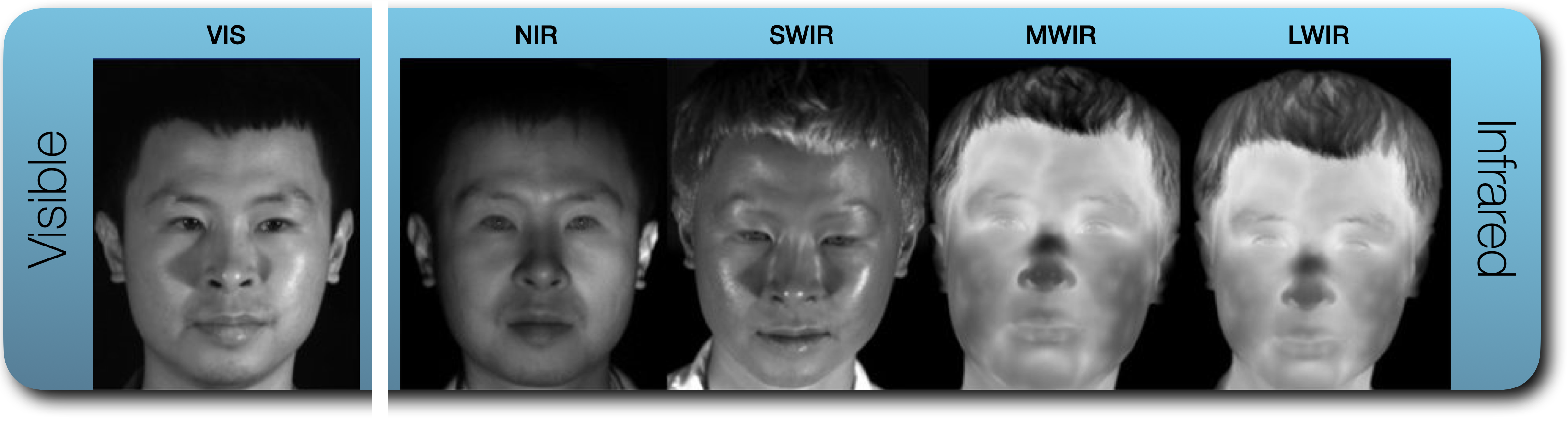}
    \caption{\textbf{Face beyond the visible}. \ A comparison of a face in visible (VIS) and infrared bands, viz., \textit{NIR}, \textit{SWIR}, \textit{MWIR} and \textit{LWIR}. We note the different physiological properties in both \textbf{Active IR} (NIR, SWIR) and \textbf{Passive IR} (MWIR, LWIR) bands. 
    Figure credit: Hu \textit{al.} \cite{hu2017heterogeneous}}
    \label{fig:NIR_SWIR_MWIR_LWIR_face}
    \vspace{-0.3 cm}
\end{figure}


\textbf{Experiment 1.} In order to demonstrate the modality gap, we conduct the following experiment inspired by Hu \textit{et al.} \cite{hu2017heterogeneous}. We select heterogeneous images pertaining to one subject from the Tufts dataset~\cite{Tufts}.
Hence, for the HFR scenarios \textit{Visible/Visible, Visible/NIR, Visible/LWIR and Visible/Sketch}, we compute both (i) the associated structural similarity (SSIM) scores and (ii) the cosine similarity measures from facial features extracted with CNNs.

Towards evaluating the modality gap in CFR related to (i), the SSIM metric \eqref{eq:SSIM}, as proposed by Wang \textit{et al.} \cite{Wang2004_SSIM}, has been widely used and offers similarity at the pixel level. In our context, SSIM indicates the similarity between a visible spectrum reference image $x$ and a gallery image $y$ (in our case infrared) by assessing the perturbation of structural information caused by factors such as luminance and contrast. Specifically,

\begin{equation}\label{eq:SSIM}
    SSIM(x,y) = \dfrac{(2\mu _x \mu _y + c_1)(2\sigma _{xy} + c_2)}{(\mu _{x}^2 + \mu _{y}^2 + c_1)(\sigma _{x}^2 + \sigma _{y}^2 + c_2)},
\end{equation}
\\ 
where, $\mu _{.}$ and $\sigma _{.}$ denote mean and standard deviation, respectively, and $\sigma _{xy}$ represents the covariance of the respective images. Here, $c_1$ and $c_2$ are constants. 
SSIM \eqref{eq:SSIM} ranges between $-1$ and $1$, with $1$ being the extreme case, when compared images $x$ and $y$ are identical. 

\begin{figure}[h!]
    \centering
    \includegraphics[width=0.80\textwidth]{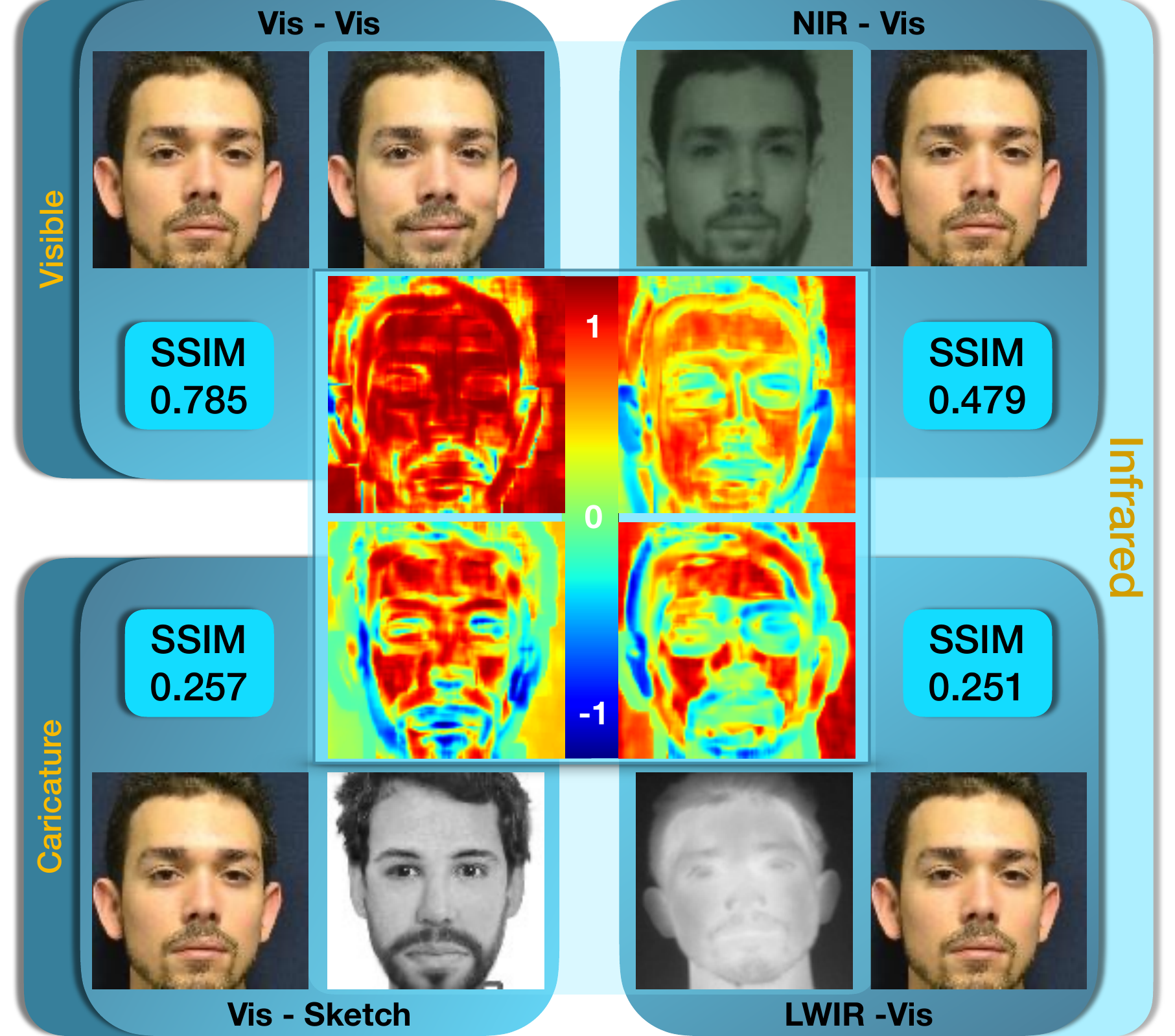}
    \caption{\textbf{Structure Similarity (SSIM) maps and scores.}  A visualization strategy to understand the human and machine visual perception when comparing two images in different scenarios: \textit{Visible} - against \textit{Visible}/\textit{NIR}/\textit{LWIR}/\textit{Sketch}.  }
    \label{fig:SSIM_score}
\end{figure}

The structural similarity maps and associated SSIM-scores computed in Experiment 1 are shown in Figure \ref{fig:SSIM_score}, highlighting face regions that are most sensitive to infrared energies. While the wavelength in the infrared band increases, the SSIM-difference increases as well. We observe different emission patterns around nose, mouth and eyes (blue panel). Furthermore, the results suggest that the largest challenges (lowest SSIM-scores) have to do with comparing an image in the visible spectrum with a sketch as well as with a thermal image. In both cases, we obtain similar SSIM-scores. 

\ 
\\
Furthermore, we proceed to showcase the spectral modality gap related to CFR in an additional experiment (ii) by utilizing the cosine similarity measure (formalized below in Equation \ref{eq:similarity score}). 
Specifically, facial biometric features are extracted from associated facial images illustrated in Figure \ref{fig:SSIM_score} with a state-of-the-art face matcher, ArcFace \cite{deng2019arcface}.
ArcFace is used to extract deep features from a visible face, as well as from its counterpart, which are then compared using the cosine similarity measure. 
The cosine similarity measure is computed based on facial features extracted from distinct spectral modalities, offering valuable insights into the biometric-level similarity of face representations. 
Table \ref{tab:modality_gap_with_cosine_similarity} reports scores pertaining to the different scenarios shown in Figure \ref{fig:SSIM_score}. We note that the scores are normalized between 0 to 1, where 1 indicates identical images.
As expected, we observe that similarity decreases significantly in cases when facial images sensed in different spectra or modalities (sketch-visual) are compared, indicating that comparison is more challenging in such cases.


\setcounter{table}{1}
\begin{table}[h!]
    \caption{Computation of cosine similarity S scores using state-of-the-art face matcher ArcFace from a visible spectrum image and its counterpart.}
    \centering
    \begin{tabular}{|c|c|c|c|c|}
    \hline
         & Vis-Vis & Vis-NIR & Vis-LWIR & Vis-Sketch \\ \hline \hline
        \textbf{Score of similarity} & $0.9545$ & $0.6652$ & $0.1475$ & $0.1024$ \\ \hline
    \end{tabular}
    \label{tab:modality_gap_with_cosine_similarity}
\end{table}



\subsection{Active IR} 
The active IR region comprises of \textit{near infrared} (NIR) and \textit{shortwave infrared} (SWIR), characterized by reflective material properties, which require an (invisible) IR light to reveal the scene. 



\textbf{NIR} is placed right next to the red-channel of the visible spectrum, with increased wavelength, which is exhibited in the visual similarities between VIS and NIR images (Figure \ref{fig:modality}). 
We observe this in Experiment 1 when comparing \textit{VIS-NIR} faces, where the SSIM score is 0.479 (Figure \ref{fig:SSIM_score}). 
NIR is instrumental in CFR, as it remains invariant to lighting-direction and low-light conditions and, hence, is employed for  monitoring and night-surveillance \cite{4107567}.

\textbf{SWIR}-imagery, similar to NIR, is visually close to visible imagery. See Figure \ref{fig:NIR_SWIR_MWIR_LWIR_face}. The SWIR-band is significantly wider than the NIR-band and enables sensing in atmospherically challenging conditions such as \textit{rain}, \textit{fog}, \textit{mist}, \textit{haze}, and common urban particulates such as \textit{smoke} and \textit{pollution} \cite{bhowmik2011thermal_SPECTRA}. 
In addition, SWIR-sensors are able to capture objects in extremely low-light conditions, which renders SWIR suitable for surveillance, allowing for long-range applications ($<$ one kilometer \cite{6272346}), as well as 
for identification purposes \cite{SWIR_2019_SPIE}.

\subsection{Passive IR } 
Beyond a wavelength of $\lambda = 3 \mu m$, the IR band is significantly emissive and aims to acquire heat-sensitive radiation emitted from a human face. Hence, thermal sensors can be designed to capture temperature variations across the facial skin tissue due to the underlying face vasculature \cite{4107566}. One of the benefits of the emission-dominated regions are that thermal imagery can be passively acquired without any external illumination, in either day or night environments \cite{wolff2005face}. Passive IR is comprised of Midwave (MWIR) and Longwave (LWIR). Note however that the range between MWIR and LWIR (ranging between $\lambda = 5 \mu m - 8\mu m$, see Figure \ref{fig:electromagnetic spectrum}) constitutes a strong atmospheric absorption window rendering image acquisition impossible. 

\textbf{MWIR} is located between SWIR and LWIR, and has both reflective and emissive properties - allowing for the sensing of different facial-skin-features including vein patterns \cite{bourlai2013mid_MWIR}. However, it is the emissive component that is predominantly exploited in CFR, which is why face images captured in MWIR tend to largely resemble heat face signatures, and are visually distinct from visible imagery (see Figure \ref{fig:NIR_SWIR_MWIR_LWIR_face}).





\textbf{LWIR} extends the infrared band up to $\lambda = 14 \mu m$ and consists of exclusively emitted radiation. This shift introduces high within-class variations and notably impedes a match with visible images (Figure \ref{fig:NIR_SWIR_MWIR_LWIR_face}). We observe that \textit{Vis-LWIR} is the most challenging scenario as per Figure \ref{fig:SSIM_score}, with the lowest SSIM score of 0.251. LWIR imagery is visually similar to that of MWIR with respect to shape and contrast (Figure \ref{fig:NIR_SWIR_MWIR_LWIR_face}). 



\subsection{Polarimetric Thermal Imagery}

\begin{figure*}[]
    \centering
    \includegraphics[width=1\textwidth]{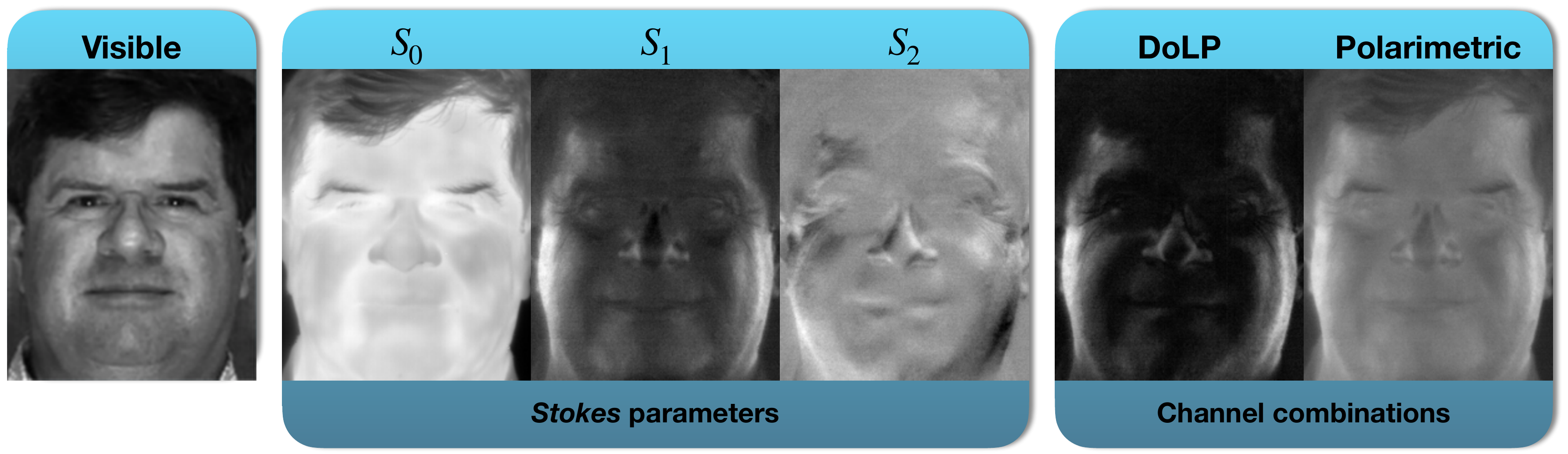}
    \caption{\textbf{Stokes parameters}. Example images of a subject captured in the visible spectrum and their corresponding conventional thermal $S_{0}$, polarization state information $S_{1}$ and $S_{2}$, and polarimetric thermal images. Related SSIM-scores are reported in Table \ref{tab:SSIM_POLAR}.}
    \label{Pola_State}
    \vspace{-0.25 cm}
\end{figure*}

Polarimetric thermal imagery aims to measure the polarization state information of the light contained in the thermal infrared spectrum, as well as in any other spectrum. By performing optical filtering of the light at precise rotational angles, say $0^{\circ}, 90^{\circ}, 45^{\circ}$ and $-45^{\circ}$ related to the horizontal axes, more geometrical and textural details are obtained, which is able to enhance the conventional intensity-based thermal images for face recognition purposes. Towards inferring polarimetric-information in the passive IR and capturing the thermal radiance, specific infrared polarimetric sensors have been used. Through light manipulation, the polarization state information is typically expressed in terms of \textit{Stokes parameters} $\Vec{S}$, as firstly introduced by G. Stokes in 1852. For more details, the reader is referred to the work of Pezzaniti and Chenault \cite{pezzaniti2020polarization}. $\Vec{S}$ is derived from the following linear combination.

\begin{equation}
\Vec{S} 
=
\begin{bmatrix}
S_{0} \\
S_{1} \\
S_{2} \\ 
S_{3}
\end{bmatrix}
=
\begin{bmatrix}
I_{0}^{\circ } + I_{90}^{\circ } \\
I_{0}^{\circ } - I_{90}^{\circ } \\
I_{45}^{\circ } + I_{-45}^{\circ } \\
I_{R}^{\circ } + I_{L}^{\circ }
\end{bmatrix},
\end{equation}
\\
where, $I_{0}^{\circ }$ and $I_{90}^{\circ }$ denote the horizontal and vertical polarized light, respectively, while $I_{45}^{\circ }$ and $I_{-45}^{\circ }$ represent the diagonal polarized light and $I_{R}^{\circ }$ and $I_{L}^{\circ }$, respectively, represent the intensity of the right and left circularly polarized light. The \textit{Stokes parameter}, $S_{0}$, represents the conventional thermal images without polarization information, $S_{1}$ and $S_{2}$ convey orthogonal polarimetric information and display additional details. Finally, $S_3$ is considered to be $0$ in most applications, due to lack of circular information in the thermal spectrum.
The degree of linear polarization (DoLP) can then be computed as
\begin{equation}
    DoLP=\frac{\sqrt{S_1^2+S_2^2}}{S_0},
\end{equation}
and it denotes the spectral radiation window of the electromagnetic spectrum which is linearly polarized. 
The polarimetric image entails the channel combinations of $S_0$, $S_1$, and $S_2$ (see Figure \ref{Pola_State}). 

\textbf{Experiment 2.} We compute SSIM scores between \textit{Visible} and \textit{Thermal} images (including each polarization state and degree of linear polarization), which we report in Table \ref{tab:SSIM_POLAR}. The scores confirm the improvement due to the use of {\em polarimetric} thermal imagery.
We note that SSIM pertaining to polarimetric imagery, outperforms the one due to thermal imagery. Therefore, polarimetric thermal imagery can facilitate CFR, which will be later confirmed in Section \ref{seq:Face Matching}. 

\begin{table}[h!]
\caption{\textbf{Experiment 2}. Computation of cosine similarity scores using state-of-the-art face matchers and structural similarity SSIM for images depicted in Figure \ref{Pola_State}. The scores are normalized to [0,1].}
\label{tab:SSIM_POLAR}
    \centering
    \begin{tabular}{|c||c|c|c|c|c|c|}
    \cline{3-7}
   \multicolumn{2}{c|}{} & \multicolumn{3}{c|}{Stokes parameters} & \multicolumn{2}{c|}{Channel combinations} \\ \hline
        \multirow{2}{*}{\textbf{Face Matchers}} & \textit{Vis}/ & \multicolumn{3}{c|}{\textit{Vis}/}  & \multicolumn{2}{c|}{\textit{Vis}/} \\
        & Vis & $S_0$ & $S_1$ & $S_2$ & DoLP & Polar \\ \hline  \hline 
        \textbf{SphereFace~\cite{liu2017sphereface}} & 1.0 & 0.499 & 0.473 & 0.486 & 0.461 & 0.698 \\ \hline          
        \textbf{AM-Softmax~\cite{ChenR19}} & 1.0 & 0.543 & 0.576 & 0.555 & 0.596 & 0.702 \\ \hline  
        \textbf{MobileFaceNet~\cite{ChenR19}} & 1.0 & 0.505 & 0.499 & 0.503 & 0.507 & 0.702 \\ \hline \hline     
        \textbf{ArcFace~\cite{deng2019arcface}}  & 1.0 & 0.641 & 0.658 & 0.587 & 0.67 & 0.684 \\ \hline \hline  
        \textbf{SSIM} & 1.0 & 0.286 & 0.236 & 0.262 & 0.209 & 0.300 \\ \hline 
    \end{tabular}
    \vspace{-0.25 cm}
\end{table}

\section{Cross-spectral Face Recognition}\label{sec:HFR}
As in classical face recognition, \textit{face detection} and \textit{alignment} are considered the first and foremost steps of CFR. CFR data is preprocessed by aligning and cropping faces to reference templates based on detected facial landmarks. Then, a standard pipeline proceeds in learning/extracting face features by using optimal network architectures and loss functions. 

\subsection{Face Preprocessing}
Very limited work has focused on facial landmark detection beyond the visible spectrum~\cite{PosterCVPRW2019,Kopaczka2019, ChuMMSP2019, anghelone2022tfld}. Since NIR and SWIR face images closely resemble visible spectrum faces, landmark detection methods developed for visible face images are directly applicable. 

\begin{figure}[h]
    \centering
    ~ 
    \begin{subfigure}[b]{1\textwidth}
        \includegraphics[width=1\textwidth]{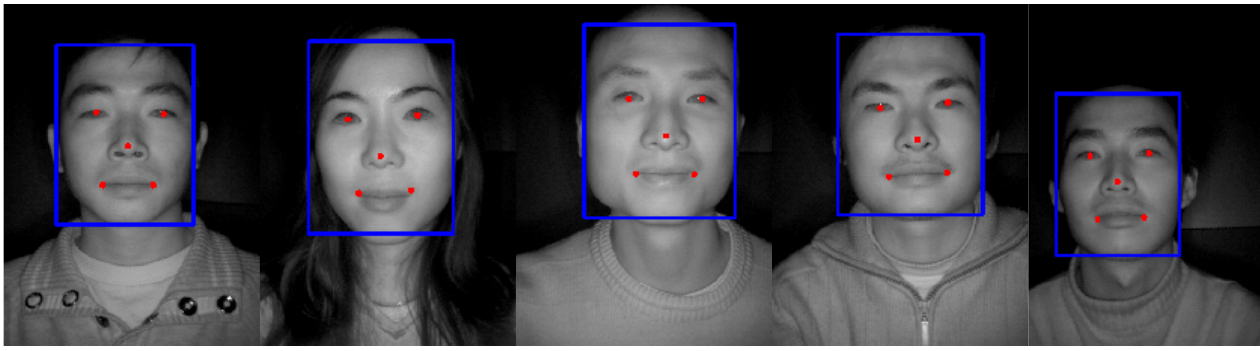}
        \caption{Face and landmark detection}
        \label{casia_nir_landmark}
    \end{subfigure}
    ~ 
    \begin{subfigure}[b]{1\textwidth}
        \includegraphics[width=1\textwidth]{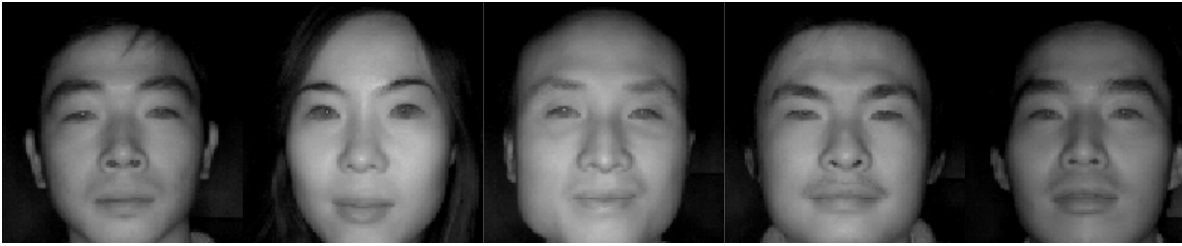}
        \caption{Face normalization}
        \label{casia_nir_cropped}
    \end{subfigure}
    \caption{\textbf{Experiment 3.} Face and landmark detection experiments conducted on near-infrared faces sampled from~\cite{CasiaNIRVIS}. Face normalization was performed by aligning the faces to the reference pose based on the detected landmarks.}\label{casia_nir_alignment}
    \vspace{-0.2 cm}
\end{figure}

\textbf{Experiment 3.} Towards demonstrating the possibility of adapting landmark detection methods developed for the visible spectrum, we utilize MTCNN~\cite{MTCNN2016} to perform face detection and landmark detection experiments on NIR faces. Next, an affine transformation is computed based on the detected landmarks to normalize the detected faces to a frontal pose (see Figure~\ref{casia_nir_alignment}). 

\begin{figure}[h!]
    \centering
    \includegraphics[width=1\textwidth]{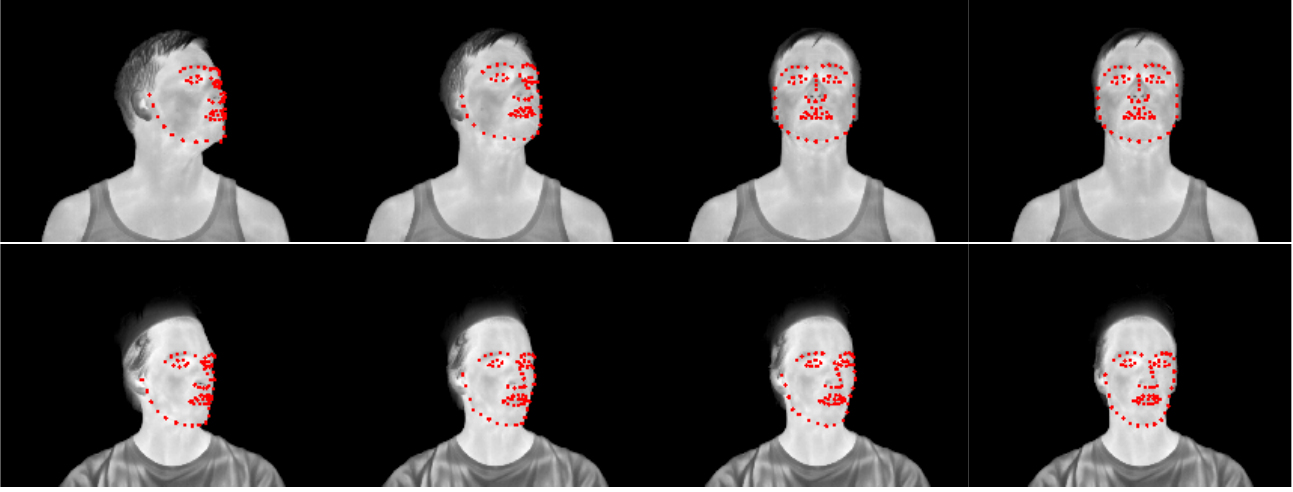}
    \caption{\textbf{Landmark detection in thermal imagery}. Example images from a thermal dataset~\cite{Kopaczka2019} showing 68-point ground-truth landmark annotations.}
    \label{fig:ir_landmark}
    \vspace{-0.4 cm}
\end{figure}

As demonstrated, while it might be relatively easy to perform landmark detection on NIR or SWIR face images, it remains a challenge in the context of thermal face images, including MWIR and LWIR. We note that thermal face images tend to have lower contrast, lower resolution and lack texture information~\cite{PosterCVPRW2019}. Poster et al.~\cite{PosterCVPRW2019} analyzed the performances of three modern deep learning based landmark detection methods on thermal face images, namely \textit{Deep Alignment Network} (DAN), \textit{Multi-Task CNN} (MTCNN), and \textit{multi-class patch based fully CNN} (PBC). These landmark detection algorithms were evaluated on the \textit{ARL polarimetric thermal face} dataset \cite{Polarimetric_Thermal_Face} with 5-landmarks.  Kopaczka et al.~\cite{Kopaczka2019} assembled a high-resolution thermal facial image dataset in the LWIR spectral band and manually annotated 68 landmark points. DAN was used as the representative deep learning landmark detector to train and evaluate on the assembled dataset. Chu and Liu~\cite{ChuMMSP2019} proposed to use U-Net as a backbone to detect 68 facial landmarks on the same dataset used in~\cite{Kopaczka2019}. The landmark detection task was designed as a multi-task learning framework by integrating emotion recognition. Poster et al.~\cite{Poster2021} used a coupled convolutional network architecture to leverage visible face data to train a thermal-only face landmark detection model. Examples of 68-point ground-truth landmark annotations are depicted in Figure~\ref{fig:ir_landmark}. 

\subsection{Deep Feature Extraction}\label{subsubseq:Convolutional neural networks}
After face images are preprocessed, two different schemes can be leveraged to compute domain-invariant features. Firstly, faces from different spectral bands can be directly ingested by a CNN, aiming to learn a shared feature representation scheme. Secondly, faces can be transformed by GANs into a target domain and features can be extracted from the transformed images. 

\subsubsection{Convolutional Neural Networks (CNNs)}\label{subsubseq:CNN}
\begin{figure}[h!]
    \centering
    \includegraphics[width=1\textwidth]{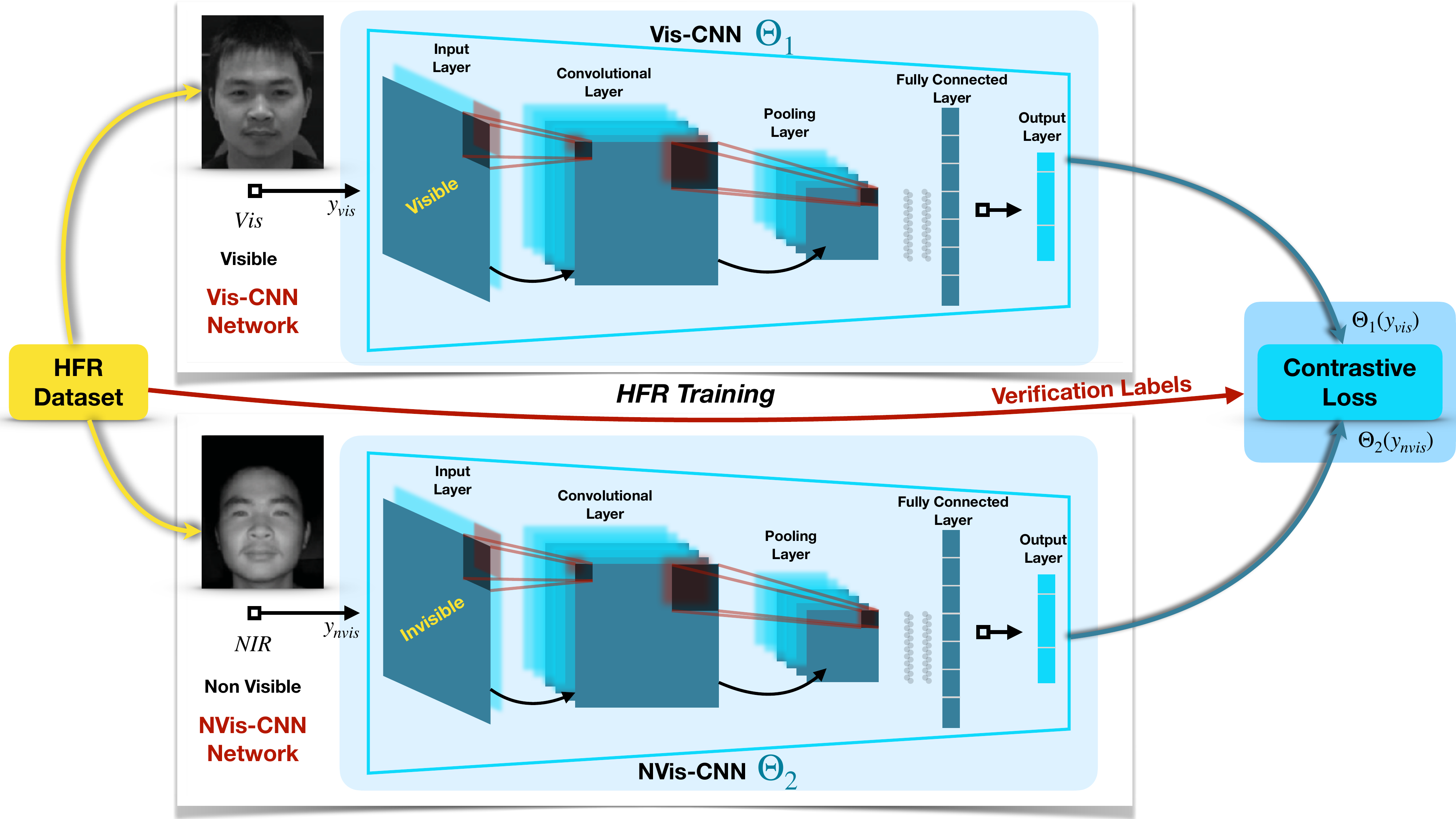}
    \caption{\textbf{CNNs.} Flowchart depicting a two-branch convolutional neural network, learning domain-invariant feature representation.}
    \label{fig:coupledCNN}
    \vspace{-0.20 cm}
\end{figure}

\textit{CNNs} are a class of neural networks that are highly efficient in learning effective feature representations for FR~\cite{liu2017sphereface, ChenR19, deng2019arcface}. For CFR, CNNs operate on specific pairs of data, e.g., NIR-VIS, formed by different spectral bands. Depending on the number of pairs, CNNs can be designed with a two-branch~\cite{2016_NIR_HFR, IDR2017, he2018wasserstein} or a three-branch architecture~\cite{Transferring_deep_rpz}. Weights from individual branches are often shared. An example of such a design is depicted in Figure~\ref{fig:coupledCNN}. Visible and NIR images represent respectively input to the network, extracting features. To learn domain-invariant features that are independent of the given spectral bands, a contrastive loss function is applied to enforce the identity constraint for these two extracted feature vectors~\cite{2016_NIR_HFR}. For triplet inputs formed by anchor, positive and negative samples, a triplet loss is generally applied to minimize the gap between different spectral bands~\cite{Transferring_deep_rpz}.  
There are several prominent CNN-architectures that can be used as backbone for the architecture in Figure~\ref{fig:coupledCNN}, some of which we enlist in Table \ref{tab:CNN_Architecture}. 
\begin{table}[h!]
\caption{\textbf{Representative architectures} used in CFR for both CNNs and GANs. }
\label{tab:CNN_Architecture}
    \centering
    \begin{tabular}{|c|l|c|p{0.3\linewidth}|}
    \hline
     \textbf{Year} &\textbf{Backbone Architecture} & \textbf{Source} & \textbf{Used in}  \\ \hline \hline
       \multirow{2}{*}{2014} & VGGNet & \cite{simonyan2014very}& \cite{Lezama_2017_CVPR, Bihn2018, DSU2019, TRGAN2020} \\  \cline{2-4}
        & GoogLeNet & \cite{szegedy2015going}& \cite{2016_NIR_HFR} \\   \hline
       \multirow{2}{*}{2015} & ResNet & \cite{he2016deep}& \cite{MMDL2019, MCCNN2019, cho2020relational}\\ \cline{2-4}
        & U-Net &\cite{RonnebergerFB15} & \cite{UNET2020},~\cite{TVGAN2018},~\cite{ThermalToVisible_autoencoders},~\cite{di2018apgan}\\ \hline
       2016 & Inception-ResNet & \cite{szegedy2016rethinking}& \cite{hu2019heterogeneous,DSU2019, ADCANs_2020} \\ \hline
       \multirow{2}{*}{2017} & DenseNet & \cite{huang2017densely}& \cite{CpGAN_US_Army, ZhangRHSP19} \\ \cline{2-4}
       & Squeeze-and-Excitation Network & \cite{hu2018squeeze} & \cite{10008000, anghelone2023ANYRES}\\ \hline
       2018 & LightCNN &\cite{wu2018light}& \cite{Transferring_deep_rpz,IDR2017,he2018wasserstein, CDL_AAAI_2018, He2019CrossspectralFC, DSU2019, cho2020relational} \\ \hline

        2019 & EfficientNet & \cite{tan2019efficientnet} & - \\ \hline
        
       2020 & Vision Transformer & \cite{dosovitskiy2020image} & \cite{Liu_2021_ICCV, luo2022memory} \\ \hline
       2021 & SwinTransformer & \cite{Liu_2021_ICCV} & - \\ \hline
    \end{tabular}
    \vspace{-0.35 cm}
\end{table}

\subsubsection{Generative Adversarial Networks (GANs)}\label{subsec:GAN}
\textit{GANs} are deep learning frameworks introduced by Goodfellow \textit{et al.} \cite{goodfellow2014generative} (2014), which consist of two sub-networks, a \textit{Generative} model \textit{G} and a \textit{Discriminative} model \textit{D} trained simultaneously. Together, \textit{G} and \textit{D} compete with each other in a minimax game. Specifically, \textit{G} aims to learn the intrinsic distribution $p_G$ over some target data $x \sim p_{data} $, and associated with a noise prior $z \sim p_z$. $G$ draws sample $z$ for creating synthetic data $G(z)$. 
In the meantime, $D$ aids the training by taking $x$ or $G(z)$ as input and performing a binary classification as to whether the input is from the real data distribution $p_{data}$ or from the generated data distribution $p_G$.
$G$ attempts to mislead $D$ to not be able to distinguish generated images from real, while $D$ attempts to make that distinction. 

The optimization competition of these two models results in the following minimax game 
\begin{equation}\label{eq:GAN metric}
    \underset{G}{min} \  \underset{D}{max} \ \mathbb{E}_{x \sim p_{data(x)}}[log(D(x))] + \mathbb{E}_{z \sim p_z}[log(1-D(G(z)))].
\end{equation}




Traditional GANs learn a mapping from a random noise input $z$ to an output image $y$, while conditional GANs learn the mapping from an observed image $x$ and a random noise $z$ in which both $G$ and $D$ are conditioned to an output image $y$ according to 
\begin{small}
\begin{equation}\label{eq:GAN metric conditional}
    \underset{G}{min} \  \underset{D}{max} \ \mathbb{E}_{x \sim p_{data(x)}}[log(D(x \vert y))] + \mathbb{E}_{z \sim p_z}[log(1-D(G(z \vert y)))].
\end{equation}
\end{small}

\begin{figure}[h!]
    \centering
    \includegraphics[width=1\textwidth]{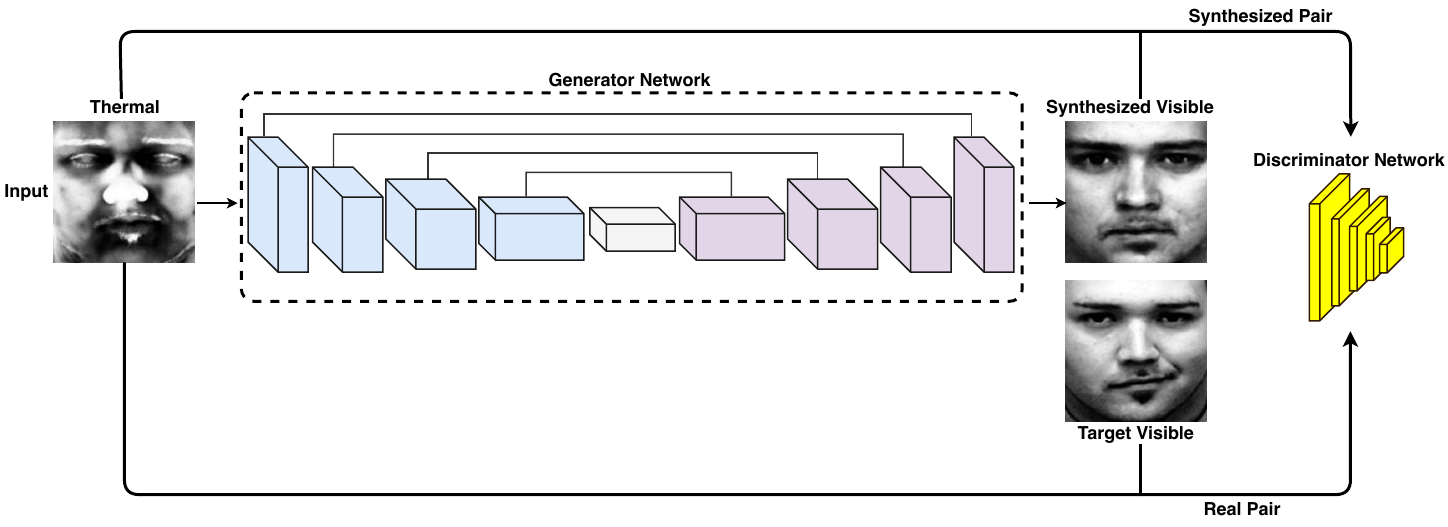}
    \caption{\textbf{GANs.} Flowchart depicting a conditional generative adversarial network for synthesizing visible faces from conditional infrared (thermal) faces.}
    \label{fig:GAN framework Cunjian}
\end{figure}

Many existing techniques have been developed based on conditional GANs to address the CFR problem (see Figure~\ref{fig:GAN framework Cunjian}). The generator takes a thermal image as input, seeking to produce a synthesized visible image as output. The discriminator is trained to distinguish between two pairs of images: the real pair, consisting of an input thermal image and a target visible image, and the synthesized pair, consisting of an input thermal image and a synthesized visible image~\cite{TVGAN2018, di2018apgan, ChenR19, DiAttribute2020}. As noted in these works, there were notable differences in the use of architectures and loss functions. Some variants of architectures such as multi-scale generator~\cite{DiAttribute2020} and multi-scale discriminator~\cite{ZhangRHSP19} were used in order to account for scale variance. There also exist variants of loss functions, including attribute loss~\cite{DiAttribute2020}, identity loss~\cite{ChenR19} and shape loss~\cite{CycleGANDetector2018}. 

\subsubsection{Loss Functions for Cross-spectral Face Recognition}\label{subsec:Loss}

\begin{table*}[h!]
    \caption{\textbf{Loss Functions.} A comprehensive list of loss functions $\mathcal{L}$ from traditional FR to CFR. Here, $N$ is the number of samples, $T$ is the number of samples in a mini-batch, $W$ is the weight matrix, $b$ is the bias term, $x_{i}$ and $y_{i}$ are the $i^{th}$ training sample and the according class label, respectively. $\theta _{y_{i}} \in \left[ \frac{k \pi}{m}, \frac{(k+1)\pi}{m} \right]$ with $k \in [0, m-1]$ and $m \geq 1$. }
    \label{tab:loss_function}
    \centering
    \small
    \begin{tabular}{|c| l c |c|l|c|}
    \hline
       \textbf{Task} &
       \textbf{Name} & \textbf{Notation} & \textbf{Reference} &  \textbf{Function $\mathcal{L}$} \\ \hline \hline

        \multirow{14}{*}{FR}
         
        & Softmax (Cross-Entropy) & $ \mathcal{L}_s$ &  \cite{sun2014deep_Cross_entropy_Loss}  & $-\dfrac{1}{N} \sum \limits_{i=1}^{N} \text{log}\left( \dfrac{e^{W^{T}_{y_i}x_i + b_{y_i}} }{\sum _{j=1}^{n} e^{W^{T}_{j}x_i + b_{j}} }\right) $ \\ 
         \cline{2-5}
         
         & Center Loss & $ \mathcal{L}_{ce}$ & \cite{wen2016discriminative_Center_Loss} & $\dfrac{1}{2} \sum \limits_{i=1}^{N} \Vert x_{i} - c_{y_i} \Vert ^{2}_{2} $   \\ 
         \cline{2-5}
         
         & Marginal Loss &$\mathcal{L}_{ma}$ & \cite{deng2017marginal_Marginal_Loss} & $\dfrac{1}{m^2 - m} \sum \limits_{i,j, i\neq j}^{m} \left( \xi - y_{ij}\left( \theta - \Vert \frac{x_i}{\Vert x_i \Vert} -  \frac{x_j}{\Vert x_j \Vert} \Vert^{2}_{2} \right)   \right) $  \\
         \cline{2-5}
         
         & Angular Softmax Loss (SphereFace) & $\mathcal{L}_{as}$ & \cite{liu2017sphereface} &  $- \dfrac{1}{N} \sum\limits_{i=1}^{N} \dfrac{e^{\Vert x_{i} \Vert \text{cos}(m \theta _{y_{i}})}}{e^{\Vert x_{i} \Vert \text{cos}(m \theta _{y_{i}})} + \sum _{j=1, j \neq y_{i}}^{n} e^{ \Vert x_{i} \Vert \text{cos}(\theta _{j}) } }$  \\
         \cline{2-5}
         
         & Large Margin Cosine Loss (CosFace) & $\mathcal{L}_{co}$ & \cite{Wang_2018_CVPR_cosface} &$ - \dfrac{1}{N} \sum\limits_{i=1}^{N} \text{log} \left( \dfrac{e^{s(\text{cos}(\theta _{y_i}) -m )}}{ e^{s(\text{cos}(\theta _{y_i}) -m )} + \sum _{j=1,j\neq y_i}^{n} e^{s \ 
         \text{cos}(\theta _j)}} \right)$  \\
         \cline{2-5}
         
         & Additive Angular Margin Loss (ArcFace) & $\mathcal{L}_{aa}$ & \cite{deng2019arcface} & $-\dfrac{1}{N} \sum \limits_{i=1}^{N} \text{log}\left( \dfrac{e^{s \ \text{cos}(\theta _{y_i}+m)}}{e^{s \ \text{cos}(\theta _{y_i}+m)} + \sum _{j=1 , j \neq y_i} e^{s \ \text{cos}(\theta_j)}} \right)$  \\
         \hline
         \hline
         
        \multirow{10}{*}{CFR}
        
         & Contrastive Loss & $\mathcal{L}_{ct}$ &  \cite{2016_NIR_HFR} - \eqref{loss:ct} & $
\begin{cases}
\frac{1}{2}\Vert f(x_i)-f(x_j)\Vert^2, &\text{if \(y_i=y_j\)} \\
\frac{1}{2} max(0,m-\Vert f(x_i)-f(x_j)\Vert )^2, &\text{else}.
\end{cases}$ \\
        \cline{2-5}

        & Triplet Loss & $\mathcal{L}_{tp}$ & \cite{Transferring_deep_rpz} - \eqref{loss:tp} &  $\sum \limits _{i=1}^{N}  \left( \Vert f(x_{i}^{a})-f(x_{i}^{p})\Vert - \Vert f(x_{i}^{a})-f(x_{i}^{n}) \Vert +m \right)$ \\ \cline{2-5}
        
         & Tetrad Margin Loss & $\mathcal{L}_{TML}$ & \cite{MMDL2019} - \eqref{loss:TML} &  $\begin{aligned}
 & \sum_{i \in \{ N,V \} }^{T} \left(\frac{z_{j}^{N}\cdot z_{j}^{V}}{\Vert z_{j}^{N}\Vert \cdot \Vert z_{j}^{V} \Vert } - \frac{z_{j}^{N}\cdot z_{l}^{V}}{\Vert z_{j}^{N}\Vert \cdot \Vert z_{l}^{V} \Vert }+m_1\right) \\
 & +\sum_{i \in \{ N,V \} }^{T} \left(\frac{z_{j}^{N}\cdot z_{j}^{V}}{\Vert z_{j}^{N}\Vert \cdot \Vert z_{j}^{V} \Vert } - \frac{z_{j}^{V}\cdot z_{l}^{V}}{\Vert z_{j}^{V}\Vert \cdot \Vert z_{l}^{V} \Vert }+m_2\right)
\end{aligned} $  \\ \hline
    \end{tabular} 
\end{table*}

CNNs (Section \ref{subsubseq:CNN}) and GANs (Section \ref{subsec:GAN}) constitute fundamental deep learning networks. Once such network architectures are designed, suitable loss functions are chosen to minimize prediction errors. Therefore, one active direction in CFR has to do with the development of efficient \textit{loss functions}~\cite{Survey_Deep_FR}. Loss functions are pertinent in deep architectures, as they aim to measure how well algorithms are able to perform correct predictions. In the CFR-context, the loss function is specifically designed to mitigate cross-spectral impact, in order to minimize intra-class variation and maximize inter-class variation~\cite{2016_NIR_HFR,Transferring_deep_rpz,MMDL2019}.

Loss functions used in CFR are frequently adapted from traditional FR. We note that the predominantly used loss function in traditional FR remains to be \textit{softmax}, which is targeted to maximize the probability that the same subject belongs to the target class. Therefore, it encourages the separability of features. \textit{Center loss}~\cite{wen2016discriminative_Center_Loss} seeks to minimize distances between deep features and their corresponding class centers by simultaneously learning a center for individual classes. However, we note that center loss does not encourage separability of features. Therefore, it is common practice to perform a joint supervision of the softmax loss with center loss.  Some other approaches combine euclidean margin-based losses with softmax loss to perform joint supervision. Recently, it has been shown that the features learned by softmax loss have an intrinsic angular distribution~\cite{liu2017sphereface}. Since then, a flurry of angular-based loss functions have been proposed to impose 
discriminative constraints on a hypersphere manifold~\cite{liu2017sphereface, Wang_2018_CVPR_cosface, deng2019arcface}. We summarize loss functions in Table \ref{tab:loss_function}. 

In most traditional FR systems, samples belonging to the same subject are treated without any differentiation. However, in CFR, samples are divided into different categories based on their spectrum information. This prompts the use of loss functions that can accept a pair of inputs~\cite{2016_NIR_HFR}, triplet of inputs~\cite{Transferring_deep_rpz}, as well as quadruplet of inputs\cite{MMDL2019}. These loss functions have achieved impressive performance. Admittedly, directly using loss functions from traditional FR for CFR is based on the premise that facial appearance from other spectral bands is close to the visible spectral band. This is applicable in the case of comparing NIR against VIS faces. With increase in wavelength, difference in the facial appearance also starts to increase. Thus, adversarial loss functions used in GAN can be used to perform the synthesis.

\subsection{Face Comparison}\label{seq:Face Matching} 

Trained CNNs or GANs for CFR produce deep feature representations for a given probe-image, which are then compared with those of the gallery images. Using the same notation introduced in Section \ref{sec: HFR Formalization}, let $X^s = \{ \textbf{x}_i^s\}_{i=1}^{n} \subset \mathcal{X}^s$ and $X^t = \{ \textbf{x}_i^t\}_{i=1}^{n} \subset \mathcal{X}^t$ denote the set of samples from the \textit{Source} (visible) and the \textit{Target} (infrared) modalities, respectively. The corresponding shared set of labels are denoted by $Y = \{ y_j \}_{j=1}^{n} \subset \mathcal{Y}$. 

Suppose $\Theta$ denotes the deep process of extracting $d$ features from a CNN, then the similarity score between two templates can be computed as a function $\mathcal{S}$ referred to as the \textit{measure of similarity} ranging between $0$ and $1$, where $1$ represents a high similarity. 
As an example, the \textit{cosine distance} is widely used to calculate such similarity, for all $k$-index identities $\in [1,n]$:

\begin{equation}\label{eq:similarity score} 
\begin{aligned}
\mathcal{S}(\textbf{x}_k^s , \textbf{x}_k^t) &= \cos (\textbf{s},\textbf{t}) = \dfrac{< \textbf{s} , \textbf{t} >}{ \Vert \textbf{s} \Vert  \Vert \textbf{t} \Vert  } \\ 
& = \frac{ \sum_{j=1}^{d} s_j t_j }{ \sqrt{\sum_{j=1}^{d}{(s_j)^2}} \sqrt{\sum_{j=1}^{d}{(t_j)^2}} },
\end{aligned}
\end{equation}

where, $\textbf{s} = \Theta (\textbf{x}_k^s)$ and $\textbf{t} = \Theta (\textbf{x}_k^t) $. Table \ref{tab:SSIM_POLAR} reports cosine similarity scores of state-of-the-art face matchers, when comparing thermal images depicted in Figure \ref{Pola_State}. 
Other Euclidean distance based measures such as $L_2$ and $L_1$ can also serve as similarity metrics.

\section{Algorithms of Different Spectra}
FR operating on imagery beyond the visible spectrum can be categorized as \textit{cross-spectrum featured-based algorithms} or \textit{cross-spectrum images synthesis algorithms} (Figure \ref{fig:overview}). The former involve comparison of infrared probe against a gallery of visible face images, within a common feature subspace. However, after the success of image synthesis based on GANs, cross-spectrum images synthesis algorithms have attempted to synthesize a ``pseudo-visible" image from an infrared image. Given the synthesized face image, both academic and commercial-off-the-shelf (COTS) FR-systems trained on visible spectrum, can be utilized for comparison. Furthermore, intensive research has been done to exploit the polarimetry thermal images to further improve CFR. 

\ 


\subsection{Reflective IR-to-Visible FR}\label{sec:Reflective_algo}

\subsubsection{Near Infrared}
\begin{figure*}[t]
    \centering
    \includegraphics[width=1\textwidth]{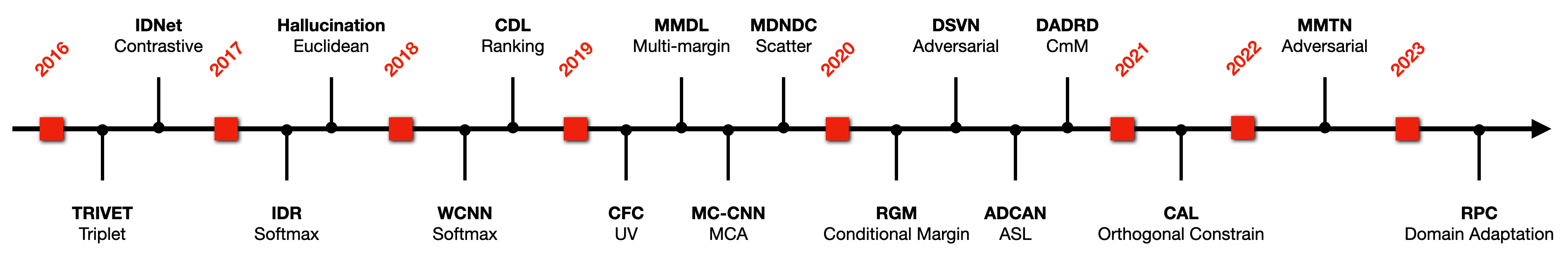}
    \caption{\textbf{Timeline of algorithm development.} NIR-to-VIS FR with their corresponding loss functions.}
    \label{fig:Timeline_NIR}
\end{figure*}

The appearance difference between NIR and VIS face images is less pronounced compared to other spectral bands and, hence, shared feature representations have often been proposed~\cite{Transferring_deep_rpz, 2016_NIR_HFR, IDR2017,he2018wasserstein, CDL_AAAI_2018}. CNNs
are suitable for NIR-to-VIS CFR scenario, since they seek to automatically extract representative face features. Nevertheless, a challenge remains, which has to do with the illumination-variation between the two modalities. We proceed to discuss a number of recent work, constituting prominent state-of-the-art approaches in matching NIR against VIS face images. 
These algorithms along with their individual loss functions are summarized in Figure~\ref{fig:Timeline_NIR}. Further, we also list in Table~\ref{NIR_VIS_comparisons} the NIR-to-VIS face comparison performances on public benchmark datasets.


\begin{table*}[t]
        \caption{Rank-1 Accuracy and Verification Rate on different datasets for the \textbf{NIR-VIS} face comparison.}
    \label{NIR_VIS_comparisons}
    \centering
    \tiny
    \begin{tabular}{|c|c|c|c|l||c|c|c|c|}
    \hline
        \textbf{Year} & \textbf{Authors} & \textbf{Method} & \textbf{Loss} & \textbf{Dataset} &   \multicolumn{3}{c|}{\textbf{Performance}} \\
         & & & & & Rank-1 & FAR 1$\% $ & FAR 0.1$\% $ \\ \hline \hline
         
         \multirow{5}{*}{2016} & \multirow{3}{*}{Liu \textit{et al.} \cite{Transferring_deep_rpz}} & \multirow{3}{*}{TRIVET}& \multirow{3}{*}{Triplet} & CASIA NIR-VIS 2.0 \cite{CasiaNIRVIS} & 95.74 & 98.1 & 91.03 \\ \cline{5-8}
         &&&& Oulu-CASIA NIR-VIS \cite{chen2009learning} & 92.2 & 67.9 & 33.6 \\  \cline{5-8}
         &&&& BUAA-VisNir face  \cite{huang2012buaa} & 93.9& 93.0 & 80.9 \\ \cline{2-8} 
         & \multirow{2}{*}{Reale \textit{et al.} \cite{2016_NIR_HFR}} & \multirow{2}{*}{IDNet} & \multirow{2}{*}{Contrastive} & CASIA NIR-VIS 2.0 \cite{CasiaNIRVIS} & 87.1 & - & 74.5 \\ \cline{5-8}
         &&&& CASIA HFB  \cite{CASIA_HFB} & 97.58 & 96.9 & 85.0 \\ \hline \hline
         
         \multirow{4}{*}{2017} & \multirow{3}{*}{He \textit{et al.} \cite{IDR2017}} & \multirow{3}{*}{IDR} & \multirow{3}{*}{Softmax} & CASIA NIR-VIS 2.0 \cite{CasiaNIRVIS} & 97.3 & 98.9 & 95.7 \\ \cline{5-8}
         &&&& Oulu-CASIA NIR-VIS \cite{chen2009learning} & 94.3 & 73.4 & 46.2 \\  \cline{5-8}
         &&&& BUAA-VisNir face  \cite{huang2012buaa} & 94.3 & 93.4 & 84.7 \\ \cline{2-8}
         & Lezama \textit{et al.} \cite{Lezama_2017_CVPR} & Hallucination & Euclidean &  CASIA NIR-VIS 2.0 \cite{CasiaNIRVIS} & 96.41 & - & - \\ \hline \hline
         
         \multirow{6}{*}{2018} & \multirow{3}{*}{He \textit{et al.} \cite{he2018wasserstein}}& \multirow{3}{*}{WCNN} & \multirow{3}{*}{Softmax} & CASIA NIR-VIS 2.0 \cite{CasiaNIRVIS} & 98.7 & 99.5 & 98.4 \\   \cline{5-8}
          &&&& Oulu-CASIA NIR-VIS \cite{chen2009learning} & 98.0 & 81.5 & 54.6 \\  \cline{5-8}
          &&&& BUAA-VisNir face  \cite{huang2012buaa} & 97.4& 96.0 & 91.9 \\ \cline{2-8}
         & \multirow{3}{*}{Wu \textit{et al.} \cite{CDL_AAAI_2018}}& \multirow{3}{*}{CDL} & \multirow{3}{*}{Ranking} & CASIA NIR-VIS 2.0 \cite{CasiaNIRVIS} & 98.6 & - & 98.32 \\ 
         \cline{5-8}
         &&&& Oulu-CASIA NIR-VIS \cite{chen2009learning} & 94.3 & 81.6 & 53.9 \\  \cline{5-8}
         &&&& BUAA-VisNir face  \cite{huang2012buaa} & 96.9 & 95.9 & 90.1 \\ \hline \hline
         
        \multirow{8}{*}{2019} & \multirow{3}{*}{He \textit{et al.} \cite{He2019CrossspectralFC}} & \multirow{3}{*}{CFC (GAN)} & \multirow{3}{*}{UV} & CASIA NIR-VIS 2.0 \cite{CasiaNIRVIS} & 98.6 & 99.2 & 97.3 \\   \cline{5-8}
          &&&& Oulu-CASIA NIR-VIS \cite{chen2009learning} & 99.9 & 98.1 & 90.7 \\  \cline{5-8}
          &&&& BUAA-VisNir face  \cite{huang2012buaa} & 99.7 & 98.7& 97.8 \\ 
         \cline{2-8}
         & \multirow{2}{*}{Cao \textit{et al.} \cite{MMDL2019}} & \multirow{2}{*}{MMDL} & \multirow{2}{*}{Multi-margin} & CASIA NIR-VIS 2.0 \cite{CasiaNIRVIS} & 99.9 & - & 99.4 \\ \cline{5-8} 
         &&&& Oulu-CASIA NIR-VIS \cite{chen2009learning} & 100.0 & - & 97.2 \\  \cline{2-8}

         & Deng \textit{et al.} \cite{MCCNN2019} & MC-CNN & MCA & CASIA NIR-VIS 2.0 \cite{CasiaNIRVIS} & 99.22 & - & 99.27 \\ \cline{2-8}
         
         &  \multirow{2}{*}{Hu \textit{et al.} \cite{hu2019heterogeneous}} &  \multirow{2}{*}{MDNDC} & \multirow{2}{*}{Scatter} & CASIA NIR-VIS 2.0 \cite{CasiaNIRVIS} & 98.9 & 99.6 & 97.6 \\   \cline{5-8}
          &&&& Oulu-CASIA NIR-VIS \cite{chen2009learning} & 99.8 & 88.1 & 65.3 \\ \hline \hline
         
         \multirow{12}{*}{2020} & \multirow{2}{*}{Cho \textit{et al.} \cite{cho2020relational}} & \multirow{2}{*}{RGM} & \multirow{2}{*}{Conditional-margin} & CASIA NIR-VIS 2.0 \cite{CasiaNIRVIS} &99.3& 99.51 & 99.02 \\   \cline{5-8}
         
          &&&& BUAA-VisNir face  \cite{huang2012buaa} & 99.67 & 99.22& - \\ \cline{2-8}
         
         & \multirow{2}{*}{Hu \textit{et al.} \cite{HuDSVN2020}} & \multirow{2}{*}{DSVN} & \multirow{2}{*}{Adversarial} & CASIA NIR-VIS 2.0 \cite{CasiaNIRVIS} & 99.0 & 99.7 & 98.6 \\ \cline{5-8}
          &&&& Oulu-CASIA NIR-VIS \cite{chen2009learning} & 100.0 & 99.3 & 95.5 \\  \cline{2-8}

         & \multirow{2}{*}{Iranmanesh \textit{et al.} \cite{CpGAN_US_Army}} & \multirow{2}{*}{CpGAN} & Contrastive+Coupled& CASIA NIR-VIS 2.0 \cite{CasiaNIRVIS} & 96.63 & - & 87.05 \\   \cline{5-8}
         
            &&& +GAN+Eucl.+Perceptual & CASIA HFB  \cite{CASIA_HFB} & 99.64 & 98.4 & 89.7 \\ \cline{2-8}
            
        & \multirow{3}{*}{ Hu \textit{et al.} \cite{ADCANs_2020}} & \multirow{3}{*}{ADCANs} & \multirow{3}{*}{ASL} & CASIA NIR-VIS 2.0 \cite{CasiaNIRVIS} & 99.1 & 99.6 & 98.5 \\ \cline{5-8}
       &&&& Oulu-CASIA NIR-VIS \cite{chen2009learning} & 99.8 & 93.2 & 78.9 \\  \cline{5-8}
          &&&& BUAA-VisNir face  \cite{huang2012buaa} & 99.8 & 99.7& 98.4 \\ \cline{2-8}
    
    & \multirow{3}{*}{ Hu \textit{et al.} \cite{hu2020dual_DADRD}} & \multirow{3}{*}{DADRD} & \multirow{3}{*}{CmM} & CASIA NIR-VIS 2.0 \cite{CasiaNIRVIS} & 99.10 & 99.6 & 98.6 \\ \cline{5-8}
       &&&& Oulu-CASIA NIR-VIS \cite{chen2009learning} & 100.0 & 98.5 & 92.9 \\  \cline{5-8}
          &&&& BUAA-VisNir face  \cite{huang2012buaa} & 99.9 & 99.9 & 99.8 \\ \hline \hline

     \multirow{3}{*}{2021} & \multirow{3}{*}{He \textit{et al.} \cite{he2021coupled}} & \multirow{3}{*}{CAL} & \multirow{3}{*}{Adversarial + Orthogonal constrain} & CASIA NIR-VIS 2.0 \cite{CasiaNIRVIS} &99.6& 99.7 & 99.4 \\   \cline{5-8}
    &&&& Oulu-CASIA NIR-VIS \cite{chen2009learning} & 100.0 & 98.9 & 93.9 \\  \cline{5-8}
          &&&& BUAA-VisNir face  \cite{huang2012buaa} & 99.9 & 99.8 & 99.4 \\ \hline \hline

    \multirow{3}{*}{2022} & \multirow{3}{*}{Luo \textit{et al.} \cite{luo2022memory}} & \multirow{3}{*}{MMTN} & \multirow{3}{*}{Adversarial + Identity + Content + Style + Cycle} & CASIA NIR-VIS 2.0 \cite{CasiaNIRVIS} &99.4& 99.8 & 99.7 \\   \cline{5-8}
    &&&& Oulu-CASIA NIR-VIS \cite{chen2009learning} & 100.0 & 98.2 & 89.0 \\  \cline{5-8}
          &&&& BUAA-VisNir face  \cite{huang2012buaa} & 99.2 & 99.5 & 97.2 \\ \hline \hline

    \multirow{3}{*}{2023} & \multirow{3}{*}{Yang \textit{et al.} \cite{yang2023robust}} & \multirow{3}{*}{RPC} & \multirow{3}{*}{Domain Adaptation} & CASIA NIR-VIS 2.0 \cite{CasiaNIRVIS} &99.7& 99.9 & 99.6 \\   \cline{5-8}
    &&&& Oulu-CASIA NIR-VIS \cite{chen2009learning} & 100.0 & 985 & 94.4 \\  \cline{5-8}
          &&&& BUAA-VisNir face  \cite{huang2012buaa} & 100.0 & 99.9 & 99.8 \\ \hline
    
    \end{tabular}
\end{table*}

\

\textbf{TRIVET}. Liu \textit{et al.}~\cite{Transferring_deep_rpz} proposed a CNN with ordinal measures (o-CNN), pre-trained on visible images from the large-scale CASIA WebFace dataset~\cite{Yi2014LearningFR} and hence is able to extract general face features, which has been fine-tuned on pairs of NIR-VIS face images, in order to learn a domain-invariant deep representation. To cope with limited image pairs, two types of NIR-VIS triplet loss functions were used, reducing intra-class variations by iteratively setting NIR and VIS images as anchors, such that the network focuses on the identity distinction instead of the spectrum classification.

\textbf{IDNet}. Reale \textit{et al.}~\cite{2016_NIR_HFR} utilized GoogLeNet with small convolutional filters, pre-trained on the visible CASIA WebFace dataset. For the NIR-to-VIS CFR scenario, they initialized two identical networks based on the pre-trained model, excluding the fully connected and softmax layers. The outputs from these two networks, termed VisNet and NIRNet, were concatenated to form a Siamese network with a contrastive loss function. This coupled deep network design effectively mapped NIR and VIS faces into a spectrum-independent feature space.


\textbf{IDR}. He \textit{et al.} \cite{IDR2017} presented a CNN-based approach targeted to map both NIR and VIS images into a common subspace by separating the feature space into a shared layer: NIR- and VIS-layer. While the shared layer encoded the modality-invariant identity information, the NIR- and VIS-layers encoded the modality-variant spectrum information, respectively. The NIR and VIS images were respectively input to the two CNN channels with shared parameters. Compared to similar CNN pipelines \cite{Transferring_deep_rpz,2016_NIR_HFR}, the proposed method jointly learned identity and spectrum information, leading to substantial performance gains on the CASIA NIR-VIS 2.0 dataset. 

\textbf{Hallucination}. Lezama \textit{et al.} \cite{Lezama_2017_CVPR} proposed an approach, which adapts a pre-trained VIS-CNN model towards generating discriminative features for both VIS and NIR face images, without retraining the network. Their approach consisted of two core components, cross-spectral hallucination and low-rank embedding, aiming at modifying both inputs and outputs of the CNN for CFR. Firstly, cross-spectral hallucination was used to transform the NIR image into the VIS spectrum by learning correspondences between NIR and VIS patches. Next, a low-rank embedding was used to restore a low-rank structure that simultaneously minimized the intra-class variations, while maximizing the inter-class variations. Three pre-trained VIS-CNN models, including \textit{VGG-S}, \textit{VGG-face} and \textit{COTS}, were evaluated on the CASIA NIR-VIS 2.0 dataset.


\textbf{WCNN}. He \textit{et al.}\cite{he2018wasserstein} proposed a novel method entitled \textit{Wasserstein Convolutional Neural Network} (WCNN) streamlined to learn invariant features between NIR and VIS face images. Similar to their previous work~\cite{IDR2017}, the proposed WCNN divided the high-level layer of the shared CNN into two orthogonal subspaces that contain modality-invariant identity features and modality-variant spectrum features. In addition, Wasserstein distance was used to measure the difference between heterogeneous feature distributions to reduce the modality gap. We note that their work represented the first attempt to formulate a probability-based distribution learning for VIS-to-NIR face comparison. WCNN outperformed the work of He \textit{et al.}~\cite{IDR2017}, improving the performance on the CASIA NIR-VIS 2.0 dataset.


\textbf{CDL}. Wu \textit{et al.}~\cite{CDL_AAAI_2018} designed a \textit{Coupled Deep Learning} (CDL) approach, identifying a shared feature space, in order to reduce modality differences in heterogeneous face comparison. The loss function of CDL consisted of two different components: (a) relevance constraints that imposed a trace norm to the softmax loss and a block-diagonal prior to the fully connected layer; and (b) cross-modal ranking that employed triplet ranking regularization to enlarge the size of training data. A semi-hard triplet selection method was also adopted to further improve the NIR-to-VIS comparison performance. Both relevance and ranking losses were jointly optimized. 


\textbf{CFC}. He \textit{et al.} \cite{He2019CrossspectralFC} proposed \textit{Cross-spectral Face Completion} (CFC), a GAN-based method, aimed at synthesizing visible spectrum (VIS) images from near-infrared (NIR) images. This heterogeneous face synthesis process comprises two complementary components: (a) texture inpainting, recovering VIS facial textures from NIR images, and (b) pose correction, mapping any pose in NIR images to a frontal pose in VIS images. These components are fused using a warping procedure within an adversarial network, regularized by UV loss, adversarial loss, and L1 reconstruction loss. The main contributions of this work include creating a novel GAN-based end-to-end framework for cross-spectral face synthesis without the need to assemble multiple image patches. The framework includes encoder-decoder structured generators and two novel discriminators to fully account for variations between NIR and VIS images. This approach simplifies the unsupervised heterogeneous face synthesis problem to a one-to-one image translation problem. The decomposition of texture inpainting and pose correction enables the generation of realistic, identity-preserving VIS face images.

\textbf{MMDL}. Cao \textit{et al.}~\cite{MMDL2019} combined heterogeneous representation network and decorrelation representation learning in order to design a \textit{Multi-Margin based Decorrelation Learning} (MMDL) framework to extract decorrelated features for both VIS and NIR images. First, the heterogeneous representation network was initialized from a VIS face model and fine-tuned on both VIS and NIR face images. Second, a decorrelation layer was appended to shared feature layer of the representation network to extract decorrelated features for both VIS and NIR images. 

\textbf{MC-CNN}. Deng \textit{et al.}~\cite{MCCNN2019} proposed a \textit{Mutual Component Convolutional Neural Network} (MC-CNN) to extract  modality-invariant features. The novelty of MC-CNN was the incorporation of a generative module, i.e., the mutual component analys (MCA), into the CNN by replacing the fully connected layer with MCA. VIS-NIR image pairs from the same subjects were first input to a shared CNN to extract common feature vectors. Subsequently, these feature vectors were fed into the MCA layer. Finally, MCA loss was included as an additional regularization to the softmax loss. 

\textbf{MDNDC}. Hu \textit{et al.} \cite{hu2019heterogeneous} proposed \textit{Multiple Deep Networks with scatter loss and Diversity Combination} (MDNDC). They used three ResNet-v1 networks; one being dedicated to feature extraction, in order to build the architecture of Multiple Deep Networks (MDN), followed by the two other networks in parallel. Jointly with the MDN, a Scatter loss (SL) \cite{hu2019discriminant} was used as a loss function, which attempts to maximize distance between the classes and minimize distance within the class to learn highly discriminative features for the CFR task.
In addition to the MDN, a joint decision strategy named Diversity Combination (DN) was introduced to auto-adjust weights of all three deep networks of the MDN and make a joint classification decision.

\textbf{RGM}. Cho \textit{et al.} \cite{cho2020relational} were interested in Relational Deep Feature Learning for CFR. Therefore, to bridge the domain gap between visible and invisible spectrum, they proposed a \textit{Relational Graph-structured Module} (RGM) focused on facial relational information. Their graph RGM performed relational modeling from node vectors that represent facial components such as lips, nose and chin. They also performed recalibration by considering global node correlation via \textit{Node Attention Unit} (NAU) to focus on the more informative nodes arising from the relation-based propagation. Furthermore, they suggested a novel conditional-margin loss function \textit{C-Softmax} for efficient projection learning on the common latent space of the embedding vector that adaptively uses the inter-class margin. 

\textbf{DSVN}. Hu \textit{et al.}~\cite{HuDSVN2020} proposed \textit{Disentangled Spectrum Variations Networks} (DSVN) to disentangle spectrum variations between VIS and NIR domains and to separate the modality-invariant identity information from modality-variant spectrum information. DSVN is comprised of two major components: spectrum-adversarial discriminative feature learning (SaDFL) and step-wise spectrum orthogonal decomposition (SSOD). The SaDFL was further divided into identity-discriminative subnetwork (IDNet) and auxiliary spectrum adversarial subnetwork (ASANet). The IDNet was used to extract identity discriminative features. The ASANet, on the other hand, was used to eliminate modality-variant spectrum information. Both IDNet and ASANet were designed to extract domain-invariant feature representations via adversarial learning. 

\textbf{ADCANs}. Hu \textit{et al.} \cite{ADCANs_2020} proposed an effective \textit{Adversarial Disentanglement spectrum variations and Cross-modality Attention Networks} (ADCANs) for the VIS-to-NIR CFR scenario. To reduce the gap of cross-modal images and solve the NIR-VIS CFR problem, the authors set-up three key components which are able to learn identity-related and modality-unrelated features. 
Firstly, they proposed a new objective loss termed \textit{Advanced Scatter Loss} (ASL), aiming at capturing within- and between-class information of the data and embedding them in the network for more effective training, focusing on categories with small inter-class distance and increasing the distance between them. 
Then, a \textit{Modality-adversarial Feature Learning} (MaFL) including an Identity-Discriminative Feature Learning Network (IDFLN) and a Modality-Adversarial Disentanglement Network (MADN) were incorporated to improve the feature representation of the identity-discriminative component as well as to highlight the spectrum variations through adversarial learning. 
Finally, a \textit{Cross-modality Attention Block} (CmAB) was introduced, in order to guide the network in selecting pertinent features and suppress noise information. CmAB sequentially applies spatial and channel attention modules to both, IDFLN and MADN, in order to increase the representation ability between them.

\begin{figure}
    \centering
    \includegraphics[width=0.80\textwidth]{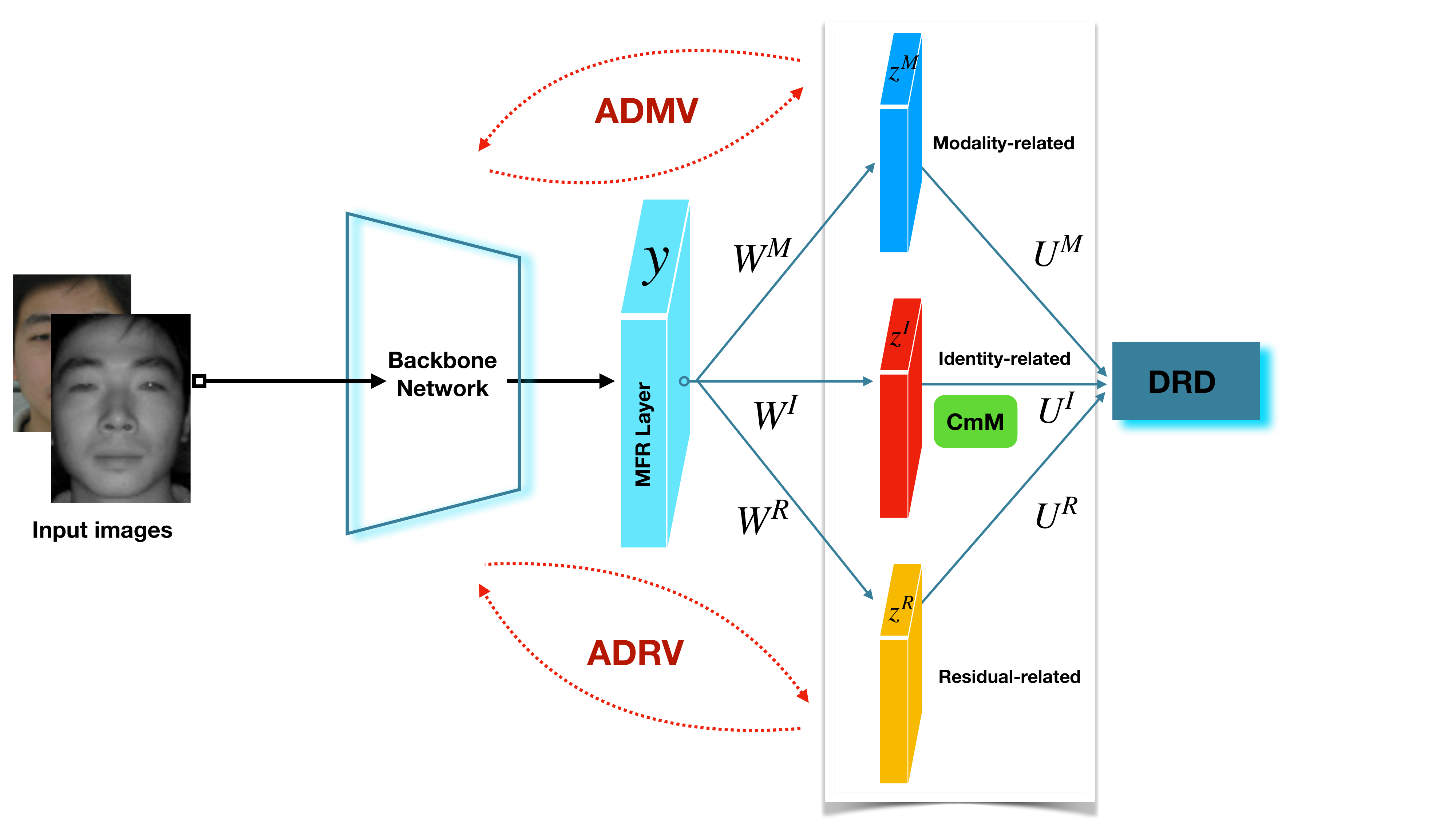}
    \caption{\textbf{Overview of the DADRD \cite{hu2020dual_DADRD} model:} learn identity-discriminative features and disentangle within-class variations. DADRD combines three key components, including \textit{CmM} loss, \textit{DADV} consisting of both \textit{ADMV} and \textit{ADRV} and \textit{DRD} embedded into the three previous \textit{MFR} layers.}
    \label{fig:DADRD_architecture}
\end{figure}

\textbf{DADRD}. Hu \textit{et al.} \cite{hu2020dual_DADRD} proposed the \textit{Dual Adversarial Disentanglement and deep Representation Decorrelation} (DADRD) approach to reduce the gap between NIR-VIS modalities, while enhancing the learning of identity-related features. At the same time, DADRD sought to effectively disentangles the additional residual-related features (i.e. PIE) rather than only extracting modality-related (i.e. NIR and VIS) and identity-related features. 
In contrast with their prior work, MDNDC \cite{hu2019heterogeneous} and ADCAN \cite{ADCANs_2020}, the DADRD method combined three key components. 
Firstly, they proposed a new objective loss termed \textit{Cross-modal Margin} (CmM), and attempted to enhance the learning identity-related features, which captures within- and between-class information of the data (i.e., occlusion, pose, distance, lighting and expressions), while also reducing the modality gap by using a center-variation item. 
Then, they introduced a \textit{Mixed Facial Representation} (MFR) which is divided into three layers: ``(I) Identity-related layer", ``(M) Modality-related layer" and ``(R) Residual-related layer".  
A \textit{Dual Adversarial Disentanglement Variation} (DADV) was then designed to reduce the intra-class variation with the help of adversarial learning, including both \textit{Adversarial Disentangled Modality Variations} (ADMV) and \textit{Adversarial Disentangled Residual Variations} (ADRV). These adversarial mechanisms disentangle the spectrum variation (in terms of data heterogeneity) and eliminate residual variations such as PIE, respectively.
Finally, a \textit{Deep Representation Decorrelation} (DRD) was embedded to the three MFR layers of DADV for the purpose of making them unrelated to each other and enhance feature representations.
The DADRD method, combining CmM, DADV and DRD into a unified framework, is depicted in Figure \ref{fig:DADRD_architecture}.

\textbf{DVG-Face.} Fu \textit{et al.}~\cite{DVGFACE2021} proposed a novel Dual Variational Generation (DVG-Face) framework to learn the joint distribution of paired heterogeneous images. Identity information from large-scale visible face data was also utilized to tackle the problem of identity diversity sampling induced by the small-scale paired heterogeneous data. To ensure identity consistency, a pairwise identity preserving loss was applied to generated paired heterogeneous images. A contrastive learning scheme was employed to train the heterogeneous face matching with positive pairs from generated paired heterogeneous images and negative pairs from different identity samplings. The proposed DVG-Face was an extension from the authors' previous work DVG~\cite{DVG2019} that greatly improved the identity diversity of generated face images.

\textbf{CAL.} He \textit{et al}.\cite{he2021coupled} introduced Coupled Adversarial Learning (CAL), a synthesis framework which offers a semi-supervised approach for learning features across spectra, specifically when dealing with a limited number of unpaired training data. Particularly, at the image level, a deep transformation network has been trained to convert NIR images to the VIS domain, addressing appearance discrepancies using carefully designed cycle loss, intensity loss, and texture loss.
At the feature level, CAL aimed at creating a deep feature space, where the cross-spectral face matching problem was approximated as a homogeneous face matching problem. Adversarial loss, along with orthogonal constraints, enhanced the ability of cross-spectral feature representation.

\textbf{MMTN.} Luo \textit{et al}. \cite{luo2022memory}
presented MMTN, an innovative Memory-Modulated Transformer Network. 
The approach followed the recognition via generation strategy, framing the problem as an unsupervised generation task with a "one-to-many" reference-based approach. To delve into prototypical style patterns from the reference domain, authors incorporated a memory module. In addition, a novel style transformer module was designed to facilitate the local fusion of input image content with the reference image's style.

\textbf{RPC.} Yang \textit{et al}. \cite{yang2023robust} introduced RPC, a domain adaptation model which integrated three key components into a unified framework. This integration aimed to uncover intrinsic relationships in cross-domain data and to facilitate the learning of domain-invariant representations. 
The first component, namely NPS, was employed to accurately estimate pseudo labels based on NIR clusters. These pseudo labels guided the network in learning representations that were invariant across domains. 
Subsequently, the DCL component was introduced to enhance the discriminate qualities of individual features within each specific domain. 
Finally, the ICL component was implemented to guide the network in learning robust and domain-independent representations.

\begin{center}
    \textbf{Summary of NIR-to-VIS Face Comparison}
\end{center}

Comparing face images in NIR against those in VIS can be coarsely classified into (a) cross-spectrum featured-based methods~\cite{Transferring_deep_rpz, 2016_NIR_HFR, IDR2017, he2018wasserstein, CDL_AAAI_2018, MMDL2019, MCCNN2019, cho2020relational, HuDSVN2020} and (b) cross-spectrum image-synthesis methods~\cite{Lezama_2017_CVPR, He2019CrossspectralFC}. Existing methods (a) predominantly seek to learn a domain-invariant deep representation between NIR and VIS samples, which is motivated by the presence of some similar textures observed in NIR as well as VIS-domains. Such approaches can incorporate two different ways to construct the inputs of the networks. Firstly, NIR and VIS images pertaining to same subjects are related in  positive pairs~\cite{2016_NIR_HFR, IDR2017,he2018wasserstein, MMDL2019, MCCNN2019, HuDSVN2020}, constituting two identical subnetworks with shared weights. A \textit{contrastive loss} in this context minimizes the intra-class distance
\begin{equation}\label{loss:ct}
\mathcal{L}_{ct}=
\begin{cases}
\frac{1}{2}\Vert f(x_i)-f(x_j)\Vert^2, &\text{if \(y_i=y_j\)} \\
\frac{1}{2} max(0,m-\Vert f(x_i)-f(x_j)\Vert )^2, &\text{else}.
\end{cases}
\end{equation}
Secondly, a triplet can be formed, where a NIR image is set as an anchor, a VIS image of the same subject as a positive sample, and an additional VIS image pertaining to a different subject as a negative sample. Alternatively, a triplet can be formed by setting a VIS image as an anchor and using NIR images from the same and different subjects as positive and negative samples, respectively~\cite{Transferring_deep_rpz, CDL_AAAI_2018}. This leads to three identical subnetworks with shared weights, where the \textit{triplet loss} is defined as
\begin{equation}\label{loss:tp}
\mathcal{L}_{tp}=\sum_{i}^{N}  \left( \Vert f(x_{i}^{a})-f(x_{i}^{p})\Vert - \Vert f(x_{i}^{a})-f(x_{i}^{n}) \Vert +m \right).
\end{equation}
Here, $x_{i}^{a}$ is an anchor image, $x_{i}^{p}$ is a positive sample and $x_{i}^{n}$ is a negative sample. The goal of triplet loss is to `push' the negative sample $x_{i}^{n}$ away from the anchor $x_{i}^{a}$ by a margin $m$ compared to the positive sample:
\begin{equation}
\Vert x_{i}^{a}-x_{i}^{p}\Vert^2+m \leq \Vert x_{i}^{a}-x_{i}^{n}\Vert.
\end{equation}
The triplet loss attempts to minimize the intra-class distance, while maximizing the inter-class distance. Unlike triplet, where a group of three samples is selected, the \textit{Tetrad Margin Loss} (TML)~\cite{MMDL2019} uses a group of four samples to construct tetrad tuples. The designed tetrad sample selection strategy was to choose four heterogeneous decorrelation representations, viz., $\{z_{j}^{N}, z_{j}^{V}, z_{k}^{N},z_{l}^{V}\}$. Herein, $\{z_{j}^{N}, z_{j}^{V}\}$ are samples from the same identity, $z_{k}^{N}$ denotes the closest NIR sample to $z_{j}^{V}$ from another identity, $z_{l}^{V}$ represents the closest VIS sample to $z_{j}^{N}$ from another identity. The tetrad margin loss is defined as:

\begin{equation}\label{loss:TML}
\begin{scriptsize}
\begin{aligned}
\mathcal{L}_{TML}(z_{j}^{N}, z_{j}^{V}, z_{l}^{V}) & =\sum_{i \in \{ N,V \} }^{T} \left(\frac{z_{j}^{N}\cdot z_{j}^{V}}{\Vert z_{j}^{N}\Vert \cdot \Vert z_{j}^{V} \Vert } - \frac{z_{j}^{N}\cdot z_{l}^{V}}{\Vert z_{j}^{N}\Vert \cdot \Vert z_{l}^{V} \Vert }+m_1\right) \\
 & +\sum_{ i \in \{ N,V \}  }^{T} \left(\frac{z_{j}^{N}\cdot z_{j}^{V}}{\Vert z_{j}^{N}\Vert \cdot \Vert z_{j}^{V} \Vert } - \frac{z_{j}^{V}\cdot z_{l}^{V}}{\Vert z_{j}^{V}\Vert \cdot \Vert z_{l}^{V} \Vert }+m_2\right).
\end{aligned}
\end{scriptsize}
\end{equation}

Subsequently, $\mathcal{L}_{TML}(z_{j}^{N}, z_{j}^{V},z_{k}^{N}, z_{l}^{V})$ can be defined as
\begin{equation}
\begin{split}
\mathcal{L}_{TML}(z_{j}^{N}, z_{j}^{V},z_{k}^{N}, z_{l}^{V})& = \mathcal{L}_{TML}(z_{j}^{N}, z_{j}^{V}, z_{l}^{V}) \\
& + \mathcal{L}_{TML}(z_{j}^{V}, z_{j}^{N}, z_{k}^{N}).
\end{split}
\end{equation}

$T$ represents the number of samples in a mini-batch, while $m_1$ and $m_2$ are the two margins. The tetrad margin loss can be regarded as the combination of two triplet losses~\cite{Transferring_deep_rpz, CDL_AAAI_2018}. Intuitively, $\mathcal{L}_{TML}(z_{j}^{N}, z_{j}^{V}, z_{l}^{V})$ is very similar to assigning a NIR image $z_{j}^{N}$ as an anchor, a VIS image $z_{j}^{V}$ of the same subject as a positive sample, and another VIS image $z_{l}^{V}$ of a different subject as a negative sample. The difference lies in the use of cosine distance other than the euclidean distance to compare the similarity. Both triplet loss and tetrad margin loss use the hard-sample mining strategy.  

In addition to the contrastive loss, triplet loss and tetrad margin loss, other loss functions have notably been proposed in the NIR-to-VIS literature. In~\cite{IDR2017}, a variant of \textit{softmax loss} was proposed in order to learn modality invariant subspace. The softmax loss function is defined as:
\begin{equation}
\begin{split}
\mathcal{L} & = \sum_{i\in \{N,V\}} \text{\ softmax}(F_i,c,\Theta, W, P_i)\\
& = -\sum_{i\in \{N,V\}} \left(\sum_{j=1}^{N} \text{log}(\hat{p}_{ij}) \ \mathbbm{1}_{\{y_{ij}=c\}}  \right), \\
& s.t. \quad P_i^{T}W=0, \quad i\in \{N,V\},
\end{split}
\end{equation}
where, $F_i$ is the fully connected layer of WCNN, $c$ is the class label and $\Theta$ is the set of WCNN parameters. $W$ is used to denote the weight matrix of the modality-invariant features (i.e., shared features across spectral domain) and $P_i$ denotes the weight matrix of the spectrum-specific features (i.e., features which are spectral dependent). $\mathbbm{1}$ is an indicator function and $p_{ij}$ denotes the predicted class probability. $P_i^{T}W$ corresponds to an orthogonal constraint imposed to make the features uncorrelated. 





\subsubsection{Shortwave Infrared}


To date, SWIR-to-visible has received limited attention. This is reflected in the scarcity of deep models and in the lack of publicly available SWIR face datasets (see Table \ref{tab:Datasets}).
We proceed to summarize existing prominent handcrafted methods \cite{BOURLAI201614, 6284307_bourlai}.
 

\textbf{pre-CNN.} Research of Bourlai \textit{et al.} \cite{5597717__Bourlai_SWIR} firstly investigated face verification related to the SWIR band. The authors introduced a \textit{geometric} and \textit{photometric normalization} (PN) scheme in order to compensate for the variable environment, coupled with \textit{contrast limited adaptive histogram equalization} (CLAHE). Based on that, three FR methods (commercial and academic) were tested using the \textit{West Virginia University} (WVU) Multi-spectral database.

Kalka \textit{et al.} \cite{6117586_Bourlai_SWIR} extended the previous work w.r.t. pre-processing \cite{5597717__Bourlai_SWIR} by adding \textit{single scale retinex} (SSR) and also explored the cross-photometric score level fusion rule. 
For the experiment, three datasets were considered including different environmental setup (fully controlled/semi controlled/uncontrolled). Their finding provided insights such as that pre-processing with PN improved FR performance, obtaining highest results up to Rank-1 accuracy of $100 \%$, when images were acquired under fully controlled conditions. 

Bourlai \textit{et al.} \cite{6284307_bourlai} pursued the intra-spectral and cross-spectral FR specially focused on (NIR/SWIR/MWIR)-imaging at various standoff distances and different environmental conditions. They merged previous work \cite{5597717__Bourlai_SWIR, 6117586_Bourlai_SWIR} in order to demonstrate that jointly, the use of independent or combined PN, CLAHE, SSR and cross-photometric score level fusion rule were able to reach better results under different environmental setups. 

\textbf{CMLD}. Cao \textit{et al.} \cite{SWIR_to_visible} proposed a local operator called \textit{Composite Multilobe Descriptors} (CMLD) with the aim to extract facial feature through all IR-spectral bands. This operator combined Gaussian function, Local binary patterns \textit{(LPB)}, Weber local descriptor and histogram of oriented gradients \textit{(HOG)}. The experimental results showed that for the SWIR-to-visible scenario, CMLD performed on a private dataset containing SWIR images (at various standoff distances), a Rank-1 accuracy of 78.7$\% $ and a verification rate of 99.5$\% $ at FAR 10$\% $.
Authors also used CMLD to conduct a study on cross-spectral \textit{partial} FR in order to understand the face area which contained useful information for CFR. They found that the eye region is most informative for FR beyond the visible.   

\textbf{VGGFace}. Bihn \textit{et al.} \cite{Bihn2018} explored the use of a pre-trained VGG-Face network on visible images to extract features from shortwave infrared (SWIR) images. They hypothesized that the VGG-Face network would perform well on both VIS and SWIR face images since their facial appearance differences are less significant compared to MWIR and LWIR images. SWIR wavelengths at 935 nm, 1060 nm, 1300 nm, and 1550 nm were evaluated, along with a composite image combining 1060 nm, 1300 nm, and 1550 nm. Deep features were extracted from the \textit{fc7} layer of the VGG-Face network, producing a 4096-dimensional feature vector. The Euclidean distance between feature representations of VIS and SWIR images was used for similarity comparison.

\textbf{GMLM-CNN}. Cao \textit{et al.} \cite{cao2022gmlm} proposed GMLM-CNN, a hybrid method leveraging traditional feature engineering with deep learning techniques for CFR. Repectively, two operators named Nominal Measurement Descriptor and Interval Measurement Descriptor (IMD) extract features at both, nominal and interval measurement levels. These features are then fused into a comprehensive representation using a Gabor Multiple-Level Measurement (GMLM) operator.
To further enhance feature selection and recognition performance, GMLM-CNN employed a PCA-based neural network that effectively filtered out redundant information and identified the most informative features.

\textbf{SSL}. Nanduri and Chellappa \cite{Nanduri_2024_WACV} used semi-supervised learning (SSL) and  explored the impact of different training datasets with limited data on the generalization capability of the trained network across various test domains, in particular SWIR vs. Visible.
Experiments demonstrated that larger models consistently outperformed smaller architectures, even with very small training sets, without compromising generalization performance. When dealing with multiple target domains and data scarcity, prioritizing training data collection for the most challenging domain (farthest from the source domain) showed to be more effective.
Wider training datasets (\textit{i.e.,} more classes with fewer samples per class) exhibited better performance compared to deep datasets (\textit{i.e.,} fewer classes with more samples per class). In the presence of noisy unlabeled data (containing samples from classes not represented in the labeled data), back-propagating entropy loss solely on the feature extractor (excluding the classifier) significantly enhanced performance.
Designing fine-tuning datasets that solely focus on increasing labeled data without incorporating novel variations (\textit{i.e.,} pose, illumination) was proven a not reliable strategy for consistently improving performance. Instead, authors stated that efforts should be directed towards incorporating data with diverse variations to effectively generalize the network's capabilities.

         
            
         

 \subsection{Emissive IR-to-Visible FR}\label{sec:Emissive_algo}

\subsubsection{Midwave Infrared}
\begin{figure}[h]
  \centering
    \includegraphics[width=0.8\textwidth]{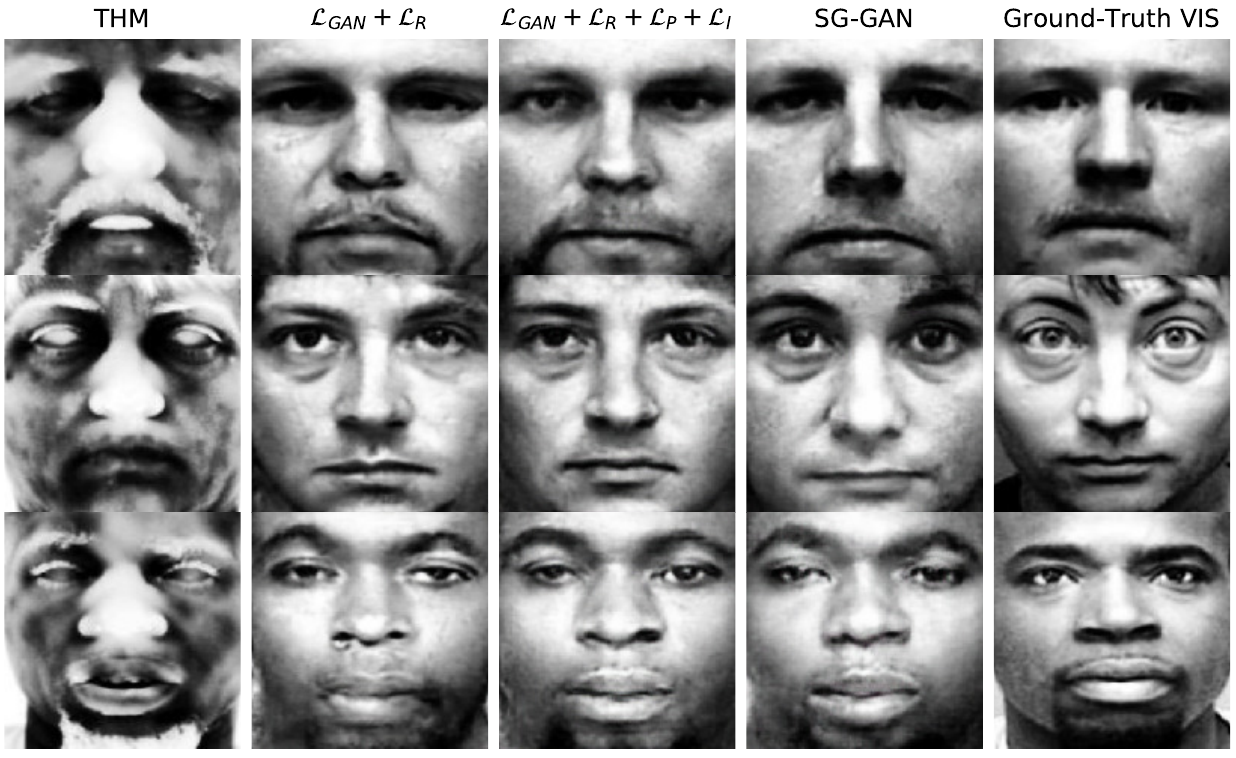}
    \caption{\textbf{Synthesizing VIS face images} from THM images on the PCSO dataset.  Compared to the use of the ``$\mathcal{L}_{GAN}+\mathcal{L}_R+\mathcal{L}_P+\mathcal{L}_I$'' loss function, the output of SG-GAN is semantically more close to the ground-truth VIS image especially around the salient facial regions.}
    \label{pcso_synthesized}
\end{figure}

Appearance difference between MWIR and VIS is more pronounced than NIR and SWIR (Figure \ref{fig:NIR_SWIR_MWIR_LWIR_face}). Therefore, learning a common feature subspace for MWIR-to-VIS is a challenging problem. In most cases, methods developed over the years have relied on GANs to synthesize visible face images from thermal face images. Illustrative examples are shown in Figure \ref{pcso_synthesized} and \ref{TR-GAN_tufts_synthesized}. 
We report the related MWIR-to-VIS face comparison performance in Table \ref{MWIR_VIS_comparisons}. 
 




         

\begin{table*}[]
      \caption{Rank-1 Accuracy and Verification Rate on different dataset for the \textbf{MWIR-VIS} face comparison.}
    \label{MWIR_VIS_comparisons} 
    \centering
    \begin{tabular}{|c|c|c||l|c|}
    \hline
        \textbf{Year} & \textbf{Authors}& \textbf{Methods} & \textbf{Dataset} &  \textbf{Performance $\% $} \\ \hline \hline
          
2017 & Sarfraz \textit{et al.} \cite{DPM_Sarfraz_Stiefelhagen} & DPM & NVESD \cite{byrd2013preview_NVESD}& 98.66 @Rank-1 \\ \hline \hline

2018 & Zhang \textit{et al.} \cite{TVGAN2018} & TV-GAN & IRIS \cite{133} & 19.90 @Rank-1   \\ \hline  \hline

2019 & Chen \textit{et al. \cite{ChenR19}} & SG-GAN & PCSO \cite{klare2012heterogeneous} & 92.16 @AUC - 15.01 @EER   \\   \hline \hline

\multirow{2}{*}{2020} &  \multirow{2}{*}{Iranmanesh \textit{et al.} \cite{CpGAN_US_Army}} & \multirow{2}{*}{CpGAN} & NVESD \cite{byrd2013preview_NVESD} & 96.10 @Rank-1 \\   \cline{4-5}
         
            &&& WSRI \cite{WSRI_dataset} & 97.80 @Rank-1 \\ \hline \hline 
            
\multirow{3}{*}{2021} &  \multirow{3}{*}{Peri \textit{et al.} \cite{peri2021synthesis_IEEE}} & Pix2Pix-ATC & MILAB-VTF(B) \cite{peri2021synthesis_IEEE}& 59.3 @AUC - 43.4 @EER \\   \cline{3-5}
         
            && CycleGAN-ATC & MILAB-VTF(B) \cite{peri2021synthesis_IEEE} & 54.9 @AUC - 46.4 @EER \\ \cline{3-5} 
            && CUT-ATC & MILAB-VTF(B) \cite{peri2021synthesis_IEEE} & 68.8 @AUC - 36.3 @EER \\ \hline
    
 \end{tabular}
\end{table*}

\

\textbf{DPM}. One of the first successful MWIR-to-VIS approaches with deep neural network was introduced by Sarfraz \textit{et al.} \cite{DPM_Sarfraz_Stiefelhagen}, termed \textit{Deep Perceptual Mapping} (DPM). Their approach was targeted to learn a non-linear mapping between visible and thermal domains (including MWIR and LWIR) while preserving the identity information. Specifically, a feed-forward neural network was constructed to regress densely computed features from the visible image to the corresponding thermal image. An MSE loss function was used to measure the perceptual difference between the visible and thermal images. Further, a regularization term with Frobenius norm of the projection matrix was also introduced. 


\textbf{TV-GAN}. Zhang \textit{et al.}~\cite{TVGAN2018} proposed a \textit{Thermal-to-Visible Generative Adversarial Network} (TV-GAN) that can synthesize visible images from their corresponding thermal images while maintaining identity information during the reconstruction. To preserve the identity information, a multi-task discriminator was designed. This discriminator served two different purposes. Firstly, one output of this discriminator was used to differentiate whether the generated samples were "real" or "fake". Secondly, the other output of this discriminator was used to perform identity classification in the context of supervised learning, given that the label identities are provided. Note that this identity classification was regarded as a closed-set FR. That means the identity associated with a given sample had already been seen in the training dataset. The proposed TV-GAN is similar to the auxiliary classifier GAN (or AC-GAN). Experiments were conducted on IRIS dataset \cite{133} and compared against three baselines approaches. TV-GAN against other GAN-methods are visualized in Figure \ref{TR-GAN_tufts_synthesized}.


\textbf{SG-GAN}. Chen \textit{et al.}~\cite{ChenR19} proposed the use of \textit{Semantic-Guided Generative Adversarial Network} (SG-GAN) to automatically synthesize visible face images from their thermal counterparts. Specifically, semantic labels, extracted by a face parsing network were used to compute a semantic loss function to regularize the adversarial network during training. These semantic cues denoted high-level facial component information associated with each pixel. Further, an identity extraction network was leveraged to generate multi-scale features to compute an identity loss function. To achieve photo-realistic results, a perceptual loss function was introduced during network training to ensure that the synthesized visible face was perceptually similar to the target visible face image.  Experiments involving \textit{PCSO} \cite{klare2012heterogeneous} face dataset showed that the proposed method achieved promising results in both face synthesis and CFR (see Figure \ref{pcso_synthesized}).

\begin{center}
    \textbf{Summary of MWIR-to-VIS Face Comparison}
\end{center}

MWIR-to-VIS CFR scenario has been scarsely studied in scientific literature. Partly this is the case due to lack of public benchmark datasets. Interestingly, early research on comparing MWIR-to-VIS face images adopted a neural network~\cite{DPM_Sarfraz_Stiefelhagen}, formulated by sequential operations of fully connected layer and non-activation layer. This simple feed-forward neural network does not involve the use of convolution units and can be represented as
\begin{equation}
H(x)=h^{N} = g(W^{N}h^{N-1}+b^N),
\end{equation}
where, $W$ is the projection matrix, $h$ is the hidden layer and $b$ is the bias. Basically, the output of current hidden layer is a multiplication of project matrix with the output of previous hidden layer. $g$ is a non-activation function to ensure that the mapping is non-linear. The objective function of DPM was formulated as
\begin{equation}
\arg\min_{W,b} \mathcal{L} =\frac{1}{M} \sum_{i=1}^{M} (\bar{x}_i-t_i)^{2}+\frac{\lambda}{N}\sum_{i=1}^{N}\left( \Vert W^{k}\Vert_{F}^{2}+\Vert b^{k}\Vert_{2}^{2}\right). 
\end{equation}
Here, the first term is the mean square error between the feature vectors of visible $\bar{x}_i$ and thermal $t_i$ image. The second term serves as a regularizer on the weight matrix $W$ and bias $b$. Although the DPM showed promising results on comparing SWIR against VIS images, the performance was still far from satisfactory. 

Recent developments of neural networks, specifically in GANs, have pushed the frontiers of SWIR-to-VIS CFR scenario. Both TV-GAN~\cite{TVGAN2018} and SG-GAN~\cite{ChenR19} are inspired by Pix2Pix~\cite{pix2pix2017}, which involves the use of adversarial loss and $\mathcal{L}_1$ loss to regularize the image synthesis process. The $\mathcal{L}_1$ loss was used to measure the per-pixel difference between synthesized visible face image and target visible face image. Compared to Pix2Pix, TV-GAN modified the output of the discriminator to perform a closed-set identification. Therefore, a new identity loss was introduced in TV-GAN. SG-GAN, on the other hand, adds more loss functions to the Pix2Pix, including perceptual loss, identity loss and semantic loss. An ablation study was used to demonstrate the individual
functionalities of the loss functions (see Figure~\ref{pcso_synthesized}). 

\subsubsection{Longwave Infrared}
\begin{figure*}[h!]
    \centering
    \includegraphics[width=1\textwidth]{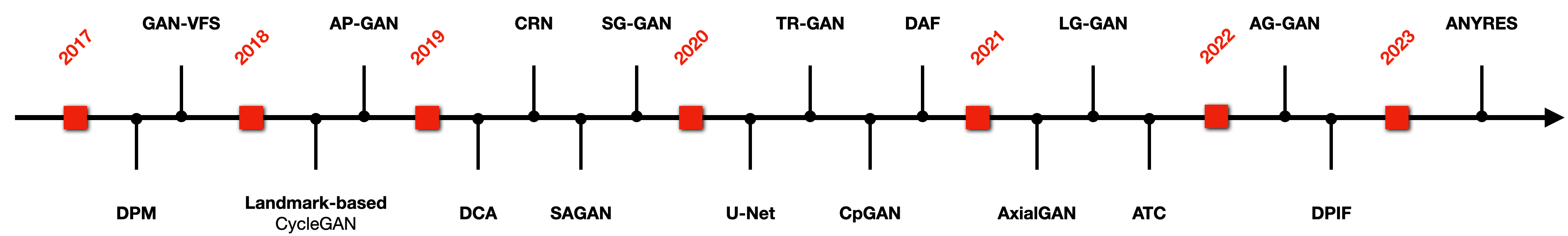}
    \caption{\textbf{Timeline of developments in algorithms:} LWIR-to-VIS face recognition.}
    \label{fig:Timeline_LWIR}
\end{figure*}

Face images captured under MWIR and LWIR tend to look very similar (Figure \ref{fig:NIR_SWIR_MWIR_LWIR_face}). Therefore, algorithms developed to address LWIR-to-VIS closely resembles that of MWIR-to-VIS. Most developed algorithms for LWIR-to-VIS CFR scenario were based on GANs. Here, we give a brief summary of recent representative works for LWIR-to-VIS comparison, related performance are reported in Table \ref{LWIR_VIS_comparisons} and we illustrate the timeline of developments in Figure \ref{fig:Timeline_LWIR}. 

\begin{table*}[t]
    \caption{Rank-1 Accuracy and Verification Rate on different dataset for the \textbf{LWIR-VIS} face Comparison.}
    \label{LWIR_VIS_comparisons}
    \centering
    \scriptsize
    \begin{tabular}{|c|c|p{0.12\linewidth}||l|c|}
    \hline
        \textbf{Year} & \textbf{Authors}& \textbf{Methods} & \textbf{Dataset} & \textbf{Performance $\% $} \\ \hline \hline

\multirow{6}{*}{2017} 
& \multirow{4}{*}{Sarfraz \textit{et al.} \cite{DPM_Sarfraz_Stiefelhagen}} & \multirow{4}{*}{DPM} &  UND X1 \cite{135} & 83.73 @Rank-1 \\ \cline{4-5}
&&& Carl \cite{CarlDatabase}  & 71.00 @Rank-1 \\ \cline{4-5}
&&& NVESD \cite{byrd2013preview_NVESD}  & 97.33 @Rank-1 \\\cline{4-5}
&&& Polarimetric thermal \cite{Polarimetric_Thermal_Face}-\textit{ext.} - \textit{Therm. Probe}  & 84.00 @Rank-1 \\\cline{2-5}

& \multirow{2}{*}{Zhang \textit{et al.} \cite{zhang2017generative_GAN_VFS}} & \multirow{2}{*}{GAN-VFS} & Polarimetric thermal \cite{Polarimetric_Thermal_Face} - \textit{Therm. Probe} & 85.61 @Rank-1 - 79.30 @AUC - 27.34 @EER \\ \cline{4-5}
&&&  ARL-VTF \cite{Poster_2021_WACV_ARL_VTF}  & 97.94 @AUC - 08.14 @EER \\ \hline \hline 

\multirow{4}{*}{2018} 
& \multirow{2}{*}{Wang \textit{et al.} \cite{CycleGANDetector2018}} & CycleGAN \cite{zhu2017unpaired} & private database & 88.90 @Rank-1 - 98.80 @Rank-3 - 99.80 @Rank-5   \\ \cline{3-5}
 && CycleGAN-Based & private database & 91.60 @Rank-1 - 99.30 @Rank-3 - 99.90 @Rank-5 \\ \cline{2-5}

& Iranmanesh \textit{et al.} \cite{iranmanesh2018deep} & CpDCNN & Polarimetric thermal \cite{Polarimetric_Thermal_Face} - \textit{Therm. Probe} & 88.57 @Rank-1
\\ \cline{2-5}

& Di \textit{et al.} \cite{di2018apgan} & AP-GAN & Polarimetric thermal \cite{Polarimetric_Thermal_Face} - \textit{Therm. Probe} & 84.16 @AUC - 23.90 @EER  \\ \hline \hline

\multirow{8}{*}{2019} 
& \multirow{3}{*}{Kantarcı \textit{et al.} \cite{ThermalToVisible_autoencoders}} & \multirow{3}{*}{DCA}             & UND X1 \cite{135} & 87.20 @Rank-1  \\ \cline{4-5}
            &&& EURECOM \cite{8553431EURECOM} & 88.33 @Rank-1 \\ \cline{4-5}
            &&&  Carl \cite{CarlDatabase}  & 85.00 @Rank-1 \\ \cline{2-5}

& Mallat \textit{et al.} \cite{CRN2019} & CRN & EURECOM \cite{8553431EURECOM} & 82.00 @LightCNN  \\ \cline{2-5}
& Iranmanesh \textit{et al.} \cite{iranmanesh2019attributeguided_AGC-GAN} & AGC-GAN & Polarimetric thermal \cite{Polarimetric_Thermal_Face} - \textit{Therm. Probe} & 89.25 @Rank-1 \\ \cline{2-5}
& \multirow{2}{*}{Di \textit{et al.} \cite{DiSAGAN2019}} & \multirow{2}{*}{SAGAN} & Polarimetric thermal \cite{Polarimetric_Thermal_Face} - \textit{Therm. Probe} & 91.49 @AUC - 15.45 @EER \\ \cline{4-5}
&&&  ARL-VTF \cite{Poster_2021_WACV_ARL_VTF}  & 99.28 @AUC - 03.97 @EER \\ \cline{2-5}

& Chen \textit{et al.} \cite{ChenR19} & SG-GAN & Polarimetric thermal \cite{Polarimetric_Thermal_Face} - \textit{Therm. Probe} & 93.08 @AUC - 14.24 @EER \\ \hline \hline

\multirow{8}{*}{2020} 
& Chatterjee \textit{et al.} \cite{UNET2020} & U-Net & Nagoya University & 69.60 @Rank-1  \\ \cline{2-5} 
& Kezebou \textit{et al.}~\cite{TRGAN2020} & TR-GAN & TUFTS \cite{Tufts} & 80.70 @Resnet50 - 88.65 @VGG16 \\ \cline{2-5}
&  \multirow{3}{*}{Iranmanesh \textit{et al.} \cite{CpGAN_US_Army}} & \multirow{3}{*}{CpGAN} & UND X1 \cite{135} & 76.40 @Rank-1 \\   \cline{4-5}
         
            &&& NVESD \cite{byrd2013preview_NVESD} & 93.90 @Rank-1 \\ \cline{4-5}
            &&&  Polarimetric thermal \cite{Polarimetric_Thermal_Face} - \textit{Therm. Probe} & 89.05 @Rank-1 \\ \cline{2-5}
            
& Di \textit{et al.} \cite{DiAttribute2020} & Multi-AP-GAN & Polarimetric thermal \cite{Polarimetric_Thermal_Face} - \textit{Therm. Probe} & 90.99 @AUC - 17.81 @EER \\  \cline{2-5}

& \multirow{2}{*}{Fondje \textit{et al.} \cite{fondje2020cross}} & \multirow{2}{*}{DAF} & Polarimetric thermal \cite{Polarimetric_Thermal_Face}-\textit{ext.} - \textit{Therm. Probe} & 94.20 @Rank-1 \\  \cline{4-5}
&&&  ARL-VTF \cite{Poster_2021_WACV_ARL_VTF}  & 99.76 @AUC - 2.30 @EER \\ \hline \hline 

\multirow{7}{*}{2021} 
& Immidisetti \textit{et al.} \cite{Immidisetti2021} & Axial-GAN & ARL-VTF \cite{Poster_2021_WACV_ARL_VTF} & 94.4 @AUC - 12.38 @EER  \\ \cline{2-5} 
& \multirow{2}{*}{Anghelone \textit{et al.}~\cite{AnghChen_LGGAN_FG2021}} & \multirow{2}{*}{LG-GAN} & Polarimetric thermal \cite{Polarimetric_Thermal_Face} - \textit{Therm. Probe} & 93.99 @AUC - 13.02 @EER\\  \cline{4-5}
&&& ARL-VTF \cite{Poster_2021_WACV_ARL_VTF} & 96.96 @AUC - 5.94 @EER\\ \cline{2-5}
& \multirow{4}{*}{Peri \textit{et al.} \cite{peri2021synthesis_IEEE}} & Pix2Pix-ATC & ARL-VTF \cite{Poster_2021_WACV_ARL_VTF} & 91.3 @AUC - 16 @EER\\ \cline{3-5}
&& CycleGAN-ATC & ARL-VTF \cite{Poster_2021_WACV_ARL_VTF}  & 96.8 @AUC - 9.2 @EER\\ \cline{3-5}

&& \multirow{2}{*}{CUT-ATC} & ARL-VTF \cite{Poster_2021_WACV_ARL_VTF}  & 97.7 @AUC - 6.9 @EER\\ \cline{4-5}
&&& TUFTS \cite{Tufts} & 87.4 @AUC - 21.3 @EER \\ \hline \hline  

\multirow{7}{*}{2022}
& \multirow{4}{*}{Chen \textit{et al.}~\cite{10008000}} & \multirow{2}{*}{AG-GAN} & SF dataset \cite{abdrakhmanova2021speakingfaces} & 89.86 @AUC - 17.68 @EER\\  \cline{4-5}
&&& ARL-VTF \cite{Poster_2021_WACV_ARL_VTF} & 98.74 @AUC - 5.56 @EER\\ \cline{3-5}

&& \multirow{2}{*}{AG-GAN+} & SF dataset \cite{abdrakhmanova2021speakingfaces} & 90.53 @AUC - 17.20 @EER\\  \cline{4-5}
&&& ARL-VTF \cite{Poster_2021_WACV_ARL_VTF} & 99.26 @AUC - 4.30 @EER\\ \cline{2-5}

& \multirow{3}{*}{Fondje \textit{et al.} \cite{fondje2022learning}} & \multirow{3}{*}{DPIF} & TUFTS \cite{Tufts} & 91.82 @AUC - 16.03 @EER\\  \cline{4-5}
&&& Polarimetric thermal \cite{Polarimetric_Thermal_Face}-\textit{ext.} & 93.06 @AUC - 14.97 @EER\\ \cline{4-5}
&&& ARL-VTF \cite{Poster_2021_WACV_ARL_VTF} & 99.99 @AUC - 0.15 @EER\\ \hline \hline

\multirow{4}{*}{2023}
& \multirow{4}{*}{Anghelone \textit{et al.}~\cite{anghelone2023ANYRES}} & \multirow{4}{*}{ANYRES} & EURECOM \cite{8553431EURECOM} & 93.65 @AUC - 14.05 @EER\\  \cline{4-5}
&&& TUFTS \cite{Tufts} & 83.09 @AUC - 24.53 @EER\\  \cline{4-5}
&&& SF dataset \cite{abdrakhmanova2021speakingfaces} & 91.39 @AUC - 15.87 @EER\\ \cline{4-5}
&&& ARL-VTF \cite{Poster_2021_WACV_ARL_VTF} & 99.88 @AUC - 1.26 @EER\\ \hline 

\end{tabular}
    \vspace{-0.40 cm}
\end{table*}


\

\textbf{Landmark-based CycleGAN}. Wang \textit{et al.}~\cite{CycleGANDetector2018} proposed a detection network to extract facial landmarks from visible faces and use that to guide the generative network to preserve geometric shapes. Their network was based on \textit{CycleGAN} \cite{zhu2017unpaired}, targeted to perform translation between the thermal and visible face images. The detection network extracted $68$-landmarks from visible faces, constructing the shape loss function. To perform the experiment, they established a new database (available online), including $33$ subjects with $792$ thermal-visible pair images. 

 
\textbf{DCA}. Kantarcı \textit{et al.} \cite{ThermalToVisible_autoencoders} introduced a \textit{Deep Convolutional Autoencoder} (DCA) based on the \textit{U-Net} \cite{RonnebergerFB15} architecture, incorporating Up Convolution and Difference of Gaussians (DoG) to learn the non-linear mapping between visible and thermal face images for CFR. They demonstrated that preprocessing and alignment improved performance by bridging the gap between these domains. The approach was tested on three publicly available thermal-visible datasets: Carl \cite{CarlDatabase}, UND-X1 \cite{135}, and EURECOM \cite{8553431EURECOM}.
Additionally, the authors manually annotated six facial landmarks on the Carl and EURECOM datasets to align the faces, which increased performance by approximately 2$\%$. They conducted experiments with three setups: the first two assessed the decoder method without applying the DoG filter as a preprocessing step, while the third tested the DoG filter's effect on the Up Convolutional Autoencoder Model proposed in this work.

\textbf{CRN}. Mallat \textit{et al.} \cite{CRN2019} proposed a solution based on \textit{Cascaded Refinement Network} (CRN). Their approach did not require a large amount of training data, owing to a limited number of training parameters. CRN is a type of CNN that consists of inter-connected refinement modules, which gradually process the image synthesis from lowest resolution (4$\times$4) to the highest resolution (128$\times$128). The refine module includes three different layers, namely input, intermediate and output layers. A contextual loss function was used to train the CRN. The motivation to choose the contextual loss was based on two conditions: (a) robustness to roughly aligned THM-VIS pairs; and (b) invariant to outlier at the pixel level in the context of per-pixel loss. FR experiments on the EURECOM \cite{8553431EURECOM} dataset showed that the proposed method achieves better performance than TV-GAN. In addition, they performed CFR using their synthesized faces (facial variation = Neutral) with two systems, namely \textit{OpenFace} and \textit{LightCNN}. 


\textbf{U-Net}. Chatterjee and Chu~\cite{UNET2020} presented a U-Net architecture with a residual network backbone to generate visible face images from thermal face images. The U-Net was modified by using residual blocks with skip-connections instead of the normal convolutional layers as the basic building components. Further, a pixel shuffle upsampling was introduced to replace the transposed convolution layers in the decoder part. A weighted combinations of two different loss functions was used to train the proposed network, namely mean squared loss and perceptual loss. The proposed method was evaluated on thermal face dataset from \textit{Nagoya University}, which contained $900$ thermal images and $900$ visible images captured simultaneously. A Rank-1 accuracy of 69.60\% was achieved.

\begin{figure}[h]
  \centering
    \includegraphics[width=1\textwidth]{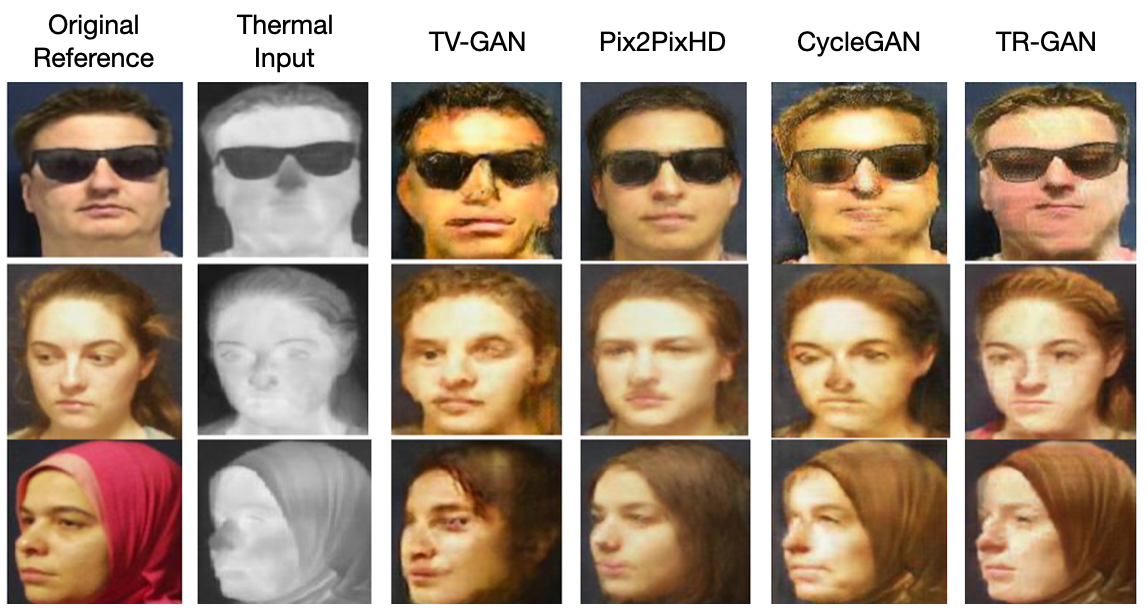}
    \caption{\textbf{Synthesizing VIS face images} from thermal images on the TUFTS dataset \cite{Tufts} with different visualization results: TR-GAN \cite{TRGAN2020}, CycleGAN \cite{ZhangRHSP19} and TV-GAN \cite{TVGAN2018}. Figure credit: Kezebou \textit{et al.} \cite{TRGAN2020}}
    \label{TR-GAN_tufts_synthesized}
\end{figure}

\textbf{TR-GAN}. Kezebou \textit{et al.}~\cite{TRGAN2020} proposed a \textit{Thermal to RGB Generative Adversarial Network} (TR-GAN) to automatically synthesize visible face images captured in the thermal domain. The TR-GAN employed an architecture similar to U-Net with cascade residual blocks for the generator. Basically, it replaced the Resnet blocks in the CycleGAN with cascaded-in-
cascaded residual blocks. This ensured that the generator synthesizes images with consistent global and local structural information. For the discriminator, the TR-GAN used the same discriminator network architecture as CyleGAN. A pretrained  VGG-Face recognition model was used to perform the face comparison after the thermal to visible image translation. The experiments were conducted on TUFTS face dataset and compared against TV-GAN, Pix2PixHD and CycleGAN. Figure \ref{TR-GAN_tufts_synthesized} shows samples of synthesized images performed by named models. In comparison, TR-GAN demonstrated superior performance in generating realistic images. 

\textbf{DAF.} Fondje \textit{et al.} \cite{fondje2020cross} proposed a \textit{Domain Adaptation Framework} (DAF) featuring a new feature mapping sub-network with: (a) a \textit{cross-domain identification} loss for effective "co-registration" and "synchronization," and (b) a \textit{domain invariance} loss for cross-domain regularization. Feature representations within each domain were extracted using a truncated pre-trained CNN (VGG16 or ResNet50), aiming to preserve discriminative information across visible and thermal spectra.
DAF was embedded in the \textit{Residual Spectral Transform} (RST), a residual block that transforms features with three specific 1x1 convolutions to enhance discriminability. Experiments using three datasets (polarimetric thermal, extended polarimetric thermal with 111 subjects, and a private dataset with 126 paired visible/thermal faces) showed significant improvement compared to DPM \cite{DPM_Sarfraz_Stiefelhagen}.

\textbf{Axial-GAN.} Immidisetti et al.~\cite{Immidisetti2021} proposed Axial-Generative Adversarial Network (Axial-GAN) to synthesize high-resolution visible images from low-resolution thermal counterparts. The proposed GAN framework designed an axial-attention layer with transformer to model long-range dependencies to facilitate long-distance face matching. Their work can simultaneously address face hallucination and translation for thermal-to-visible face matching. Evaluations on ARL-VTF thermal and multi-modal polarimetric thermal face recognition datasets obtained promising performances compared to SAGAN and HiFaceGAN. 

\begin{figure}
    \centering
    \includegraphics[width=0.75\textwidth]{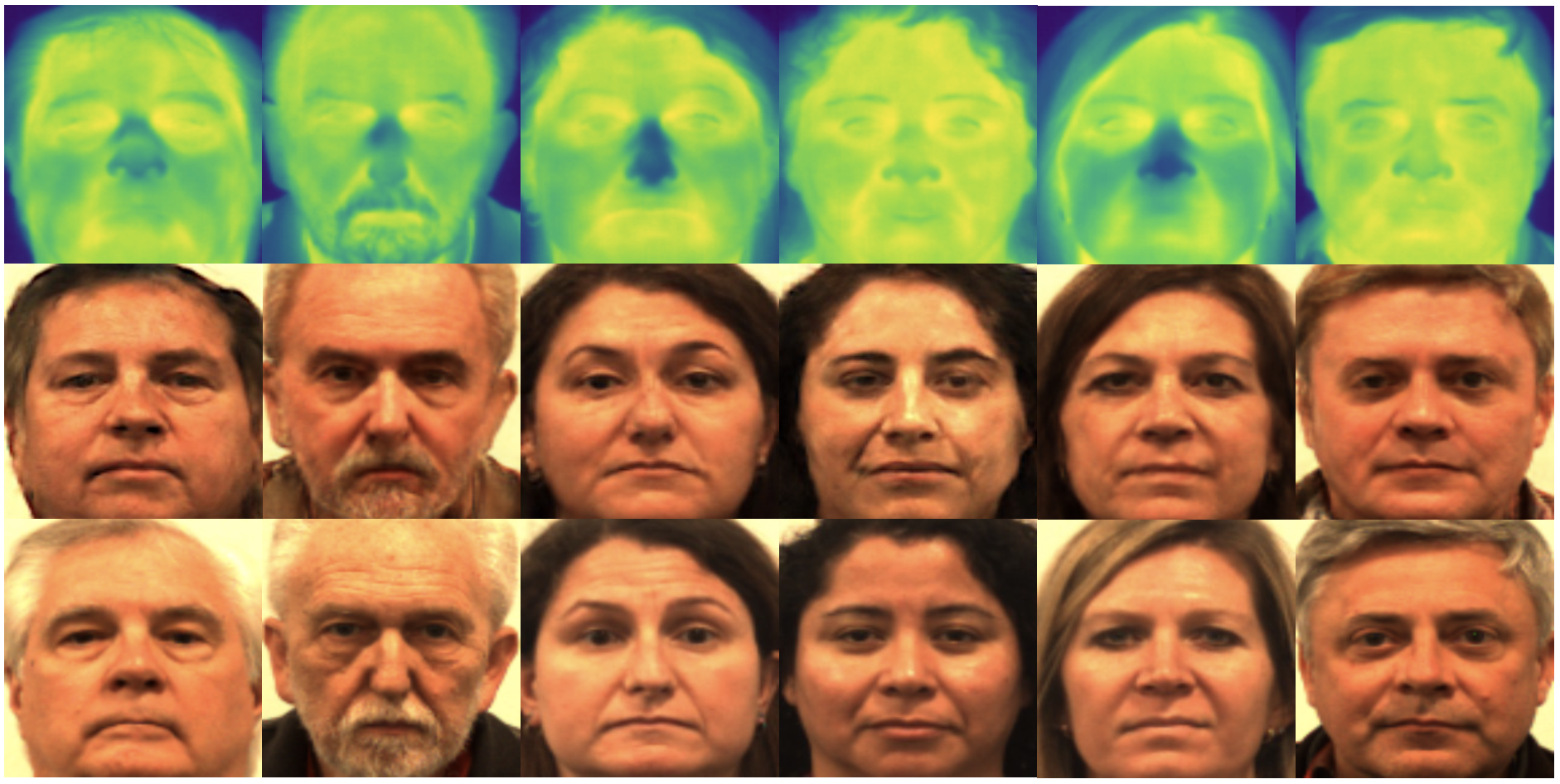}
    \caption{\textbf{Synthesizing VIS face images} from thermal images on ARL-VTF dataset \cite{Poster_2021_WACV_ARL_VTF} using LG-GAN method. Top row is the thermal input, middle row is the LG-GAN generation and bottom row is the ground truth image.}
    \label{fig:LG-GAN_samples}
\end{figure}

\textbf{LG-GAN.} Anghelone \textit{et al.}~\cite{AnghChen_LGGAN_FG2021} proposed a Latent-Guided Generative Adversarial Network (LG-GAN) which disentangle the identity from other confounding factors. Hence, an input face image is explicitly decomposed into an \textit{identity latent code} that is spectral-invariant and a \textit{style latent code} that is spectral-dependent. Thermal-to-Visible translation is performed by switching the thermal style code with the opposite visible style code and recombined with the identity code. Interpretation of the identity code offers useful insights in explaining salient facial structures that are essential to generate high-fidelity face images. By using such disentanglement, LG-GAN is able to preserve the identity during the spectral transformation and thus achieve promising face recognition results with respect to visual quality (Figure \ref{fig:LG-GAN_samples}).

\textbf{CUT-ATC.} Peri \textit{et al.} \cite{peri2021synthesis_IEEE} presented a comprehensive study on synthesis-based approach for thermal-to-visible face verification. They explored Pix2Pix, CycleGAN and Contrastive Unpaired Translation (CUT) off-the-shelf domain adaptation algorithms for the task of generating realistic synthetic samples. They further demonstrated that additional custom loss such as Pixel wise correspondence loss and identity loss are instrumental for addressing the thermal-to-visible face verification. Authors highlighted the impact of the face alignment and encompass their method named ATC with above loss functions in a end-to-end thermal-to-visible system comprising ATC: Alignment, spectrum Translation and Classification steps.

\textbf{AG-GAN.} Chen \textit{et al.} \cite{10008000} offered critical insights with an explainable GAN entitled Attention-Guided Generative Adversarial Network (AG-GAN). 
Towards enhancing interpretability and investigating specific facial features learned (across spectra), AG-GAN aims at \textit{finding representations} that allow for \textit{extracting} robust biometric features beyond the visible, as well as interpreting the generation process from thermal to visible. 
While AG-GAN is designed to encode thermal face images into attention maps learnt with supervised attention weights from an auxiliary-domain classifier, an alternative AG-GAN+ version uses \textit{Squeeze and Excitation} to generate in an unsupervised manner attention weights serving to build the attention map.

\textbf{DPIF.} Fondje \textit{et al.} \cite{fondje2022learning} introduced an innovative framework known as the Domain and Pose Invariance Framework (DPIF) for matching off-pose thermal images to frontal visible faces. 
This framework relies on modified base architectures for extracting image representations. Additionally, a new sub-network is introduced to simultaneously learn \textit{pose} and \textit{domain} invariance using a novel joint-loss function. This function combined the proposed \textit{pose-correction} and \textit{cross-spectrum} losses. 
Notably, the authors developed a compact and efficient domain adaptation framework. It is worth highlighting that the DPIF framework stands out from previous image-to-image translation approaches by not requiring {\em a priori} information related to the head pose of probe imagery.

\begin{figure}[h!]
    \centering
    \includegraphics[width=1\textwidth]{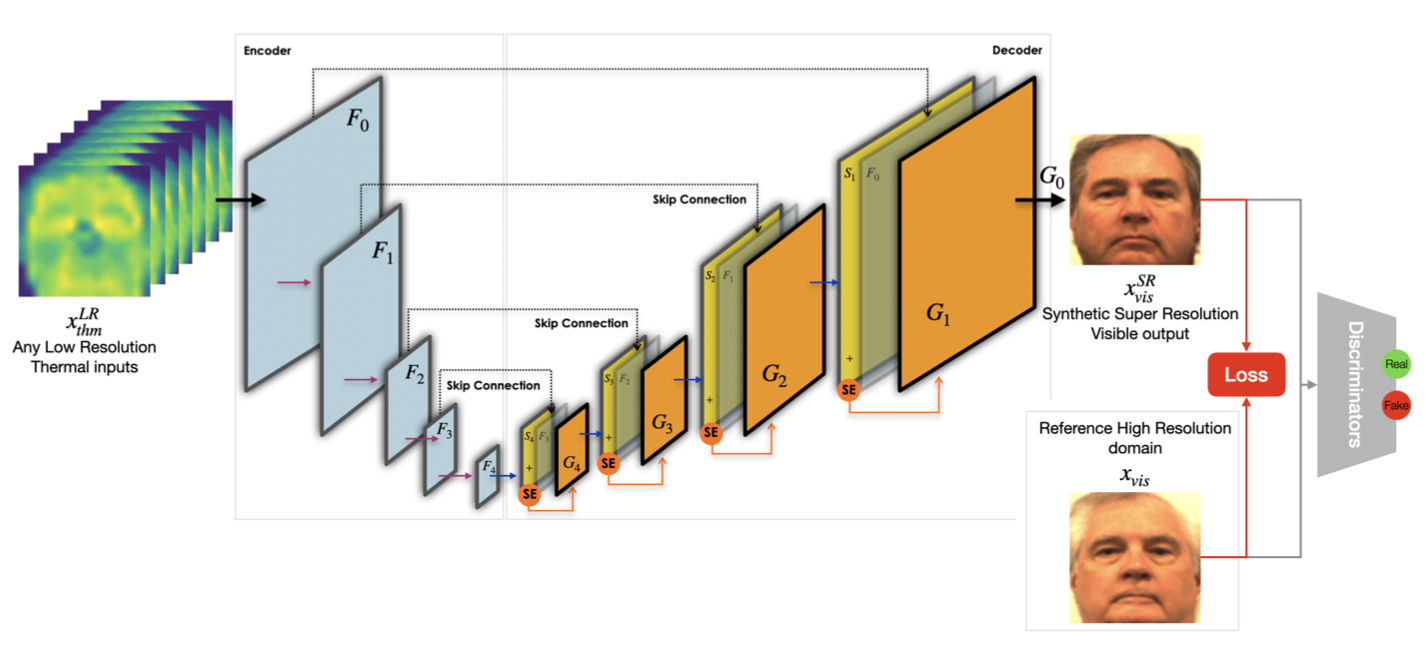}
    \caption{\textbf{Training of ANYRES \cite{anghelone2023ANYRES}.} The generator accepts any (low)-resolution thermal face $x^{LR}_{thm}$ as input. It comprises an encoder-decoder bridged by skip connections and gated by Squeeze and Excitation (SE) blocks, which play the role of gate modulator and enable resolution-wise relationships towards bringing a flexible control for balancing encoded features with decoded super resolved features. The discriminators are aimed at distinguishing real images $x_{vis}$ from generated synthetic ones $x_{vis}^{SR}$.}
    \label{fig:ANYRES_archi}
\end{figure}

\textbf{ANYRES.} Anghelone \textit{et al.} \cite{anghelone2023ANYRES} utilized a pyramidal UNet-based architecture (as related in Figure \ref{fig:ANYRES_archi}) to translate any (low) resolution thermal face images into high resolution visible face images, placing emphasis on maintaining the cross-spectral identity of depicted individual over any resolution. Authors proposed an algorithm suitable to real world operational scenarios, in which humans are situated at arbitrary distances from the camera, resulting in multi-scale LR thermal face images. 
Deviating from the state-of-the-art, where resolution is generally fixed with respect to the input, ANYRES is able to operate at any input resolution ranging from low to high.

\begin{center}
    \textbf{Summary of LWIR-to-VIS Face Comparison}
\end{center}

LWIR-to-VIS CFR scenario resembles MWIR-to-VIS CFR scenario as face images captured in LWIR and MWIR spectral bands are almost indistinguishable. In the previous works of MWIR-to-VIS comparison, algorithms developed based on Pix2Pix have been used to address the challenge~\cite{TVGAN2018,ChenR19}. The Pix2Pix model was designed only to learn forward mapping from one domain to another, while CycleGAN was streamlined to learn both the forward and inverse mappings simultaneously using cycle-consistency loss. In the work of~\cite{CycleGANDetector2018}, CycleGAN was used to learn such bi-directional mappings for LWIR and VIS face images. The objective function of CycleGAN is defined as
\begin{equation}
\begin{split}
\mathcal{L}(G,F,D_X,D_Y) &=\mathcal{L}_{GAN}(G,D_Y,X,Y) \\
& + \mathcal{L}_{GAN}(F,D_X,Y,X) \\
& + \lambda \mathcal{L}_{cyc}(G,F).
\end{split}    
\end{equation}
Here, $G$ is used to learn the forward mapping from $X$ to $Y$, i.e., $G:\{X \to Y\}$. The corresponding discriminator is $D_Y$. Therefore, $\mathcal{L}_{GAN}(G,D_Y,X,Y)$ can be described by
\begin{equation}
\begin{split}
\mathcal{L}_{GAN}(G,D_Y,X,Y) & = \mathbb{E}_{y\sim p_{data}(y)}[log(D_Y(y))] \\
& + \mathbb{E}_{x\sim p_{data}(x)}[log(1-D_Y(G(x)))].
\end{split}    
\end{equation}
Similarly, $F$ is used to learn the inverse mapping from $Y$ to $X$, i.e., $F:\{Y \to X\}$. The corresponding discriminator is $D_X$. $\mathcal{L}_{GAN}(F,D_X,Y,X)$ can be described by
\begin{equation}
\begin{split}
\mathcal{L}_{GAN}(F,D_X,Y,X) & = \mathbb{E}_{x\sim p_{data}(x)}[log(D_X(x))] \\
& + \mathbb{E}_{y\sim p_{data}(y)}[log(1-D_X(G(y)))].
\end{split}    
\end{equation}
In addition to adversarial losses of $\mathcal{L}_{GAN}(G,D_Y,X,Y)$ and $\mathcal{L}_{GAN}(F,D_X,Y,X)$, a cycle consistency loss was also introduced to ensure that an input image $x$ in domain $X$ can be successfully reconstructed after going through both the forward mapping $G$ and the inverse mapping $F$, i.e., $x \to G(x) \to F(G(x)) \to \bar x$. The cycle consistency loss is described by
\begin{equation}
\begin{split}
\mathcal{L}_{cyc} (G,F)&=\mathbb{E}_{x\sim p_{data}(x)} \Vert F(G(x))-x \Vert_{1} \\
& + \mathbb{E}_{y\sim p_{data}(y)} \Vert G(F(y))-y \Vert_{1} .
\end{split}    
\end{equation}
Using CycleGAN alone allowed the generation of visually pleasant visible images from thermal images. However, adding more constraints to restrict the mappings is envisioned further improve the synthesis quality. In~\cite{CycleGANDetector2018}, a 68-point landmark detection network was used to extract the landmarks from the visible images. This can assist the CycleGAN to preserve the geometrical shapes of the reconstructed samples. The shape loss was defined as
\begin{equation}
\mathcal{L}_{shape}=\frac{1}{68} [(y_s-G(x)_{s})^2+(y_s-F(G(y))_{s})^2],
\end{equation}
where, $x$ is a thermal image and $G(x)$ is a synthesize visible image from generator $G$. $y$ is a visible image and $F(G(y))$ is a reconstructed visible image.

\subsubsection{Longwave with polarimetry}

\begin{figure*}[h!]
    \centering
    \includegraphics[scale = 0.25]{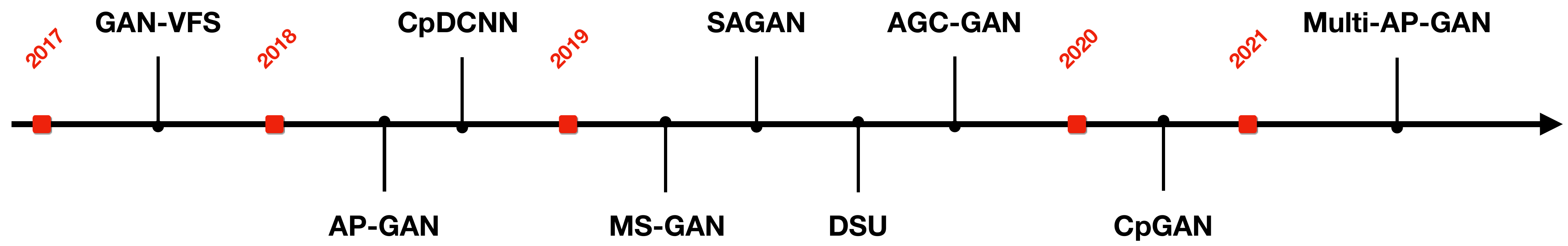}
    \caption{\textbf{Timeline of developments in algorithms:} LWIR Polarimetric-to-VIS face recognition.}
    \label{fig:Timeline_LWIR_Pola}
\end{figure*}

Polarimetric imagery in the thermal band has shown several advantages, particularly in bringing more details for both geometrical and textural information. In addition, \textbf{Experiment 2} with Table \ref{tab:SSIM_POLAR} were designed to show the effectiveness of algorithms in comparing face.                  
All LWIR polarimetry-to-VIS face comparison performance are reported on the Table \ref{LWIR Polarimetry_comparisons} and timeline of developments in algorithms are summarized in the Figure \ref{fig:Timeline_LWIR_Pola}.

\begin{table*}[t]
\caption{Rank-1 Accuracy and Verification Rate on different dataset for the \textbf{LWIR Polarimetry-VIS} face comparison.}
    \label{LWIR Polarimetry_comparisons}
    \centering
    \scriptsize
    \begin{tabular}{|c|c|c||l|c|}
    \hline
        \textbf{Year} & \textbf{Authors}& \textbf{Methods} & \textbf{Dataset} & \textbf{Performance $\% $} \\ \hline \hline
         
         2017 
         & Zhang \textit{et al.} \cite{zhang2017generative_GAN_VFS} & GAN-VFS & Polarimetric thermal \cite{Polarimetric_Thermal_Face} - \textit{Polar. Probe} & 93.82 @Rank-1 - 79.90 @AUC - 25.17 @EER \\ \hline \hline 
         
         \multirow{2}{*}{2018} & 
         Di \textit{et al.} \cite{di2018apgan} & AP-GAN & Polarimetric thermal \cite{Polarimetric_Thermal_Face} - \textit{Polar. Probe} &  88.93 @AUC - 19.02 @EER  \\ \cline{2-5}
         & Iranmanesh \textit{et al.} \cite{iranmanesh2018deep} & CpDCNN & Polarimetric thermal \cite{Polarimetric_Thermal_Face} - \textit{Polar. Probe} & 94.08 @Rank-1 \\ \hline \hline  
         
         \multirow{3}{*}{2019} 
         & Di \textit{et al.} \cite{DiSAGAN2019} & SAGAN & Polarimetric thermal \cite{Polarimetric_Thermal_Face} - \textit{Polar. Probe} & 96.41 @AUC - 10.02 @EER\\ \cline{2-5}
         
         & Iranmanesh \textit{et al.} \cite{iranmanesh2019attributeguided_AGC-GAN} & AGC-GAN & Polarimetric thermal \cite{Polarimetric_Thermal_Face} - \textit{Polar. Probe} & 96.54 @Rank-1 \\ \cline{2-5}
         
         & Zhang \textit{et al.} \cite{ZhangRHSP19} & MS-GAN & Polarimetric thermal \cite{Polarimetric_Thermal_Face}-\textit{ext.} - \textit{Polar. Probe} & 98.00 @AUC - 7.99 @EER\\ \hline \hline
         
         2020
         & Iranmanesh \textit{et al.} \cite{CpGAN_US_Army} & CpGAN & Polarimetric thermal \cite{Polarimetric_Thermal_Face} - \textit{Polar. Probe} & 95.49 @Rank-1 \\ \cline{2-5} \hline \hline
         
         2021
         & Di \textit{et al.} \cite{DiAttribute2020} & Multi-AP-GAN & Polarimetric thermal \cite{Polarimetric_Thermal_Face} - \textit{Polar. Probe} &  93.72 @AUC - 14.75 @EER \\  \hline
         
         \end{tabular}
\end{table*}

\ 

\textbf{GAN-VFS}. Zhang \textit{et al.} \cite{zhang2017generative_GAN_VFS} proposed a \textit{Generative Adversarial Network-based Visible face Synthesis} (GAN-VFS) algorithm to synthesize visible face images from their corresponding polarimetric images. The proposed GAN-VFS method contained three different modules, including visible feature extraction module, guidance sub-network module and visible image reconstruction module. The visible feature extraction module was used to extract the visible features from from the raw
polarimetric image. The guidance subnetwork module ensured that extracted visible features were instrumental in reconstructing visible images. The output of the guidance subnetwork constituted the guided features, which were finally used to reconstruct the visible image. This was optimized by the combination of identity loss and perceptual loss.


\textbf{AP-GAN}. Di \textit{et al}.~\cite{di2018apgan} proposed an \textit{Attribute Preserved Generative Adversarial Network} (AP-GAN) to synthesize the visible image from the polarimetric thermal image. A pre-trained VGG-Face model was first used to extract the attributes from the visible face image. Then, the extracted attributes were used as a prior to ensure that the synthesized visible face image from the polarimetric thermal image would share similar attributes. Finally, a pre-trained FR network was used to extract features from both visible and synthesized visible face images for comparison. 

\textbf{CpDCNN}. Iranmanesh \textit{et al.} \cite{iranmanesh2018deep} presented a \textit{Coupled Deep Convolutional Neural Network} (CpDCNN) to find and hence learn deep global discriminative features between polarimetric LWIR image faces and the corresponding visible faces. They used both thermal and polarimetric state information to bring more detail and then enhance the performance of thermal-to-visible CFR scenario. 
Their method was based on two VGG-16 coupled architectures, one \textit{Vis-DCNN} pretrained on the Imagenet dataset, dedicated to the visible spectrum and another one \textit{Pol-DCNN} initialized with the former one weights, focused to the polarimetric LWIR imagery. Together, Vis-DCNN and Pol-DCNN were linked via a \textit{contrastive} loss function $\mathcal{L}_{ct}$. 

\begin{figure*}[t!]
    \centering
    ~ 
    \begin{subfigure}[b]{0.52\textwidth}
        \includegraphics[width=1\textwidth,height=4.5cm]{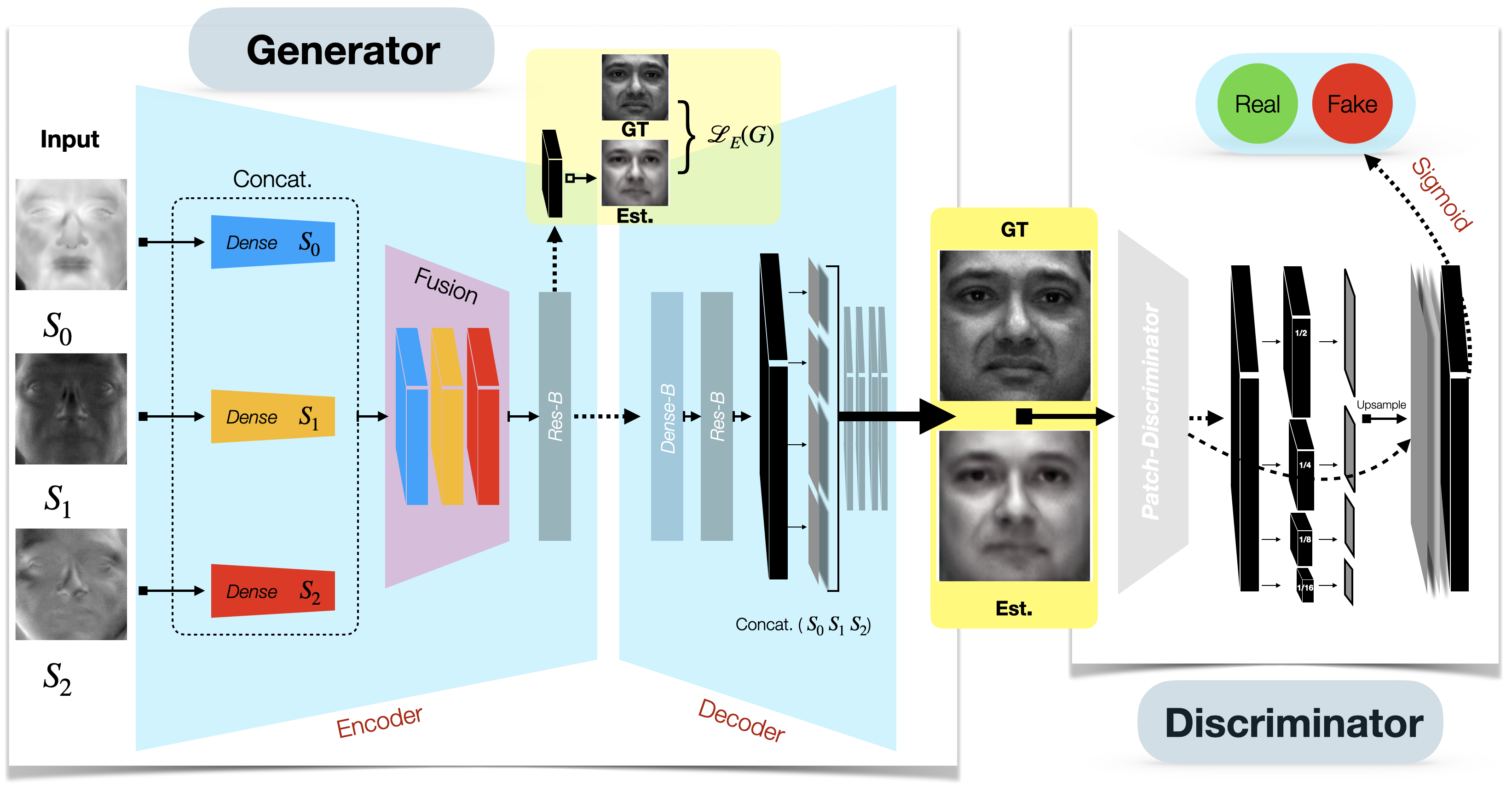}
        \caption{ \textbf{Flowchart of MS-GAN model.} The generator include a multi-stream feature-level fusion encoder-decoder network based on \textit{dense}-residual blocks. On the other hand, the discriminator involve a multi-scale patch-discriminator structure.}
        \label{fig:zhang2018synthesis_architecture}
    \end{subfigure}
    ~ 
    \begin{subfigure}[b]{0.45\textwidth}
        \includegraphics[width=1\textwidth,height=4.5cm]{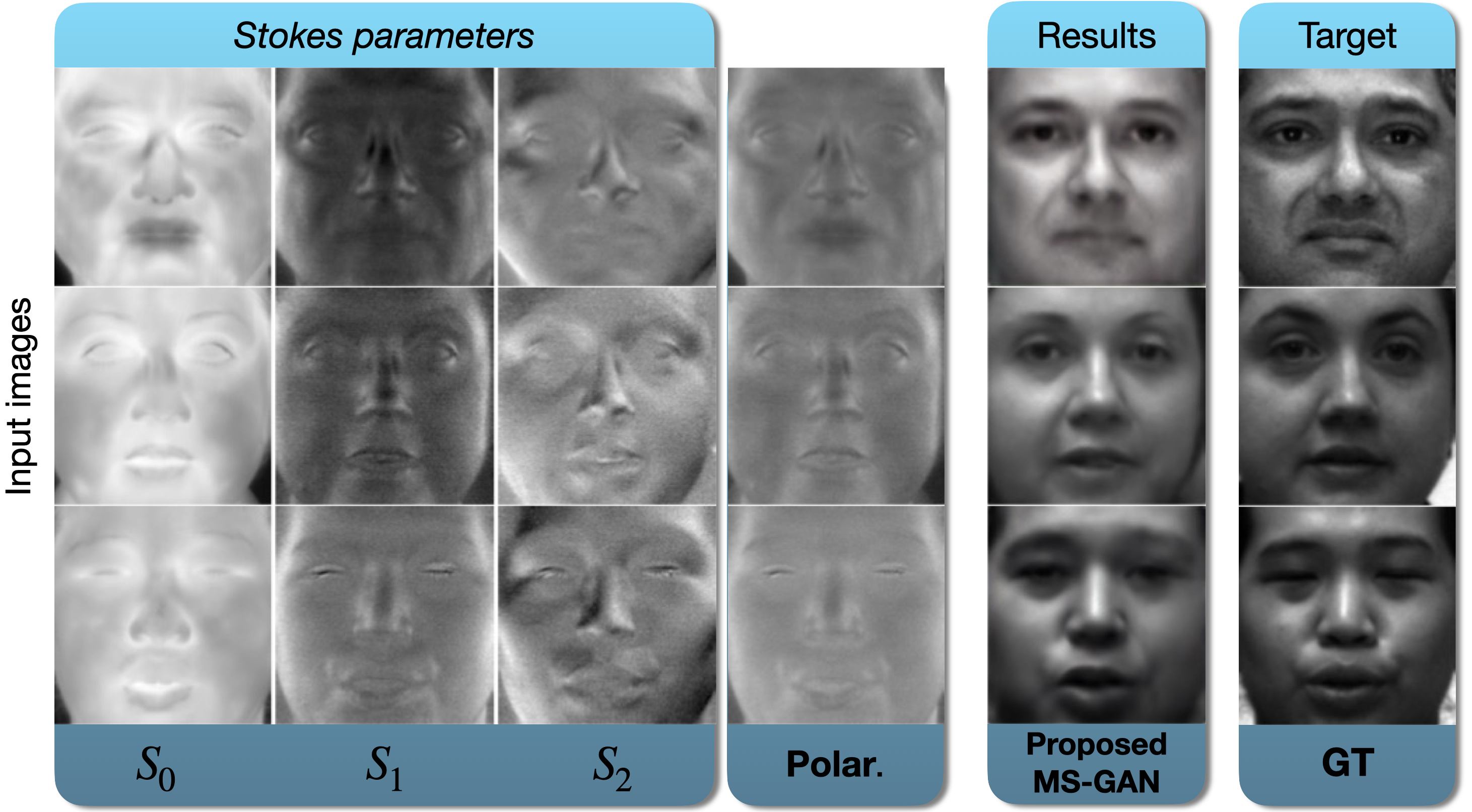}
        \caption{\textbf{Sample results of MS-GAN method.} Synthesizing   visible face images from the \textit{Stokes} parameter $S_0$, $S_1$, $S_2$ and the polarimetric thermal inputs. 
        Figure credit: Zhang \textit{al.} \cite{ZhangRHSP19}}
        \label{MS_GAN_Sample_results}
    \end{subfigure}
    \caption{ \textbf{Overview of the \textbf{MS-GAN} \cite{ZhangRHSP19} model:} Synthesize high-quality visible images.}
\end{figure*}

\textbf{MS-GAN}. Zhang \textit{et al.} \cite{ZhangRHSP19} were interested to synthesize high-quality visible images from polarimetric thermal images in order to generate photo realistic visible face images. Note that authors previous work \textbf{GAN-VFS} \cite{zhang2017generative_GAN_VFS} had adressed the same problem. 
Therefore, they presented a new \textit{Multi-Stream feature-level} (MS-GAN) fusion method based on GAN. The proposed architecture consisted of a \textit{generator} sub-network constructed using an encoder-decoder network based on dense-residual blocks, and a \textit{discriminator} sub-network to capture features at multiple scales. Further, to ensure that the encoded features contain geometric and texture information, a deep-guided sub-network was stacked at the end of the encoding part. Figure \ref{fig:zhang2018synthesis_architecture} illustrates the overview of the proposed architecture and Figure \ref{MS_GAN_Sample_results} shows corresponding visual results. 
Moreover, an extended sub-dataset from \cite{Polarimetric_Thermal_Face} was collected consisting of visible and polarimetric facial signature from 111 subjects. To be consistent with previous methods \cite{7791170, zhang2017generative_GAN_VFS}, two different protocols were introduced. Once the visible images were synthesized via the protocol presented below, the goal was to extract features in order to compare the \textit{synthesized visible} face with a \textit{visible} face. For the feature extraction task, they used the second last fully connected layer of the VGG-face network \cite{parkhi2015deep}. Best performances have been achieved with protocol 2. 



\textbf{SAGAN}. Di \textit{et al.}~\cite{DiSAGAN2019} proposed a synthesis network to convert between polarimetric thermal and visible images based on the \textit{Self-Attention GAN} (SAGAN). The self-attention module was added before the last convolutional layer of the generator and it was used as a weight matrix to refine the feature maps. 

\textbf{DSU}. Pereira \textit{et al.}~\cite{DSU2019} suggested that high-level CNN features trained in visible spectral images was domain independent and could be used to encode faces captured in different domains. The low-level CNN features that directly connected to the input signal were domain dependent. These low-level features could be learned by domain specific feature detectors, termed Domain Specific Units (DSU).

\textbf{AGC-GAN}. Iranmanesh and Nasrabadi  \cite{iranmanesh2019attributeguided_AGC-GAN} proposed
a novel \textit{Attribute-Guided Coupled Generative Adversarial
Network} (AGC-GAN) architecture that used facial attributes to improve the thermal-to-visible face recognition performance. The proposed AGC-GAN had two different generators, viz., Vis-GAN and Pol-GAN. The Vis-GAN and Pol-GAN were used to synthesize the corresponding visible image from the visible and polarimetric images, respectively. These two sub-networks were coupled using the contrastive loss function, aiming to find the latent feature vector that was shared by polarimetric face images and their corresponding visible counterparts. 

\begin{figure*}[]
    \centering
    \includegraphics[width=1\textwidth]{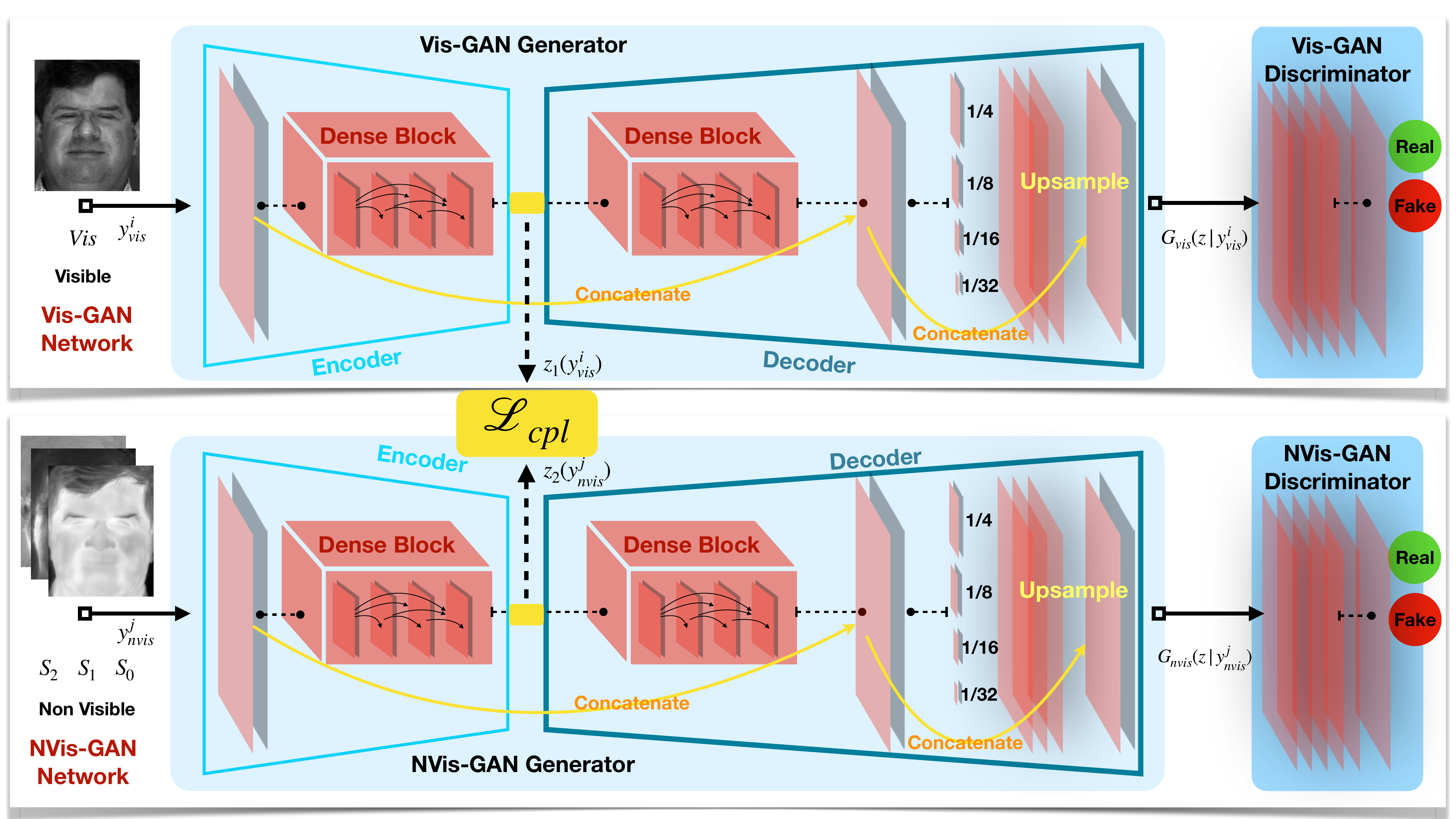}
    \vspace{-0.1 cm}
    \caption{Flowchart of \textbf{CpGAN} \cite{CpGAN_US_Army} using two GAN based sub-networks (Vis-GAN and NVis-GAN) coupled by contrastive loss function $\mathcal{L}_{cpl}$.}
    \label{CpGAN_Network}
\end{figure*}

\textbf{CpGAN}. Iranmanesh \textit{et al.} \cite{CpGAN_US_Army} proposed a \textit{Coupled Generative Adversarial Network} (CpGAN) depicted in Figure \ref{CpGAN_Network}, to synthesize visible image from non-visible image such as polarimetric thermal (channel input of $S_0$, $S_1$ and $S_2$) for CFR task. 

CpGAN contained two GAN-based sub-networks, with each two appropriate generators and discriminators, dedicated to visible (Vis) and non-visible (NVis) input images. The proposed network was capable of transforming the visible and non-visible modalities into a common discriminative embedding subspace and subsequently synthesizing the visible images from that subspace. In order to efficiently synthesize a realistic visible image from the non-visible modality, a densely connected encoder-decoder structure was used as the generator in each sub-network. The final objective of their proposed CpGAN was to learn a joint deep embedding that captured the interrelationship between the visible and non-visible facial imagery for spectrally invariant face recognition. 

         
          

\textbf{Multi-AP-GAN}. Di \textit{et al.}~\cite{DiAttribute2020} proposed a novel \textit{Multi-Scale Attribute Preserved Generative Adversarial Network} (Multi-AP-GAN) to synthesize the visible image from the polarimetric thermal image with preserved attributes. This work was an extension to their previous work on AP-GAN~\cite{di2018apgan}. The attributes were extracted from visible face images by a pre-trained VGG-Face model. The Multi-AP-GAN consisted of two major components: a multimodal compact bilinear (MCB) pooling-based generator and a generator with the multi-scale architecture. The MCB pooling method was used to concatenate  image feature vector and semantic attribute vector, rather than using a simple concatenation. The multi-scale generator was designed to generate multiple intermediate outputs at different resolution scales. The discriminator was applied to individual multi-scale resolution outputs. The proposed Multi-AP-GAN was regularized by a multi-scale adversarial loss, a perceptual loss, an identity loss and a reconstruction loss.

\begin{center}
    \textbf{Summary of Polar-LWIR to VIS Face Comparison}
\end{center}

Comparing polarimetric thermal to visible face images remains a challenge problem due to large domain discrepancy. Nearly all works have proposed GANs to synthesize visible images from the polarimetric thermal images~\cite{ZhangRHSP19, DiSAGAN2019, CpGAN_US_Army, DiAttribute2020}, rather than using CNNs to learn a domain-invariant feature representation. Associated GANs were based on conditional GANs, where mapping was learned from an observed image $x$ and a random noise vector $z$ to an output $y$: $G:\{x,z\}\leftrightarrow y$. The objective function of a conditional GAN is defined~\cite{pix2pix2017} as
\begin{equation}
\begin{split}
\mathcal{L}_{GAN}(G,D) & = \mathbb{E}_{x,y \sim p_{data}(x,y)}[\log D(x,y)]  \\
                                        & +  \mathbb{E}_{x \sim p_{data}(x)}[\log (1-D(x, G(x)))],\\
\end{split}
\end{equation}
where $G$ is the generator, $D$ is the discriminator, $x$ is the thermal image and $y$ is the target visible image. $p_{data}(x)$ indicates that $x$ is from the true data distribution and $p_{data}(x,y)$ indicates that both $(x,y)$ are from the true data distribution. The objective of generator $G$ is to synthesize visually realistic visible face images from thermal face images, while the discriminator $D$ is structured to distinguish the target visible face images from the synthesized ones, conditioned on the input thermal image. In other words, the input to a discriminator $D$ is a pair of samples $(x,y)$. The adversarial loss function $\mathcal{L}_{GAN}(G,D)$ only ensures that we can generate synthesized visible images $G(x)$ that are ''real", but it does not provide information whether the generated sample is close to the target visible image at the per-pixel level. This can be remedied by using a per-pixel loss function, $L_1$
\begin{equation} \label{eu_eqn}
\mathcal{L}_1=\mathbb{E}_{x,y \sim p_{data}(x,y)} \Vert  G(x)-y \Vert_{1}.
\end{equation}
The final objective function is described by
\begin{equation}
\begin{split}
G^{*} & = \arg \min_{G}\max_{D} \lambda_G\mathcal{L}_{GAN}(G,D)+\lambda_1\mathcal{L}_1(G).
\end{split}
\end{equation}
Using jointly, adversarial loss $\mathcal{L}_{GAN}(G,D)$ and  $\mathcal{L}_1(G)$ serves as a foundation for these proposed methods~\cite{ZhangRHSP19, DiSAGAN2019, CpGAN_US_Army, DiAttribute2020}. To further improve the comparison performance, more loss functions are employed to constrain the mapping space.

\section{Datasets}\label{sec:datasets}
\begin{figure*}[h]
    \centering
    ~ 
    \begin{subfigure}[b]{0.45\textwidth}
        \includegraphics[width=0.9\textwidth,height=3cm]{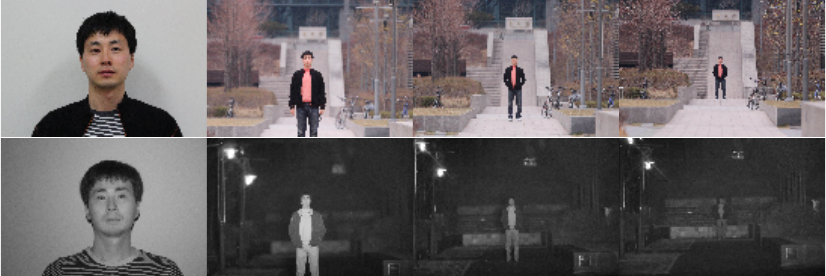}
        \caption{LDHF NIR-VIS}
        \label{ldhf_nir}
    \end{subfigure}
    ~ 
    \begin{subfigure}[b]{0.45\textwidth}
        \includegraphics[width=0.9\textwidth,height=3cm]{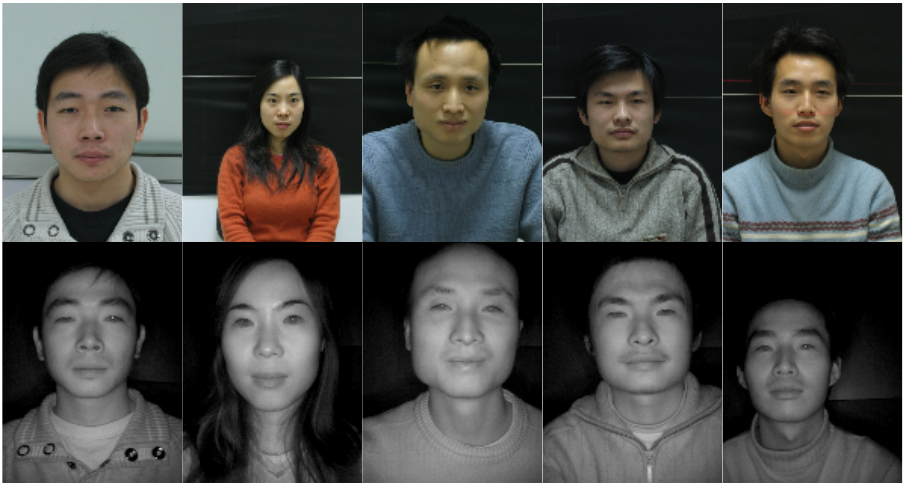}
        \caption{CASIA NIR-VIS}
        \label{casia_nir}
    \end{subfigure}
    \begin{subfigure}[b]{0.45\textwidth}
        \includegraphics[width=0.9\textwidth,height=3cm]{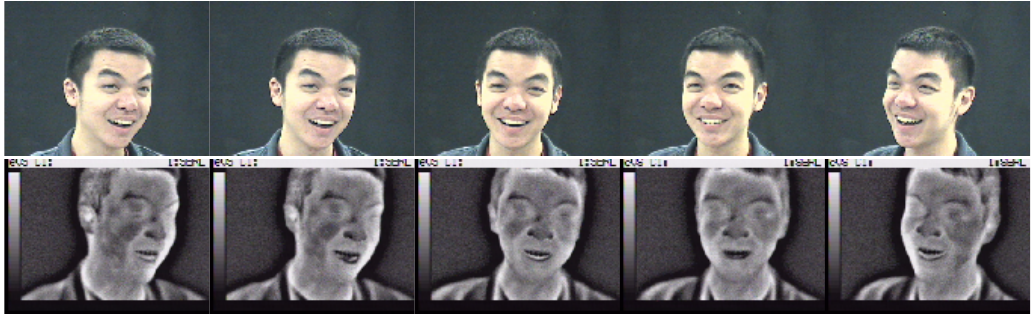}
        \caption{IRIS MWIR-VIS}
        \label{iris_mwir}
    \end{subfigure}
    \begin{subfigure}[b]{0.45\textwidth}
        \includegraphics[width=0.9\textwidth,height=3cm]{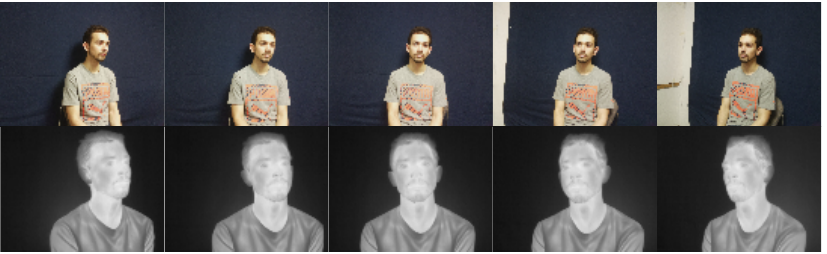}
        \caption{TUFTS LWIR-VIS}
        \label{tufts_lwir}
    \end{subfigure}    
    \vspace{-0.3 cm}
    \caption{Illustration of heterogeneous face recognition datasets.}\label{HFR_datasets}
    
\end{figure*}

Most significant research achievements made over the past decade have heavily relied on new datasets and on designing benchmarks for evaluating CFR-algorithms.  
As deep learning approaches are known to be data-hungry, training data and hence publicly available datasets are a pertinent key stone in the advancement of CFR.
We proceed to review such existing datasets, grouping them with respect to spectral bands, in which images were acquired. We summarize discussed datasets in Table~\ref{tab:Datasets}.  Figure~\ref{HFR_datasets} illustrates samples from representative CFR-datasets. 

\textbf{NIR-VIS}. The \textit{CASIA HFB}~\cite{CASIA_HFB} is an early benchmark dataset developed for CFR. \textit{CASIA HFB} comprises of 57 male and 43 female individuals (100 in total). Per subject 4 NIR and 4 VIS face images, as well as a 3D depth image have been acquired. \textit{CASIA NIR-VIS 2.0} \cite{CasiaNIRVIS} is an extension to the \textit{CASIA HFB} dataset and constitutes one of the first large NIR-VIS face dataset, containing 17,580 NIR-VIS face images of 725 subjects with variations in pose, expressions, eyeglasses, and distances. Compared to HFB, NIR-VIS 2.0 has several important features: (a) the number of subjects has a threefold increase; (b) the age distribution of the subjects is significantly larger; and (c) an evaluation protocol was proposed for benchmarking. Based on these dataset properties, \textit{CASIA NIR-VIS 2.0} has been the most widely used benchmark dataset for evaluating NIR-VIS CFR. In addition to \textit{CASIA HFB} and \textit{NIR-VIS 2.0} datasets, there also exist benchmark datasets that exhibit other properties. The \textit{Oulu-CASIA NIR-VIS} dataset~\cite{chen2009learning} is composed of 80 subjects with six expression variations (anger, disgust, fear, happiness, sadness, and surprise). As the facial images of this database are captured under different environments from two institutes, the related illumination conditions differ slightly. We note that compared to \textit{CASIA NIR-VIS 2.0}, \textit{Oulu-CASIA NIR-VIS} is more suitable to study NIR-VIS CFR under facial expression changes. In addition, per subject 48 VIS samples and 48 NIR samples are provided.
The \textit{BUAA-VisNir}  dataset~\cite{huang2012buaa} is often used for domain adaptation evaluation. This dataset contains 150 subjects, with 9 VIS images and 9 NIR images of each subject that correspond to 9 distinct poses or expressions. The training set and testing set are composed of 900 images of 50 subjects and 1800 images of the remaining 100 subjects, respectively. This evaluation protocol is more challenging due to the larger pose and illumination variations observed between the training and test sets. 
To further expand the range of variability of training samples, a new Large-Scale Multi-Pose High-Quality database \textit{LAMP-HQ} \cite{yu2019lamp} containing over 73000 images from 573 individuals with large diversities in pose, illumination, attribute, scene, and accessory was proposed.
Nevertheless, these datasets have yet to address the problem of outdoor CFR at a long distance. The \textit{LDHF-DB} dataset~\cite{Maeng2012NighttimeFR} offers subjects captured at distances of 60 meters, 100 meters, and 150 meters, in both VIS and NIR spectral bands. Therefore, LDHF-DB is highly instrumental in investigating cross-distance and cross-spectral NIR-VIS FR.


\textbf{SWIR-VIS}. The \textit{PRE-TINDERS}~\cite{TINDERS} dataset is composed of 48 frontal face subjects and a total of 384 images captured at two wavelengths: visible spectrum and SWIR spectrum at 1550 nm. 4 images per subject are available for each spectral band: 2 images have neutral expression and 2 images depict the person talking (open mouth). All images in this dataset are acquired at close distance from the camera (about 2 meters) in a single session. A light source at 1550 nm is illuminating the face of the subjects for images captured in the SWIR spectral band. The original resolutions of the acquired images before normalization are $640\times512$ for SWIR images and $1600\times1200$ for visible images. The \textit{TINDERS} \cite{TINDERS} dataset is composed of 48 frontal face subjects having images at the same two spectral bands. The SWIR images are acquired at two long ranges (50 and 106 meters). At both distances, 4 or 5 images per class are available: two/three entail neutral expression and two have talking expression. A total of 478 images are available in the SWIR band. The visible (color) images were collected at a short distance and in two sessions (3 images per session), and all of them have neutral expression, resulting in a total of 288 images. The original resolutions of images before normalization are $640\times512$ for SWIR images and $640\times480$ for visible images.

\textbf{MWIR-VIS}. The \textit{WSRI} dataset \cite{WSRI_dataset} comprises of 1615 visible and 1615 MWIR images from 64 subjects. There are 25 images per individual including different facial expressions. Visible images were captured at resolution $1004\times1004$, while MWIR images were captured at resolution $640\times512$. Face images from both modalities were preprocessed, being resized to $235\times295$ pixels. Training and testing set were split randomly into a set of 10 subjects and 54 subjects, respectively. \textit{MILAB-VTF(B)} \cite{peri2021synthesis_IEEE} is the largest collection of long-range unconstrained paired MWIR-Visible face images, including 400 different identities with both, indoor and outdoor acquisitions. 


\textbf{LWIR-VIS}. The \textit{UND X1} dataset \cite{135} contains LWIR and visible images related to 241 subjects with different variations in lighting, expression and time lapse. The original resolutions of the images are 1600x1200 pixels for the visible modality and 320x240 pixels for the LWIR modality. Both modalities are resampled to 150x110 pixels after preprocessing. The training set composed of 159 subjects was captured in the visible and LWIR modalities, with only one image per subject. On the other hand, the test set contained remaining 82 subjects with multiple images per subject. This database is challenging due to the low resolution and noise present in the LWIR imagery. This leads to significant difference between the two modalities in this dataset. The \textit{TUFTS} \cite{Tufts} dataset contains over 10,000 images from 113 individuals. Images were acquired in various modalities, including visible, thermal images, near infrared images images of individual face. The \textit{EURECOM} \cite{8553431EURECOM} dataset contains seven different facial variations, five different illumination conditions, five different types of occlusions, and four different head poses that belong to $50$ subjects. In total, there are $2100$ images in the dataset. The \textit{ARL-VTF} \cite{Poster_2021_WACV_ARL_VTF} dataset is the newest and largest collection of time-synchronized visible-thermal face images presented in this survey. The dataset includes 395 subjects and 549,712 images with variations in expression, pose and eyewear. ARL-VTF is endowed with annotations, metadata as well as standardized protocols for fair evaluation.  

\textbf{Polar-VIS}. The \textit{Polarimetric Thermal Face} dataset \cite{Polarimetric_Thermal_Face} (also called ARL-MMFD) contains polarimetric LWIR and visible face images of 60 subjects, later extended to 111 subjects \cite{ZhangRHSP19}. Data was collected at three different distances: Range 1 (2.5 m), Range 2 (5 m), and Range 3 (7.5 m). At each range, baseline and expressions data were collected and annotated. In the baseline condition, the subject was asked to maintain a neutral expression while looking at the polarimetric thermal sensor. On the other hand, in the expression condition, the subject was asked to count numerically upwards from one, resulting in different expressions in the mouth to eye regions. Each subject has 16 images of visible and 16 polarimetric LWIR images in which four images are from the base-line condition and the remaining 12 images are from the expression condition.

\textbf{Multiple-Spectral}. The \textit{NVESD} dataset \cite{byrd2013preview_NVESD} was collected jointly by the U.S. Army NVESD in 2013 from 50 different subjects in different scenarios and settings including physical exercise (fast walk) and different ranges (1m, 2m and 4m). The dataset is composed of 450 images in each modality. The images were captured simultaneously from different identities with the original resolution of $640 \times 480$ pixels for all of the modalities. After preprocessing, the image resolution is resampled to $174 \times 174$ and the dataset is split into training and testing sets.

\textbf{In-the-Wild Dataset.}
The \textit{Thermal Faces in the Wild} (TFW) dataset \cite{kuzdeuov2021tfw} was the first dataset collected in both indoor and outdoor environments under controlled and in-the-wild conditions, respectively. The TFW dataset includes 145 subjects and a total of $9202$ images with variation in pose. In addition, images are associated with manually labeled bounding boxes and five facial landmarks.



\begin{table*}
\caption{A quick reference summary of the main databases of face images acquired in the infrared spectrum.}
    \label{tab:Datasets}
    \centering
    \begin{tabular}{|c|l|c||c|c|c|c|c|c||c|c|}
    \hline 
         \multirow{2}{*}{\textbf{Year}} & \multirow{2}{*}{\textbf{Name}} & \multirow{2}{*}{\textbf{Ref.}} & \multicolumn{6}{c||}{\textbf{Spectrum}} & \multicolumn{2}{c|}{\textbf{Number of}}   \\
         & & & \scriptsize{\textit{Vis}} & \scriptsize{\textit{NIR}} & \scriptsize{\textit{SWIR}} &\scriptsize{\textit{MWIR}} & \scriptsize{\textit{LWIR}} & \scriptsize{\textit{Pola.}} & \textit{Subjects} & \textit{Images} \\
        \hline \hline 
        
2003 & 
UND X1  &  \cite{135}  & \checkmark &  & & & \checkmark & & 241 & 4584 \\ \hline \hline 

2006 & IRIS-M3 & \cite{134} & \checkmark &  & & & \checkmark & & 82 & 2624 \\ \hline \hline 

\multirow{2}{*}{2009} 
& Oulu-CASIA NIR-VIS & \cite{chen2009learning} & \checkmark & \checkmark & & & & & 80 & 7680 \\ \cline{2-11}
& CASIA HFB & \cite{CASIA_HFB}  & \checkmark & \checkmark & & & & & 100 & 5097 \\ \hline \hline

\multirow{3}{*}{2010}
& TINDERS &  \cite{TINDERS}  &\checkmark&&\checkmark&&&& 48 & 1255 \\ \cline{2-11}
& USTC-NVIE & \cite{wang2010natural} &\checkmark&&&&\checkmark& & 103 & 7460 \\ \cline{2-11}
& PolyU-NIR & \cite{137} &&\checkmark&&&& & 335 & 34000 \\ \hline \hline

2011 
& PRE-TINDERS & \cite{TINDERS} &\checkmark&&\checkmark&&&& 48 & 576 \\ \hline \hline 

\multirow{4}{*}{2012}
& BUAA-VisNir & \cite{huang2012buaa} & \checkmark & \checkmark & & & & & 150 & 2700 \\ \cline{2-11} 
& LDHF-DB & \cite{Maeng2012NighttimeFR} & \checkmark & \checkmark & & & & & 100 & - \\ \cline{2-11}
& Equinox  & \cite{131} &\checkmark & &\checkmark & \checkmark & \checkmark &&  90 & 25000 \\ \cline{2-11}  
& IRIS & \cite{133} & \checkmark &&&&\checkmark &&  32& 8456 \\ \hline \hline

\multirow{4}{*}{2013}
& NVESD & \cite{byrd2013preview_NVESD} &\checkmark & &\checkmark & \checkmark & \checkmark &&  50 & 900 \\ \cline{2-11} 
& CASIA NIR-VIS 2.0 & \cite{CasiaNIRVIS} &\checkmark & \checkmark &&&&& 725 & 17580 \\\cline{2-11}
& Carl Database & \cite{CarlDatabase} & \checkmark & \checkmark & & & \checkmark & & 41 & 7380 \\ \cline{2-11}
& PCSO & \cite{klare2012heterogeneous} & \checkmark &&&&\checkmark && 1003 &3011\\
\hline \hline

2015 
& ND-NIVL & \cite{ND-NIVL_Dataset} &\checkmark & \checkmark &&&&& 574 & 24605 \\
\hline \hline

2016 
& Polarimetric thermal &  \cite{Polarimetric_Thermal_Face}  & \checkmark &&&& \checkmark &\checkmark & 60 & 14400 \\ \hline \hline 

\multirow{2}{*}{2018}
& EURECOM & \cite{8553431EURECOM} & \checkmark & & & & \checkmark & & 50 & 2100 \\ \cline{2-11}
& UL-FMTV & \cite{ghiassuniversite_UL-FMTV_database} &\checkmark & & & \checkmark & & & 238 & -  \\ \hline \hline 

2019
& IJB-MDF & \cite{9186007}  & \checkmark &  & \checkmark & \checkmark & \checkmark & & 251 & - \\ \hline \hline

\multirow{3}{*}{2020}
& LAMP-HQ & \cite{yu2019lamp}  & \checkmark & \checkmark & & & & & 573 & 73616 \\ \cline{2-11} 
& Tufts  & \cite{Tufts} & \checkmark & \checkmark &&&\checkmark && 113 & 10000 \\ \cline{2-11}
& HQ-WMCA & \cite{Heusch_TBIOM_2020} & \checkmark & \checkmark & \checkmark & & \checkmark & & 51 & 58080 \\ \hline \hline

\multirow{4}{*}{2021}
& ARL-VTF & \cite{Poster_2021_WACV_ARL_VTF} &\checkmark & & & & \checkmark & & 395 & 549712 \\ \cline{2-11} 
& MILAB-VTF(B) & \cite{peri2021synthesis_IEEE} &\checkmark & & & \checkmark  & & & 400 & - \\  \cline{2-11} 
& TFW & \cite{kuzdeuov2021tfw} & & & & & \checkmark & & 145 & 9202 \\  \cline{2-11} 
& Speaking faces & \cite{abdrakhmanova2021speakingfaces} & \checkmark & & & & \checkmark & & 142 & 7668 \\ \hline \hline

\multirow{2}{*}{2022}
& SF-TL54 & \cite{kuzdeuov2022sf} & \checkmark &  & & & \checkmark & & 142 & 2256 \\ \cline{2-11}
 & MCXFace & \cite{George_IEEETIFS_2022} & \checkmark & \checkmark & \checkmark & & \checkmark & & 51 & 7406 \\ \hline \hline

\multirow{2}{*}{2023} 
& BYDB & \cite{anghelone2023computer} & \checkmark &  \checkmark & \checkmark &  & \checkmark & & 412 & 1476292 \\ \cline{2-11}
& LVT & \cite{10345997} & \checkmark & & & & \checkmark & & 52 & 612 \\ \hline \hline

- & WSRI & \cite{WSRI_dataset} &\checkmark & & & \checkmark & & & 64 & 3230 \\ \hline   

    \end{tabular}
\end{table*}

\section{Open Research Challenges}\label{sec:Future Direction}

Despite significant achievements, CFR still faces a number of challenges. In this section, we proceed to identify and discuss some key challenges that are essential for further success of CFR.

\textbf{Scarcity of datasets.} Unlike traditional FR, where millions of faces can be leveraged to train deep learning based models, CFR remains significantly impeded by lack of large-scale heterogeneous face datasets. In addition, very few existing datasets include annotations of face bounding box and landmarks~\cite{Kopaczka2019}. Such annotations are essential for developing automated face and landmark detection for CFR. 

\textbf{Large standoff distance.} Limited work has focused on CFR at large standoff distances~\cite{Maeng2012NighttimeFR}. Most existing studies assume that faces are captured at short standoff distances. However, capturing faces at large standoff distances results in face image quality degradation, including low-resolution, blur and noise.  Early attempts towards addressing this challenge have sought to restore high-quality images from degraded ones~\cite{Maeng2012NighttimeFR, KANG2014}. Such methods were developed by learning a locally linear mapping between  low-quality patches and high-quality patches. The linear mapping assumption has been relaxed at the patch-level, while in practice, the mapping is often considered non-linear. In this context, it has been demonstrated that GANs are highly suitable to model non-linear mappings.  


\textbf{Modality gap.} The key problem with CFR is the apparent disparity between visible and invisible spectrum, which strongly impacts the comparison performance (see Experiment 2). Although cross-spectrum featured-based methods, such as \textit{transfer learning} or \textit{domain adaptation} have achieved good results, especially in the \textit{Active IR} band, they remain less efficient in the \textit{Passive IR} band. An obvious solution is to incorporate details related to both geometrical and textural information. Unsupervised-processing methods with deep generative models, such as cross-spectrum images synthesis, appear to be a promising trend, particularly with improvements in \textit{GANs}. So far, "synthetic data" has not yet outperformed the state-of-the-art in CFR, but we foresee promising results in the near future.




\textbf{Privacy.}
Infrared imaging is a keystone biometric solution in terms of security and privacy \cite{hu2017heterogeneous}. Firstly, humans are unable to recognize individuals from IR images, as IR appearances do not resemble RGB images (Figure \ref{fig:NIR_SWIR_MWIR_LWIR_face}). Secondly, IR imaging ensures pseudo-anonymized recognition and compliance with legal requirements regarding personal data and template protection. Finally, IR is robust against attempted diversion and spoofing attacks, as artifacts are directly identifiable by appearing differently on face images. While intuitive, the above has not been studied rigorously in scientific literature.

\textbf{Presentation Attacks}. FR systems have shown to be vulnerable to presentation attacks, where adversaries use artifacts such as prints, video replays, makeup, and 3D masks to undermine the system. Reflection difference between live faces and spoof faces under NIR~\cite{JIANG201930, NIR-ANTISPOOFING}, SWIR~\cite{SteinerSWIRSpoofing} or LWIR~\cite{spinoulas2020multispectral} spectral bands can be exploited to detect such presentation attacks. CFR can provide a unique insight, complementing RGB-imagery, as it imparts liveness detection, as well as shape determination~\cite{Heusch_TBIOM_2020}. 

\textbf{Next generation of synthesis models for CFR}.
As the field of deep learning continues to evolve, new and innovative methods for generating synthetic images are emerging. While GANs have been popular in recent years for translating thermal face images into visible-like face images \cite{peri2021synthesis_IEEE, AnghChen_LGGAN_FG2021, ChenR19AGGAN, ChenR19, anghelone2023ANYRES}, they still face several limitations such as \textit{mode collapse} and \textit{instability} during training. 
In contrast, models based on the diffusion process, such as the Diffusion Probabilistic Models (DPMs), offer a promising alternative for generating high-quality synthetic images with greater stability and less distortion, preserving identity better.
Hence, transitioning from GANs to DPMs opens up exciting new possibilities for generating interspectral images with improved realism and accuracy, rendering them a worthwhile avenue of future research with new scientific keystones.
In this context, a pioneer work has explored thermal to visible face translation by proposing a Denoising Diffusion Probabilistic Model \cite{nair2023t2v}. 
By exploring and developing these new models, the field of computer vision will advance, towards creating more sophisticated and reliable systems for a wide range of applications related to CFR.




\section{Conclusions}
In this survey, we reviewed methods that have been developed for cross-spectral face recognition (CFR), placing emphasis on recently proposed deep CNN-architectures. We discussed associated benefits and limitations, as well as the datasets used. CFR is endowed with the ability to identify people in challenging illumination settings, as well as to detect presentation attacks. Such benefits have been exploited in applications such as \textit{defense}, \textit{surveillance} and \textit{public safety}. 

While traditional FR is restricted to comparing images in the visible spectrum, CFR is inherently more challenging, as it must address both subject identification and spectral variation. In order to separate spectral variation from identity, approaches such as domain-invariant feature representation have been proposed. Despite such CNN-based approaches usually performing well in minimizing spectral variation between NIR and visible spectral bands, the large modality gap introduced by capturing face images in the thermal spectral band can be effectively addressed using GANs. In CNN-based approaches designed to learn domain-invariant features, the \textit{triplet} loss function has demonstrated better performance than the \textit{contrastive} loss function, since the former considers both intra-class minimization and inter-class maximization. For GAN-based approaches, the \textit{identity} loss function is essential to ensure that synthesized faces preserve the same identity as the ground-truth face. Other loss functions such as \textit{attribution} loss, \textit{shape} loss, \textit{perceptual} loss or \textit{semantic} loss are utilized as auxiliary loss functions.

We here conclude that the NIR-to-visible CFR scenario has been extensively studied in the literature, achieving high accuracy. However, SWIR and MWIR sub-bands remain widely unused due to their limited practicality and high equipment costs. We note that the COVID-19 pandemic has brought to the fore a significant decrease in the cost of LWIR sensors, rendering such sensors more accessible for a wider range of applications. This, coupled with the increasing demand for surveillance and protection solutions, e.g., against spoofing attacks, has made LWIR imaging an appealing option for various industries.

In terms of spectral band selection, NIR is preferred for reflective IR-to-Visible FR in the majority of applications. There is not much difference between MWIR or LWIR spectral bands in emissive IR-to-Visible face recognition. Nevertheless, the use of polarimetry for LWIR has been attributed to better face comparison results. While NIR and SWIR spectral bands can produce relatively good face resolutions, this remains a challenge for MWIR and LWIR sensors. Towards improving the poor resolution associated with thermal images, there is a necessity to develop high-resolution sensors by integrating super-resolution post-processing schemes. CFR has been impeded by lack of large publicly available face datasets. With the emergence of more comprehensive and large-scale datasets, such as ARL-VTF \cite{Poster_2021_WACV_ARL_VTF} and MILAB-VTF(B) \cite{peri2021synthesis_IEEE}, we foresee further improvements in CFR performance. 

To conclude, CFR has several challenging issues, and this survey has reiterated the significant need for systematic theoretical research in this field, rather than empirical experimentation with different loss functions and networks. Such concerns necessitate the introduction of robust and reliable methods to identify forged images and videos.











\bibliographystyle{unsrt}
\bibliography{egbib}

\end{document}